\documentclass[]{youtu} 

\usepackage{natbib}

\usepackage{times}
\usepackage{latexsym}
\usepackage[T1]{fontenc}
\usepackage[utf8]{inputenc}
\usepackage{bbold}
\usepackage{inconsolata}

\usepackage{microtype}
\usepackage{parskip}
\usepackage{ragged2e}
\usepackage{amsmath}
\usepackage{amssymb}
\usepackage{pifont}    
\newcommand{\cmark}{\ding{51}}
\newcommand{\xmark}{\ding{55}}
\newcommand{\pmark}{\ding{108}~}

\usepackage{xcolor}
\definecolor{softpurple}{RGB}{120,80,160}

\usepackage{graphicx}
\usepackage{float}
\usepackage{adjustbox}

\usepackage{booktabs}
\usepackage{array}
\usepackage{multirow}
\usepackage{tabularx}
\usepackage{makecell}
\usepackage{arydshln}
\newcolumntype{Y}{>{\raggedright\arraybackslash}p{0.24\textwidth}}

\newcommand{\betweensize}{\fontsize{7pt}{6pt}\selectfont}

\usepackage{caption}
\usepackage{subcaption}

\usepackage{enumitem}

\usepackage{fvextra}   
\usepackage{listings}
\usepackage{xurl}

\usepackage[ruled,vlined]{algorithm2e}
\usepackage{algorithmic}

\newcommand{\turn}[2]{\textbf{#1}: #2 \par\vspace{0.3em}}
\newcommand{\action}[1]{\textit{\textcolor{blue}{(#1)}}}
\newcommand{\thought}[1]{\textit{\textcolor{softpurple}{[#1]}}}
\newcommand{\env}[1]{\textit{\textcolor{teal}{<#1>}}}

\title{AdaMARP: An \underline{Ada}ptive \underline{M}ulti-\underline{A}gent Interaction Framework for General Immersive \underline{R}ole-\underline{P}laying
}
\author{\small Zhenhua Xu\textsuperscript{$1 \dagger$}, Dongsheng Chen\textsuperscript{$2 \dagger$} \\ Shuo Wang\textsuperscript{$2$}, Jian Li\textsuperscript{$2$}, Chengjie Wang\textsuperscript{$2$},  Meng Han\textsuperscript{$1$}, Yabiao Wang\textsuperscript{$2$}}
\vspace{-5mm}
\affiliation{$^1$Zhejiang University,\textsuperscript{$2$}Tencent Youtu Lab}

\date{Jan. 16, 2026}
\correspondence{mhan@zju.edu.cn}
\projectleader{caseywang@tencent.com}
\sourcecode{https://xuzhenhua55.github.io/AdaMARP}

\youtufinalcopy

\begin{document}
\vspace{0.6cm}
\abstract{
LLM role-playing seeks to portray arbitrary characters in interactive narratives, yet existing systems often lack immersion and adaptability. They typically under-model \emph{dynamic environment information} and assume a largely static scene/cast, offering limited support for multi-character orchestration, scene transitions, and on-the-fly character introduction.
We propose an adaptive multi-agent interaction framework dubbed \textbf{AdaMARP}, which featuring an immersive message format that interleaves \texttt{[Thought]}, \texttt{(Action)}, \texttt{$<$Environment$>$}, and \texttt{Speech}, and an explicit \textbf{Scene Manager} that controls role-playing via discrete actions (\texttt{init\_scene}, \texttt{pick\_speaker}, \texttt{switch\_scene}, \texttt{add\_role}, \texttt{end}) with rationales.
To train these abilities, we construct \textbf{AdaRPSet} for the Actor Model and \textbf{AdaSMSet} for supervising orchestration decisions, and introduce \textbf{AdaptiveBench} for trajectory-level evaluation.
Experiments across multiple backbones and scales show consistent gains: AdaRPSet improves character consistency, environment grounding, and narrative coherence—an 8B actor outperforming several commercial LLMs, while AdaSMSet enables smoother scene transitions and more natural role introductions, surpassing Claude Sonnet 4.5 with only 14B LLMs.
}
 \maketitle

\section{Introduction}
\label{sec:introduction}
\renewcommand{\thefootnote}{}
\footnotetext{$\dagger$ Equal contribution.}

\setcounter{footnote}{0}
\renewcommand\thefootnote{\arabic{footnote}}

Recent advances in large language models (LLMs) have substantially improved their general-purpose capabilities across many language tasks~\citep{tan2025prospectretrospectreflectivememory,zhang2025osccognitiveorchestrationdynamic,zhuang2025selftaughtagenticlongcontext,zhang2025llmcasdynamicneuronperturbation,xuCTCCRobustStealthy2025a,kong2025surveyllmdrivenaiagent,xuEverTracerHuntingStolen2025b,li2025iag,xu2025copyright}. One prominent application is \emph{LLM role-playing}—prompting an LLM to portray a character and interact with a user in an ongoing narrative—which has gained rapid popularity~\citep{shanahanRolePlayLargeLanguage2023,tsengTwoTalesPersona2024,chenPersonaPersonalizationSurvey2024} and spawned a growing ecosystem of commercial products\footnote{
  \hangindent=1.8em
  \url{https://character.ai/} \\
  \url{https://replika.ai/} \\
  \url{https://spicychat.ai/} \\
}. This motivates research on \textbf{general role-playing}: enabling an LLM to convincingly portray \emph{arbitrary} user-defined or fictional characters~\citep{wangCoSERCoordinatingLLMBased2025}, rather than a small pre-defined roster~\citep{yuNeekoLeveragingDynamic2024,yangHyCoRAHyperContrastiveRoleAdaptive2025}.

Despite this progress, existing general role-playing models often fall short in \textbf{immersion}. A large body of prior work focuses primarily on generating verbal \texttt{Speech} alone~\citep{shaoCharacterLLMTrainableAgent2023,chenLargeLanguageModels2023,liChatHaruhiRevivingAnime2023,wangRoleLLMBenchmarkingEliciting2024,zhou-etal-2024-characterglm,luLargeLanguageModels2024,zhouCharacterBenchBenchmarkingCharacter2024}. Subsequent efforts incorporate \texttt{Action}~\citep{tuCharacterEvalChineseBenchmark2024,heCrabNovelConfigurable2025} or additionally model \texttt{Thought} (e.g., CoSER)~\citep{wangCoSERCoordinatingLLMBased2025}. However, a crucial signal remains under-modeled: \textbf{dynamic environment information}. In narratives, environment is not mere decoration—it shapes atmosphere and causality, linking actions, world changes, and subsequent dialogue. This motivates \textbf{RQ1}: \textit{How can we design a more immersive message representation that tightly couples character behavior with evolving environments?}

Most role-playing systems assume a \textbf{static interaction structure}: a fixed scene and a fixed character interacting with the user. Even when multiple characters are involved~\citep{chenLargeLanguageModels2023,wangCoSERCoordinatingLLMBased2025}, existing frameworks seldom provide a \emph{systematic} way to choose the next speaker with a brief rationale, and rarely support narrative-level dynamics such as scene transitions or on-the-fly character introduction as the plot evolves. These limitations motivate \textbf{RQ2}: \textit{How can we design a role-playing framework that supports dynamic multi-character orchestration, scene transitions, and on-the-fly character introduction?}

To address RQ1, we propose an \textbf{immersive messaging configuration} that explicitly interleaves four elements—\texttt{[Thought]}, \texttt{(Action)}, \texttt{$<$Environment$>$}, and \texttt{Speech}—which may appear flexibly within a single turn. This formulation enhances situational grounding and facilitates narratives in which environmental states and their dynamics actively participate in the interaction loop.

To address RQ2, we formalize the notion of \textbf{adaptive role-playing}, where the system (i) dynamically selects the most appropriate next speaker among multiple roles (including the user), (ii) switches scenes as the plot progresses, and (iii) introduces new characters when required by the narrative. Based on this formulation, we present \textbf{AdaMARP} (An \underline{Ada}ptive \underline{M}ulti-\underline{A}gent Interaction Framework for General Immersive \underline{R}ole-\underline{P}laying), which models role-playing as the interaction of three agents: an \underline{Actor Model} that portrays all non-user roles, a \underline{User Model} (simulated by an LLM or replaced by a real user), and a \underline{Scene Manager} that performs high-level control. The Scene Manager operates over a discrete action space (\texttt{init\_scene}, \texttt{pick\_speaker}, \texttt{switch\_scene}, \texttt{add\_role}, \texttt{end}) and outputs explicit rationales to guide subsequent generation; in particular, it always begins an episode with a single \texttt{init\_scene} action that establishes the initial scene context.

To train models that can follow this adaptive framework, we construct two datasets. First, we build \textbf{AdaRPSet}~(\underline{Ada}ptive \underline{R}ole-\underline{P}laying Data\underline{set}) for training the Actor Model, consisting of (i) \textbf{AdaRPSet-Extracted}, extracted from narrative books with LLM-assisted scene extraction and profile synthesis, and (ii) \textbf{AdaRPSet-Synthesis}, an LLM-synthesized corpus explicitly covering dynamic phenomena such as scene switching and role addition across 20 topics. 
Second, to improve orchestration quality, we construct \textbf{AdaSMSet}~(\underline{Ada}ptive \underline{S}cene \underline{M}anager Data\underline{set}) based on the synthesized trajectories to supervise the Scene Manager. This dataset includes inserted \texttt{pick\_speaker} decisions and their rationales.

Finally, to evaluate the adaptive role-playing abilities of the model beyond sentence-level metrics, we propose \textbf{AdaptiveBench}, a simulation-based benchmark that generates full dialogue trajectories under Scene-Manager control and evaluates performance at the trajectory level. We assess the Actor Model with a rubric covering character consistency, environmental grounding, interpersonal interaction, narrative progression, and instruction compliance, and evaluate the Scene Manager in terms of scene understanding, speaker discipline, role-introduction judgment, and overall quality.

Experiments across multiple backbones and model scales demonstrate the effectiveness of our data and supervision strategy. \textbf{AdaRPSet} consistently improves the Actor Model in AdaMARP, yielding stronger character consistency, richer environment-grounded narration, and more coherent progression under dynamic scene and cast changes. Notably, an 8B Actor Model trained with AdaRPSet outperforms several proprietary role-playing LLMs (e.g., GPT-4o-mini, Gemini-2.5-Pro, and Doubao-1.5-Pro-Character) in our trajectory-level evaluation. 
\textbf{AdaSMSet} further enhances the Scene Manager's orchestration quality, allowing it to surpass Claude Sonnet 4.5 at the 14B scale. This is reflected in more appropriate scene transitions and better-justified role introductions.

Our main contributions are summarized as follows:
\begin{itemize}[leftmargin=1.2em, labelsep=0.4em,
                itemsep=0pt, parsep=0pt, topsep=1pt, partopsep=0pt]
    \item We propose \textbf{AdaMARP}, an adaptive framework that integrates an environment-aware message format (interleaving thought, action, environment, and speech) with a discrete-action Scene Manager to enable dynamic multi-character orchestration, supporting both scene transitions and on-the-fly role additions.
    \item We construct \textbf{AdaRPSet} and \textbf{AdaSMSet}, two large-scale datasets designed to supervise immersive role portrayal and high-level narrative control (e.g., scene transitions, role additions), respectively.
    \item We introduce \textbf{AdaptiveBench} for trajectory-level evaluation \textbf{in adaptive role-playing}. Experiments demonstrate that our 8B Actor Model outperforms GPT-4o-mini, while our 14B Scene Manager surpasses Claude Sonnet 4.5.
\end{itemize}

\section{Related Work}
\label{sec:related_work}

\paragraph{Character-Specific Role-Playing.}
Many approaches specialize an LLM to one or a few fixed characters, achieving strong fidelity but limited generalization to new roles (e.g., CharacterLLM~\citep{shaoCharacterLLMTrainableAgent2023}, Neeko~\citep{yuNeekoLeveragingDynamic2024}, HyCoRA~\citep{yangHyCoRAHyperContrastiveRoleAdaptive2025}).
Consequently, they are ill-suited for open-ended scenarios where the model must adopt arbitrary user-defined personas without per-character retraining.

\paragraph{General Role-Playing via Data Construction.}
General role-playing requires models to portray arbitrary user-defined or fictional characters. A prevalent line of work constructs large-scale multi-character corpora to improve persona coverage and avoid overfitting to a fixed roster, e.g., ChatHaruhi~\citep{liChatHaruhiRevivingAnime2023}, DITTO~\citep{chenLargeLanguageModels2023}, CharacterGLM~\citep{zhou-etal-2024-characterglm}, ROLEPERSONALITY~\citep{ranCapturingMindsNot2024}, SimChat~\citep{yangCraftingCustomisableCharacters2025}, BeyondDialogue~\citep{yuDialogueProfileDialogueAlignment2025}, Crab~\citep{heCrabNovelConfigurable2025}, TAILORGEN~\citep{gaoTailorRPARetrievalBasedFramework2025}, and CoSER~\citep{wangCoSERCoordinatingLLMBased2025}. While these efforts improve dialogue quality (sometimes incorporating thought/action), most do not explicitly model dynamic environments as first-class signals, nor provide systematic supervision for scene transitions and cast changes. For a systematic comparison, we summarize the key differences between existing datasets and ours in Table~\ref{tab:dataset_comparison}.


\paragraph{General Role-Playing via Training Strategies.}
Another line improves role-playing with specialized objectives or optimization, such as persona-aware contrastive learning (PCL)~\citep{jiEnhancingPersonaConsistency2025}, metacognition-driven training (R-CHAR)~\citep{qinRCHARMetacognitionDrivenFramework2025}, and RL variants that address reward noise or design implicit rewards (CPO~\citep{yeCPOAddressingReward2025}, CogDual~\citep{liuCogDualEnhancingDual2025}). Our work targets general role-playing with supervised fine-tuning, so we mainly compare against data-construction approaches; training-strategy methods are largely complementary.

\begin{table*}[t]
\centering
\scriptsize
\setlength{\tabcolsep}{3pt}
\renewcommand{\arraystretch}{0.95}
\begin{adjustbox}{width=\textwidth}
\begin{tabular}{l cccccc c cccccc cccc c}
\toprule
& \multicolumn{7}{c}{Character} & \multicolumn{6}{c}{Conversations} & \multicolumn{4}{c}{Message Components} & \\
\cmidrule(lr){2-8}\cmidrule(lr){9-14}\cmidrule(lr){15-18}
Dataset
& I\&A & P\&P & SS & AIA & SHC & PHA & \#Char
& Init. Scene & Multi-Char & Interloc. & Rel. & \#Dial. & Avg. Turn
& Speech & Action & Thought & Env. & Open \\
\midrule

ChatHaruhi~\citep{liChatHaruhiRevivingAnime2023} &
\cmark & \xmark & \xmark & \xmark & \xmark & \xmark & 32 &
\xmark & \xmark & \xmark & \xmark & 54{,}726 & Fixed 2 &
\cmark & \xmark & \xmark & \xmark & \cmark \\

CharacterLLM~\citep{shaoCharacterLLMTrainableAgent2023} &
\multicolumn{6}{c}{See Appendix~\ref{app:dataset_comparison_details}} & 9 &
\cmark & \xmark & \cmark & \cmark & 14{,}300 & 13.2 &
\cmark & \xmark & \pmark & \xmark & \cmark \\

HPD~\citep{chenLargeLanguageModels2023} &
\cmark & \cmark & \xmark & \cmark & \xmark & \xmark & 113 &
\cmark & \cmark & \cmark & \cmark & 1{,}042 & 13.8 &
\cmark & \xmark & \xmark & \xmark & \cmark \\

RoleLLM~\citep{wangRoleLLMBenchmarkingEliciting2024} &
\multicolumn{6}{c}{See Appendix~\ref{app:dataset_comparison_details}} & 100 &
\xmark & \xmark & \xmark & \xmark & 168{,}093 & Fixed 2 &
\cmark & \xmark & \xmark & \xmark & \cmark \\

CharacterGLM~\citep{zhou-etal-2024-characterglm} &
\cmark & \pmark & \xmark & \pmark & \cmark & \cmark & 250 &
\xmark & \xmark & \cmark & \cmark & 1{,}034 & 15.8 &
\cmark & \xmark & \xmark & \xmark & \cmark \\

DITTO~\citep{luLargeLanguageModels2024} &
\cmark & \xmark & \xmark & \cmark & \cmark & \cmark & 4{,}002 &
\xmark & \xmark & \xmark & \xmark & 7{,}186 & 5.10 &
\cmark & \xmark & \xmark & \xmark & \cmark \\

ROLEPERSONALITY~\citep{ranCapturingMindsNot2024} &
\multicolumn{6}{c}{See Appendix~\ref{app:dataset_comparison_details}} & 46 &
\xmark & \xmark & \xmark & \xmark & 32{,}089/32{,}767 & Fixed 2 or 5 &
\cmark & \xmark & \xmark & \xmark & \cmark \\

SimChat~\citep{yangCraftingCustomisableCharacters2025} &
\cmark & \cmark & \cmark & \cmark & \xmark & \cmark & 68 &
\cmark & \xmark & \cmark & \xmark & 13{,}971 & 10.3 &
\cmark & \xmark & \pmark & \xmark & \cmark \\

BeyondDialogue~\citep{yuDialogueProfileDialogueAlignment2025} &
\cmark & \cmark & \cmark & \xmark & \cmark & \xmark & 331 &
\cmark & \xmark & \cmark & \cmark & 3{,}552 & 6.54 &
\cmark & \xmark & \xmark & \xmark & \cmark \\

Crab~\citep{heCrabNovelConfigurable2025} &
\cmark & \cmark & \cmark & \cmark & \xmark & \cmark & 18{,}424 &
\cmark & \xmark & \cmark & \cmark & 41{,}631 & 4.96 &
\cmark & \cmark & \xmark & \xmark & \cmark \\

TAILORGEN~\citep{gaoTailorRPARetrievalBasedFramework2025} &
\multicolumn{6}{c}{See Appendix~\ref{app:dataset_comparison_details}} & 5 &
\xmark & \xmark & \xmark & \xmark & 7{,}671 & Fixed 2$^\ast$ &
\cmark & \xmark & \xmark & \xmark & \xmark \\

CoSER~\citep{wangCoSERCoordinatingLLMBased2025} &
\cmark & \cmark & \xmark & \cmark & \cmark & \cmark & 17{,}966 &
\cmark & \cmark & \cmark & \cmark & 29{,}798 & 13.20 &
\cmark & \cmark & \cmark & \pmark & \cmark \\

\textbf{Ours} &
\cmark & \cmark & \cmark & \cmark & \cmark & \cmark & 29{,}335 &
\cmark & \cmark & \cmark & \cmark & 22{,}425 & 20.08 &
\cmark & \cmark & \cmark & \cmark & \cmark \\
\bottomrule
\end{tabular}
\end{adjustbox}
\caption{Comparison of general role-playing datasets. Profile subfields are abbreviated as:
I\&A (Identity \& Appearance), P\&P (Personality \& Psychology), SS (Speaking Style),
AIA (Abilities, Interests \& Achievements), SHC (Social \& Historical Context), and PHA (Personal History Arc).
\#Char denotes the number of distinct characters in the dataset.
Init. Scene indicates whether dialogues are provided with an explicit initial scene or setting.
Interloc. indicates whether the user (or dialogue partner) is assigned a specific profile.
Rel. denotes whether explicit relationships among characters are modeled or annotated in dialogues.
\#Dial. refers to the total number of dialogues.
Env. is short for Environment, indicating whether environmental descriptions are included.
Open indicates whether the dataset is open-sourced.
\pmark\ and $^\ast$ denote partial/implicit support or values inferred from the reported pipeline; detailed clarifications on symbol interpretations and dataset releases are provided in Appendix~\ref{app:dataset_comparison_details}.}
\label{tab:dataset_comparison}
\end{table*}

\section{Method}
\label{sec:method}

\subsection{Design Principles of AdaMARP}
\label{subsec:design_principles}


\subsubsection{Comprehensive Character Information}
\label{subsubsec:character_info}
To support nuanced and consistent role-playing, AdaMARP adopts a structured profile for the main character that covers seven complementary dimensions: (I) Identity \& Appearance, (II) Personality \& Psychology, (III) Speaking Style, (IV) Abilities, Interests \& Achievements, (V) Social \& Historical Context, (VI) Personal History Arc, and (VII) Relationships. Together, these dimensions provide both intrinsic traits and narrative/social grounding, enabling coherent behavior, stable interaction patterns, and evolution-aware characterization across scene changes and long trajectories. 
Detailed specifications are provided in Appendix~\ref{app:sec:character_profile_details}.

\subsubsection{Immersive Messaging Configuration}
\label{subsubsec:immersive_message}
AdaMARP adopts a structured message representation that extends prior role-playing formats beyond plain dialogue~\citep{shaoCharacterLLMTrainableAgent2023,chenLargeLanguageModels2023,liChatHaruhiRevivingAnime2023,wangRoleLLMBenchmarkingEliciting2024,zhou-etal-2024-characterglm,luLargeLanguageModels2024,zhouCharacterBenchBenchmarkingCharacter2024} and action-augmented interaction~\citep{tuCharacterEvalChineseBenchmark2024,heCrabNovelConfigurable2025}. Following recent practice of explicitly modeling internal states (e.g., reasoning or thoughts)~\citep{wangCoSERCoordinatingLLMBased2025,yangCraftingCustomisableCharacters2025}, we further introduce \textbf{environment-aware} descriptions as a first-class component of each turn.

Concretely, environmental signals play two complementary roles in our setting.
First, they provide \emph{atmospheric grounding}, using environmental cues to evoke or amplify character emotions.
Second, they enable \emph{interaction dynamics} between characters and their surroundings, where character actions update the environment state and external events (e.g., a sound behind a door) influence subsequent decisions and dialogue.
To operationalize this design, we adopt a unified message format that interleaves \texttt{Thought}, \texttt{Action}, \texttt{Environment}, and \texttt{Speech}: internal thoughts are enclosed in square brackets \texttt{[]}, physical actions in parentheses \texttt{()}, environment states or changes in angle brackets \texttt{<>}, and unmarked text denotes \texttt{Speech}.


\begin{figure*}[t]
    \centering
    \includegraphics[width=\textwidth]{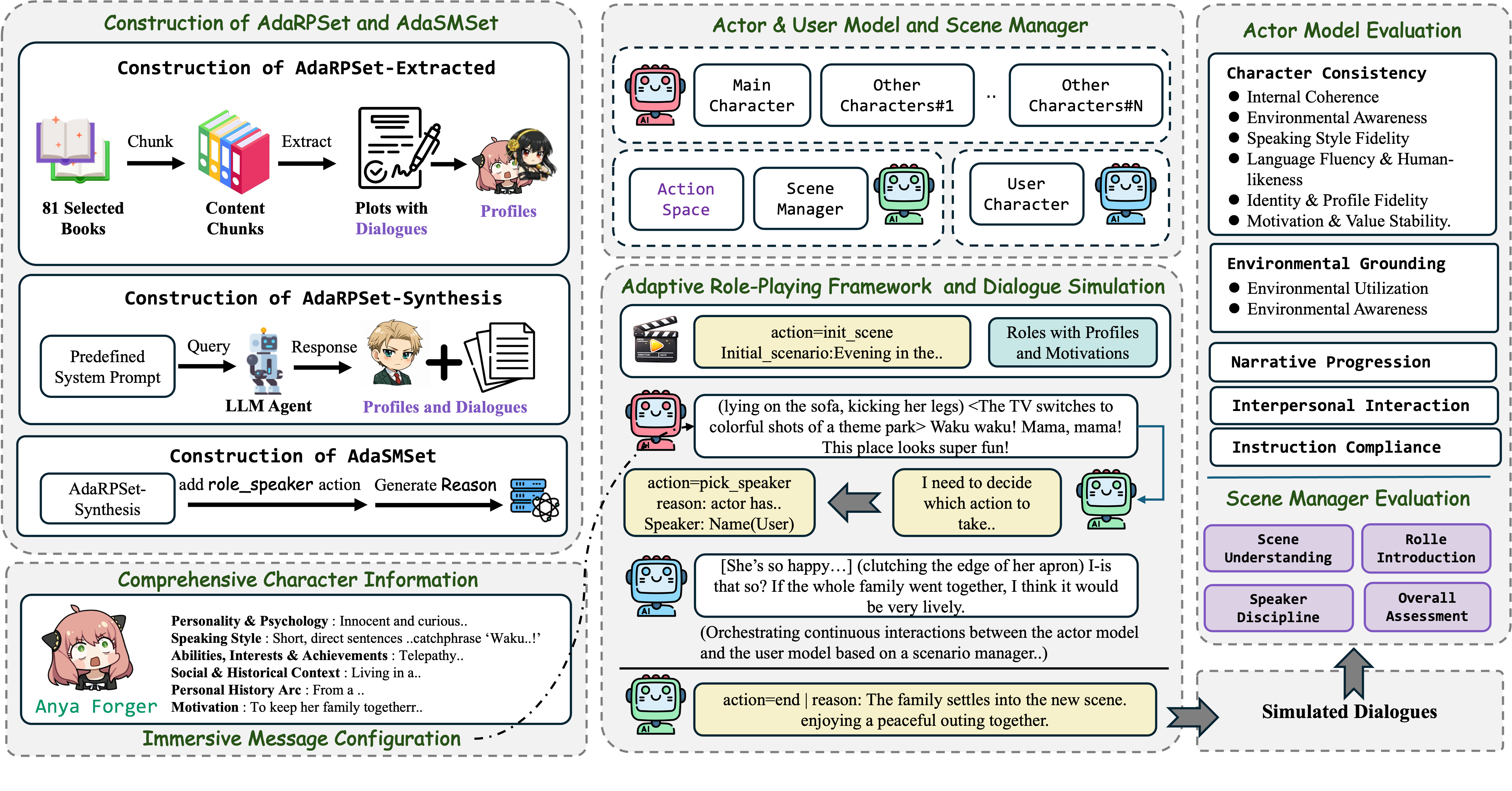}
    \caption{Overall framework of \textbf{AdaMARP}.
    The left part illustrates the construction of AdaRPSet and AdaSMSet and the components of a comprehensive role profile.
    The upper middle part shows the roles played by the three agents, while the lower middle part depicts the dialogue trajectory generation under the orchestration of the scene manager, given predefined roles and an initial scenario. The right part presents the evaluation of the actor models and the scene manager based on the generated trajectories.}
    \label{fig:main_framework}
\end{figure*}

\subsubsection{Adaptive Role-Playing Framework}
\label{subsubsec:adaptive_framework}
We propose the \textbf{AdaMARP framework} to support role-playing episodes in which the active speaker, the scene, and even the cast may change over time. The framework models three interacting agents. The \textbf{Actor Model} $\mathcal{A}$ generates messages for all non-user characters, including the main character and any auxiliary roles. The \textbf{User Model} $\mathcal{U}$ represents the user side of the interaction and is simulated by an LLM by default, while remaining replaceable by a human user. The \textbf{Scene Manager} $\mathcal{S}$ performs high-level orchestration by tracking dialogue history and role/scene states.

Each role (including the user) is associated with a structured profile and a \emph{scene-specific motivation}, which conditions behavior in the current context. At each step, the Scene Manager selects an action from a discrete control space $\mathcal{M}=\{\texttt{init\_scene},\allowbreak \texttt{pick\_speaker},\allowbreak \texttt{switch\_scene},\allowbreak \texttt{add\_role},\allowbreak \texttt{end}\}$. The first action issued in an episode is always \texttt{init\_scene}, which provides the initial scene description and bootstraps the interaction history. For every action $m\in\mathcal{M}$, $\mathcal{S}$ outputs an explicit rationale. When $m=\texttt{switch\_scene}$, $\mathcal{S}$ additionally produces an updated scene description; when $m=\texttt{add\_role}$, it specifies the new role’s name, profile, and motivation; when $m=\texttt{pick\_speaker}$, it designates which role (including the user) should produce the next in-character message. The complete pseudo-code of the AdaMARP framework is presented in Algorithm~\ref{alg:adaptive_marps}, while additional implementation details and prompt templates for $\mathcal{A}$, $\mathcal{U}$, and $\mathcal{S}$ are provided in Appendix~\ref{app:sec:details_of_adaptive_role-playing_framework}.

\begin{algorithm}[t]
\caption{Adaptive Role-Playing Framework}
\label{alg:adaptive_marps}
\KwIn{Initial role set $\mathcal{R}_0$, role profiles $\{\mathcal{P}_i\}$, role motivations $\{\Omega_i\}$}
\KwOut{Role-Playing interaction trajectory $\mathcal{H}$}
\textbf{Agents:} Scene Manager $\mathcal{S}$, Actor Model $\mathcal{A}$, User Model $\mathcal{U}$\;
Initialize dialogue history $\mathcal{H}_0 \leftarrow \emptyset$\;
$t \leftarrow 0$\;
$\mathcal{S}$ outputs $m_0 = \texttt{init\_scene}$ with initial scene description\;
Append action output to $\mathcal{H}_1$\;
$t \leftarrow 1$\;
\While{True}{
    Construct interaction state $\mathcal{G}_t = (\mathcal{R}_t, \mathcal{H}_t)$\;
    $\mathcal{S}$ selects action $m_t \in \mathcal{M}$ with rationale\;
    \If{$m_t = \texttt{pick\_speaker}$}{
        Select role $r_k \in \mathcal{R}_t$\;
        Generate in-character response using $\mathcal{A}$ or $\mathcal{U}$ conditioned on
        $\mathcal{P}_k$, $\Omega_k$, $\{\mathcal{P}_j\}_{j \neq k}$, and $\mathcal{H}_t$\;
        Append response to $\mathcal{H}_{t+1}$\;
    }
    \ElseIf{$m_t = \texttt{switch\_scene}$}{
        Append new scene description to $\mathcal{H}_{t+1}$\;
    }
    \ElseIf{$m_t = \texttt{add\_role}$}{
        Instantiate new role $r_{n_t+1}$ with profile $\mathcal{P}_{n_t+1}$ and motivation $\Omega_{n_t+1}$\;
        $\mathcal{R}_{t+1} \leftarrow \mathcal{R}_t \cup \{r_{n_t+1}\}$\;
        Append action output to $\mathcal{H}_{t+1}$\;
    }
    \ElseIf{$m_t = \texttt{end}$}{
        \textbf{break}\;
    }
    $t \leftarrow t + 1$\;
}
\Return $\mathcal{H}_t$\;
\end{algorithm}

\subsection{Dataset Curation}
\label{subsec:dataset_curation}
Grounded in the above design principles, we curate two datasets to support the AdaMARP framework: \textbf{AdaRPSet} for training the Actor Model (Section~\ref{subsubsec:dataset_for_actor_model}) and \textbf{AdaSMSet} for supervising the Scene Manager (Section~\ref{subsubsec:dataset_for_scene_manager}). Section~\ref{subsec:evaluation_framework} further presents an evaluation framework tailored to assess both components within AdaMARP.

\subsubsection{Dataset for Actor Model}
\label{subsubsec:dataset_for_actor_model}

To satisfy the immersive messaging principle in Section~\ref{subsubsec:immersive_message}, we construct \textbf{AdaRPSet}, a two-part corpus: \textbf{AdaRPSet-Extracted}, grounded in literary texts to teach the unified message format and more human-like behaviors, and \textbf{AdaRPSet-Synthesis}, designed to cover dynamic demands that are scarce in extracted data, such as scene transitions and on-the-fly character introduction.

\noindent \textbf{AdaRPSet-Extracted.}
Following CoSER~\citep{wangCoSERCoordinatingLLMBased2025}, we select 81 representative books from Goodreads’ Best Books Ever list (Table~\ref{tab:selected_books}) and obtain their full texts\footnote{\url{https://www.goodreads.com/list/show/1.Best_Books_Ever}}. We build AdaRPSet-Extracted via three stages: \textbf{Chunking}, \textbf{LLM-based Extraction}, and \textbf{LLM-based Profile Generation}. We segment each book by chapters when available and merge adjacent segments into chunks under a size budget. We then prompt a strong LLM to identify coherent scenes and extract multi-character interaction trajectories (without restricting to two speakers~\citep{yuDialogueProfileDialogueAlignment2025}), emitting each trajectory \emph{directly} in our unified messaging format (including \textit{thought}, \textit{action}, and \textit{environment} when applicable), grounded in the source text. Finally, we aggregate scenes per character and prompt the LLM to synthesize a seven-dimensional profile (Section~\ref{subsubsec:character_info}). Prompts and pipeline details are provided in Appendix~\ref{app:subsec:adarpset_extracted}.

\noindent \textbf{AdaRPSet-Synthesis.}
While AdaRPSet-Extracted supports well-formed messages conditioned on profiles and scenes, it is less robust to \emph{dynamic} role-playing (e.g., scene switching and character insertion) and offers limited interaction diversity. We therefore construct AdaRPSet-Synthesis by prompting an advanced LLM to generate plot-level trajectories. Each trajectory specifies an initial scenario, a main character plus additional characters (each with a profile and initial motivation), and a multi-turn interaction in our unified format. It also includes \emph{scene manager} control messages (e.g., \texttt{Action=add\_role}, \texttt{Action=switch\_scene}) to explicitly model role addition and scene transitions; we exclude \texttt{pick\_speaker} to avoid turn-selection artifacts. We enforce at least one scene switch and one role addition per trajectory. We instantiate 20 themes with 50 trajectories each, using 45/5 for train/test (AdaptiveBench). The complete synthesis prompts and configurations are detailed in Appendix~\ref{app:subsec:adarpset_synthesis}, after which the resulting trajectories are standardized into a unified training-sample format as described in Appendix~\ref{app:subsec:training_sample_format}.

\subsubsection{Dataset for Scene Manager}
\label{subsubsec:dataset_for_scene_manager}
To explicitly supervise high-level control in AdaMARP, we construct a dedicated dataset for the Scene Manager, \textbf{AdaSMSet}. We build it on top of \textbf{AdaRPSet-Synthesis}, whose trajectories already include scene-manager actions for \texttt{add\_role}, \texttt{switch\_scene}, and \texttt{end}. We then add the remaining key supervision for $\mathcal{S}$—\textbf{speaker selection}—by inserting \texttt{pick\_speaker} messages between consecutive character turns and prompting a strong instruction-following language model to generate the corresponding \texttt{Reason} field. Construction details and prompts are provided in Appendix~\ref{app:subsec:adasmset}.

\ding{117} We provide detailed statistical analyses of both AdaRPSet and AdaSMSet in Appendix~\ref{app:sec:statistical_analysis_adarpset}.

\subsection{Actor Model and Scene Manager Training}
\label{subsec:training_actor_scene_manager}

We train AdaMARP using supervised fine-tuning on two specialized corpora. The Actor Model $\mathcal{A}$ is fine-tuned on \textbf{AdaRPSet} (comprising both Extracted and Synthesis subsets), where each training sample follows the unified role-playing message format detailed in Table~\ref{tab:training_sample_formatter}. The Scene Manager $\mathcal{S}$ is trained on \textbf{AdaSMSet}, which provides explicit supervision for control decisions. Specifically, $\mathcal{S}$ learns to predict the next action from the set $\mathcal{M}$ (e.g., \texttt{pick\_speaker}, \texttt{switch\_scene}, \texttt{add\_role}, \texttt{end}) along with its requisite arguments—such as \texttt{reason}, \texttt{new\_scene}, or \texttt{new\_role\_*}—conditioned on the interaction history~(See Table~\ref{tab:training_scene_manager}). 

\subsection{AdaptiveBench: Evaluation Framework}
\label{subsec:evaluation_framework}

\subsubsection{Dialogue Simulation}
\label{subsubsec:dialogue_simulation}
We introduce \textbf{AdaptiveBench}, a simulation framework designed to evaluate both the Actor Model and the Scene Manager by generating full interaction trajectories. We derive 100 evaluation seeds from the held-out split of AdaRPSet-Synthesis (5 instances per topic across 20 topics), preserving the initial role configurations and scene descriptions. For each seed, we simulate an episode involving three agents: an LLM-based user simulator $U$, the Actor Model $\mathcal{A}$ portraying all non-user roles, and the Scene Manager $\mathcal{S}$ issuing control actions (e.g., \texttt{pick\_speaker}, \texttt{switch\_scene}). At every step, $\mathcal{S}$ orchestrates the flow, and the designated speaker (either $U$ or $\mathcal{A}$) generates the response. Each simulation runs for a fixed horizon of $T=20$ dialogue turns (excluding $\mathcal{S}$ messages), yielding 100 complete trajectories for subsequent evaluation.

\subsubsection{Trajectory Evaluation}
\label{subsubsec:trajectory_evaluation}
Since AdaMARP introduces a structured message format (incorporating environmental information) and dynamic states (e.g., scene transitions, role additions), existing role-playing evaluation frameworks~\citep{brahmanLetYourCharacters2021,zhouCharacterGLMCustomizingSocial2024,yuanEvaluatingCharacterUnderstanding2024,tuCharacterEvalChineseBenchmark2024,ahnTimeCharaEvaluatingPointinTime2024,samuelPersonaGymEvaluatingPersona2025,zhangRolePlotSystematicFramework2025,luRoleMRCFineGrainedComposite2025} are insufficient to capture these dimensions. We therefore adopt a \emph{trajectory-level} protocol tailored for adaptive role-playing in AdaptiveBench, evaluating both the Actor Model and the Scene Manager on complete simulated dialogues.

\paragraph{Actor Model evaluation.}
For each simulated trajectory $\tau=\{m_t\}_{t=1}^{T}$, we provide an LLM-as-Judge with the role set $\mathcal{R}$ (profiles), motivations $\Omega$, the initial scene $\mathcal{E}_0$, and the full dialogue trajectory $\tau$. The judge assigns independent $0$--$10$ scores to each sub-metric under five dimensions: (I) \textbf{Character Consistency}: Internal Coherence, Speaking Style Fidelity, Language Fluency \& Human-likeness, Identity \& Profile Fidelity, Motivation \& Value Stability; (II) \textbf{Environmental Grounding}: Env Awareness, Env Utilization; (III) \textbf{Interpersonal Interaction}: Contextual Responsiveness, Relationship Awareness; (IV) \textbf{Narrative Progression}: Attractiveness, Stability; (V) \textbf{Instruction Compliance}: Compliance. We report all sub-metric scores directly; the complete rubric and judge prompts are provided in Appendix~\ref{app:subsec:actor_model_evaluation}.

\paragraph{Scene Manager evaluation.}
We evaluate the Scene Manager with the same inputs as Actor evaluation (i.e., $\mathcal{R}$, $\mathcal{M}$, $\mathcal{E}_0$, and the full trajectory $\tau$), and ask an LLM to output four independent $0$--$10$ scores: (I) \textbf{Scene Understanding}; (II) \textbf{Speaker Discipline}; (III) \textbf{Role Introduction Judgment}; (IV) \textbf{Overall Assessment}. The complete scoring rubric and judge prompts are provided in Appendix~\ref{app:subsec:scene_manager_evaluation}. 

\ding{117} To complement the above dataset construction and evaluation framework, we provide a detailed analysis of token consumption across both data curation and AdaptiveBench evaluation in Appendix~\ref{app:sec:token_consumption_analysis}.

\section{Experiment}
\label{sec:experiment}

\subsection{Experimental Settings}
\label{subsec:experimental_settings}

\noindent \textbf{Models.}
We evaluate a diverse set of proprietary and open-source large language models. Our proprietary models include GPT-4o-mini~\citep{openai2024gpt4technicalreport}, GPT-5-Chat~\citep{openai2025gpt5systemcard}, Gemini-2.5-Pro~\citep{comanici2025gemini25pushingfrontier}, Claude Sonnet 4.5~\citep{anthropic2024claude4}, and Doubao-1.5-Pro-Character~\citep{bytedance2024doubao}. For open-source baselines, we consider three categories.
(i) \emph{Roleplay-oriented models}: Index-1.9B-Character~\footnote{\url{https://huggingface.co/IndexTeam/Index-1.9B-Character}}.
(ii) \emph{Instruction-tuned models}: Qwen2.5-7B/14B/72B-Instruct~\citep{qwen2025qwen25technicalreport} and Llama-3.1-70B-Instruct~\citep{grattafiori2024llama3herdmodels}.
(iii) \emph{Reasoning-focused models}: Qwen3-14B and QwQ-32B~\citep{yang2025qwen3technicalreport}. To explicitly examine whether role-playing fine-tuning can effectively steer \emph{base} models, we additionally include Llama-3.1-8/70B ~\cite{grattafiori2024llama3herdmodels} as controlled baselines.

\noindent \textbf{Datasets and Training.}
We fine-tune the Actor Model on AdaRPSet and the Scene Manager on AdaSMSet, both curated in Section~\ref{subsec:dataset_curation}. All models are trained with full fine-tuning; detailed configurations and hyperparameters are provided in Appendix~\ref{app:sec:training_details}.

\noindent \textbf{Baselines.}
We focus on \emph{general-purpose} role-playing and thus exclude character-specific methods (e.g., Neeko and HyCoRA) and single-turn settings (e.g., RoleLLM). We compare with three recent open-source baselines: BeyondDialogue~(Beyond), Crab, and CoSER, which model increasingly rich signals (speech $\rightarrow$ action $\rightarrow$ thought); CoSER is the closest due to its multi-role support. 
We design two sets of inference-time system prompts—\texttt{Basic} and \texttt{Enhance}—for both the Actor Model and the Scene Manager (See Appendix~\ref{app:sec:prompt_templates}). Unless otherwise specified, we report results using the \texttt{Enhance} prompt for both components.

\noindent \textbf{Evaluation Metrics.}
For the Actor Model, we primarily use AdaptiveBench to evaluate all models (open-source, proprietary, and baselines) on the five dimensions and twelve sub-metrics defined in Section~\ref{subsubsec:trajectory_evaluation}. We further report results on two public role-playing evaluation methods, CharacterArena~\citep{yeCPOAddressingReward2025} and CharacterBench~\citep{zhouCharacterBenchBenchmarkingCharacter2024}, to examine performance under different prompts and metric designs. For the Scene Manager, since it is newly introduced in this work, we evaluate it only with AdaptiveBench. Details of CharacterArena and CharacterBench are provided in Appendix~\ref{app:sec:additional_evaluation_frameworks}. For AdaptiveBench, we report the mean score and standard deviation for each metric, 
where the standard deviation is computed over all evaluation samples within the benchmark.

\subsection{Main Results}
\label{subsec:main_results}

\subsubsection{Actor Model}
Following the experimental setup described above, we evaluate all Actor Models (including open-/closed-source LLMs and re-trained baselines) on AdaptiveBench with a unified configuration: the User Model is simulated by Doubao-1.5-Pro-Character and the Manager Model is simulated by GPT-4o-mini, both equipped with our \texttt{Enhance} system prompts. We then use GPT-5-Chat as the judge model for trajectory-level scoring (we ablate different judge models in Appendix~\ref{app:actor_eval_ablation}). The results are shown in Table~\ref{tab:actor_model_adaptivebench_result_main}. Since a score of 5 denotes a neutral/acceptable level (see Table~\ref{tab:actor_eval_prompt_part1}\textasciitilde~\ref{tab:actor_eval_prompt_part2}), we draw the following observations.

\begin{table*}[t]
\centering
\resizebox{\textwidth}{!}{
\begin{tabular}{l c c c c c c c c c c c c c}
\hline
\multicolumn{14}{l}{\makecell[l]{
\textbb{ICoh}: Internal Coherence ~~~
\textbb{SSF}: Speaking Style Fidelity ~~~
\textbb{LFH}: Language Fluency \& Human-likeness ~~~
\textbb{IPF}: Identity \& Profile Fidelity ~~~
\textbb{MVS}: Motivation \& Value Stability \\
\textbb{EA}: Environmental Awareness ~~~
\textbb{EU}: Environmental Utilization ~~~
\textbb{CR}: Contextual Responsiveness ~~~
\textbb{RA}: Relationship Awareness ~~~
\textbb{ATT}: Attractiveness ~~~
\textbb{STB}: Stability ~~~
\textbb{IC}: Instruction Compliance \\
\textbb{CC}: Character Consistency ~~~
\textbb{EG}: Environmental Grounding ~~~
\textbb{II}: Interpersonal Interaction ~~~
\textbb{NP}: Narrative Progression
}} \\
\hline
\multirow{2}{*}{\textbb{Models}}
& \multicolumn{5}{c}{\textbb{CC}}
& \multicolumn{2}{c}{\textbb{EG}}
& \multicolumn{2}{c}{\textbb{II}}
& \multicolumn{2}{c}{\textbb{NP}}
& \multicolumn{1}{c}{\textbb{IC}}
& \multirow{2}{*}{\textbb{Average}} \\
\cmidrule(lr){2-6}
\cmidrule(lr){7-8}
\cmidrule(lr){9-10}
\cmidrule(lr){11-12}
\cmidrule(lr){13-13}
& \textbb{ICoh} & \textbb{SSF} & \textbb{LFH} & \textbb{IPF} & \textbb{MVS}
& \textbb{EA} & \textbb{EU}
& \textbb{CR} & \textbb{RA}
& \textbb{ATT} & \textbb{STB}
& \textbb{IC}
&  \\
\hline

\multicolumn{14}{l}{\textbf{Close-source LLMs}} \\
\hline
GPT-4o-mini
& 8.99$\pm$0.10 & 8.19$\pm$0.39 & 8.89$\pm$0.31 & 9.50$\pm$0.50 & 8.88$\pm$0.57
& 8.71$\pm$0.48 & 7.83$\pm$0.45
& 9.20$\pm$0.40 & 8.15$\pm$0.36
& 8.17$\pm$0.43 & 9.24$\pm$0.43
& 9.64$\pm$0.56
& 8.78 \\

GPT-5-Chat
& \underline{9.09$\pm$0.99} & \underline{8.84$\pm$0.92} & \underline{8.94$\pm$0.95} & \underline{9.85$\pm$1.01} & \underline{9.03$\pm$1.04}
& \underline{9.03$\pm$1.03} & \underline{8.39$\pm$1.05}
& \underline{9.43$\pm$1.07} & \underline{8.52$\pm$1.00}
& \underline{8.90$\pm$0.97} & \underline{9.47$\pm$1.08}
& \textbf{9.70$\pm$0.46}
& \underline{9.10} \\

Gemini-2.5-Pro
& 7.97$\pm$0.41 & 7.57$\pm$0.60 & 7.57$\pm$0.55 & 8.92$\pm$0.46 & 8.02$\pm$0.53
& 8.04$\pm$0.62 & 7.17$\pm$0.68
& 8.30$\pm$0.57 & 7.87$\pm$0.72
& 8.27$\pm$0.58 & 7.88$\pm$0.70
& 9.22$\pm$0.50
& 8.07 \\

Claude Sonnet 4.5
& \textbf{9.35}$\pm$0.54 & \textbf{8.97}$\pm$0.22 & \textbf{8.99}$\pm$0.41 & \textbf{9.94}$\pm$0.28 & \textbf{9.27}$\pm$0.66
& \textbf{9.13}$\pm$0.58 & \textbf{8.39}$\pm$0.63
& \textbf{9.78}$\pm$0.44 & \textbf{8.90}$\pm$0.48
& \textbf{9.23}$\pm$0.53 & \textbf{9.65}$\pm$0.65
& \underline{9.61$\pm$0.49}
& \textbf{9.27} \\

Doubao-1-5-Pro-Character
& 8.65$\pm$0.48 & 7.82$\pm$0.48 & 8.74$\pm$0.46 & 9.17$\pm$0.53 & 8.42$\pm$0.71
& 8.21$\pm$0.65 & 7.26$\pm$0.67
& 8.91$\pm$0.40 & 7.86$\pm$0.53
& 7.91$\pm$0.49 & 8.86$\pm$0.55
& 9.28$\pm$0.71
& 8.42 \\

\hdashline

\multicolumn{14}{l}{\textbf{Open-source LLMs}} \\
\hline
\multicolumn{14}{l}{\textbf{$<$14B}} \\
\hline

Index-1.9B-Character
& 5.62$\pm$1.46 & 4.63$\pm$1.35 & 6.81$\pm$1.08 & 6.15$\pm$1.59 & 6.49$\pm$1.39
& 6.17$\pm$1.28 & 4.82$\pm$1.32
& 6.84$\pm$1.35 & 5.43$\pm$1.36
& 5.13$\pm$1.27 & 6.69$\pm$1.41
& 5.47$\pm$2.61
& 5.85 \\

Qwen2-7B-Beyond-RP*
& 5.32$\pm$1.99 & 4.77$\pm$2.13 & 6.53$\pm$1.81 & 6.05$\pm$2.38 & 6.13$\pm$2.05
& 5.75$\pm$1.66 & 4.54$\pm$1.68
& 6.39$\pm$2.09 & 5.34$\pm$1.92
& 5.07$\pm$1.94 & 6.29$\pm$1.96
& 5.12$\pm$2.81
& 5.61 \\

Qwen2.5-7B-Instruct
& 8.61$\pm$0.99 & 7.70$\pm$0.91 & 8.70$\pm$0.96 & 9.03$\pm$1.02 & 8.35$\pm$1.06
& 8.24$\pm$1.06 & 7.33$\pm$0.95
& 8.81$\pm$0.97 & 7.80$\pm$0.89
& 7.73$\pm$0.96 & 8.89$\pm$0.97
& 9.23$\pm$0.72
& 8.37 \\

Qwen2.5-7B-Instruct-Beyond
& 6.38$\pm$1.75 & 6.02$\pm$1.71 & 7.42$\pm$1.21 & 7.52$\pm$1.73 & 7.10$\pm$1.45
& 6.07$\pm$1.39 & 4.78$\pm$1.34
& 7.48$\pm$1.41 & 6.56$\pm$1.44
& 5.91$\pm$1.45 & 7.37$\pm$1.47
& 6.71$\pm$2.95
& 6.61 \\

Qwen2.5-7B-Instruct-Crab
& 6.68$\pm$1.28 & 5.70$\pm$1.29 & 7.50$\pm$0.94 & 7.45$\pm$1.42 & 7.34$\pm$1.18
& 6.95$\pm$1.13 & 5.66$\pm$1.16
& 7.45$\pm$1.37 & 6.39$\pm$1.18
& 6.04$\pm$1.45 & 7.62$\pm$1.08
& 7.72$\pm$1.79
& 6.88 \\

Qwen2.5-7B-Instruct-CoSER
& 8.37$\pm$0.66 & 7.52$\pm$0.69 & 8.51$\pm$0.58 & 8.97$\pm$0.67 & 8.37$\pm$0.76
& 7.57$\pm$0.79 & 6.51$\pm$0.83
& 8.69$\pm$0.54 & 7.78$\pm$0.52
& 7.51$\pm$0.72 & 8.69$\pm$0.63
& 9.33$\pm$0.55
& 8.15 \\

\textbf{Qwen2.5-7B-Instruct-Ours}
& \underline{8.86$\pm$0.92} & \underline{8.28$\pm$0.97} & \underline{8.83$\pm$0.94} & \underline{9.51$\pm$1.08} & \underline{8.81$\pm$1.07}
& \underline{8.54$\pm$1.01} & \underline{7.65$\pm$0.90}
& \underline{8.99$\pm$0.94} & \underline{8.11$\pm$0.94}
& \underline{8.38$\pm$0.98} & \underline{9.04$\pm$0.99}
& \textbf{9.70$\pm$0.46}
& \underline{8.72} \\

Llama-3.1-8B-Crab
& 7.69$\pm$0.90 & 6.91$\pm$0.90 & 8.07$\pm$0.72 & 8.48$\pm$0.94 & 8.24$\pm$0.92
& 7.43$\pm$0.80 & 6.34$\pm$0.85
& 8.50$\pm$0.84 & 7.42$\pm$0.84
& 7.11$\pm$0.87 & 8.45$\pm$0.95
& 7.54$\pm$1.98
& 7.68 \\

Llama-3.1-8B-CoSER*
& 8.66$\pm$0.61 & 7.71$\pm$0.61 & 8.72$\pm$0.45 & 9.06$\pm$0.61 & 8.68$\pm$0.79
& 7.72$\pm$0.78 & 6.71$\pm$0.85
& 9.00$\pm$0.56 & 7.98$\pm$0.59
& 7.51$\pm$0.76 & 8.97$\pm$0.51
& 9.16$\pm$1.06
& 8.32 \\

\textbf{Llama-3.1-8B-Ours}
& \textbf{9.00$\pm$0.00} & \textbf{8.39$\pm$0.49} & \textbf{8.93$\pm$0.26} & \textbf{9.69$\pm$0.48} & \textbf{9.08$\pm$0.63}
& \textbf{8.77$\pm$0.49} & \textbf{7.84$\pm$0.44}
& \textbf{9.26$\pm$0.46} & \textbf{8.36$\pm$0.52}
& \textbf{8.52$\pm$0.50} & \textbf{9.39$\pm$0.49}
& \underline{9.49$\pm$0.54}
& \textbf{8.89} \\

\hdashline
\multicolumn{14}{l}{\textbf{14--72B}} \\
\hline

Qwen2.5-14B-Instruct
& 8.90$\pm$0.71 & 7.99$\pm$0.87 & 8.90$\pm$0.33 & 9.32$\pm$1.06 & 8.74$\pm$1.07
& 8.75$\pm$0.52 & 7.78$\pm$0.52
& 9.03$\pm$0.36 & 8.04$\pm$0.60
& 8.17$\pm$0.49 & 9.17$\pm$0.58
& 9.57$\pm$0.90
& 8.70 \\

\textbf{Qwen2.5-14B-Instruct-Ours}
& \underline{8.98$\pm$0.14} & 8.34$\pm$0.47 & \underline{8.96$\pm$0.20} & 9.66$\pm$0.47 & 8.94$\pm$0.56
& 8.69$\pm$0.46 & 7.77$\pm$0.42
& 9.10$\pm$0.30 & 8.18$\pm$0.43
& 8.61$\pm$0.49 & 9.11$\pm$0.37
& \underline{9.70$\pm$0.46}
& 8.84 \\

Qwen2.5-72B-Instruct
& 8.93$\pm$0.26 & 8.10$\pm$0.33 & \underline{8.96$\pm$0.20} & 9.40$\pm$0.53 & 8.72$\pm$0.66
& 8.61$\pm$0.55 & 7.66$\pm$0.51
& 9.13$\pm$0.39 & 8.14$\pm$0.42
& 8.05$\pm$0.46 & 9.17$\pm$0.45
& \textbf{9.71$\pm$0.45}
& 8.71 \\

\textbf{Qwen2.5-72B-Instruct-Ours}
& 8.95$\pm$0.22 & 8.42$\pm$0.49 & 8.86$\pm$0.34 & \underline{9.67$\pm$0.47} & 8.79$\pm$0.56
& 8.55$\pm$0.52 & 7.71$\pm$0.45
& 9.14$\pm$0.37 & 8.26$\pm$0.51
& 8.62$\pm$0.48 & 9.09$\pm$0.41
& 9.64$\pm$0.48
& 8.81 \\

Llama-3.1-70B-Instruct
& 8.91$\pm$0.90 & 8.08$\pm$0.89 & 8.88$\pm$0.91 & 9.36$\pm$1.06 & 8.79$\pm$1.08
& 8.39$\pm$1.03 & 7.58$\pm$1.01
& 9.07$\pm$0.98 & 8.14$\pm$0.93
& 8.39$\pm$0.98 & 9.10$\pm$0.99
& 9.49$\pm$1.09
& 8.68 \\

Llama-3.1-70B-CoSER*
& 8.90$\pm$0.30 & 7.98$\pm$0.35 & 8.88$\pm$0.32 & 9.41$\pm$0.53 & 8.77$\pm$0.65
& 8.14$\pm$0.65 & 7.19$\pm$0.67
& 9.09$\pm$0.32 & 8.14$\pm$0.40
& 8.07$\pm$0.43 & 9.08$\pm$0.37
& 9.65$\pm$0.48
& 8.61 \\

\textbf{Llama-3.1-70B-Ours }
& 8.90$\pm$0.71 & 8.14$\pm$0.82 & 8.83$\pm$0.75 & 9.42$\pm$0.84 & \underline{8.99$\pm$1.11} & 8.55$\pm$1.00 & 7.63$\pm$0.93 & 9.05$\pm$0.89 & 8.24$\pm$0.86 & 8.37$\pm$0.99 & 9.21$\pm$0.78 & 9.05$\pm$0.90 & 8.70 \\

Qwen3-14B
& 8.91$\pm$1.32 & \underline{8.64$\pm$1.31} & 8.67$\pm$1.31 & \underline{9.67$\pm$1.45} & 8.91$\pm$1.42
& \underline{8.85$\pm$1.36} & \underline{8.19$\pm$1.33}
& 9.05$\pm$1.42 & 8.22$\pm$1.31
& \underline{8.73$\pm$1.34} & 9.25$\pm$1.45
& 9.43$\pm$1.10
& \underline{8.88} \\

QwQ-32B
& \textbf{9.19}$\pm$0.39 & \textbf{8.99}$\pm$0.22 & \textbf{8.89}$\pm$0.34 & \textbf{9.96}$\pm$0.20 & \textbf{9.11}$\pm$0.51
& \textbf{9.15}$\pm$0.50 & \textbf{8.72}$\pm$0.69
& \textbf{9.35}$\pm$0.54 & \underline{8.43$\pm$0.51}
& \textbf{9.17}$\pm$0.43 & \textbf{9.45}$\pm$0.52
& 9.26$\pm$0.64
& \textbf{9.14} \\

\hline
\end{tabular}}
\caption{Actor model evaluation results on \textbf{AdaptiveBench}, with GPT-5-Chat as the judge model. \textbf{Bold} indicates the best performance within the same model scale, and \underline{underline} indicates the second-best performance within the same scale. Models marked with * are evaluated using their officially released checkpoints without additional training, while unmarked baselines are re-trained under our experimental setup. $\pm$ denotes the standard deviation computed over all samples.}

\label{tab:actor_model_adaptivebench_result_main}
\end{table*}

\paragraph{(1) Strong LLMs, instruction-tuned models, and reasoning models excel in adaptive role-playing.}
Claude Sonnet 4.5 and GPT-5-Chat lead the proprietary tier, while the open-source reasoning model QwQ-32B is competitive.
Notably, Qwen2.5-7B-Instruct already surpasses Gemini-2.5-Pro overall, and Qwen2.5-14B-Instruct approaches GPT-4o-mini; scaling Qwen2.5 to 72B brings diminishing returns, suggesting a bottleneck for standard instruction-tuned models.
Among 14B models, the reasoning-oriented Qwen3-14B achieves the best overall performance.
We further analyze the impact of prompting strategies by comparing \texttt{Basic} and \texttt{Enhance} settings in Appendix~\ref{app:actor_prompting_strategy}, where we find that the simpler \texttt{Basic} prompting mostly outperforms \texttt{Enhance}, a counterintuitive result suggesting that adaptive role-playing benefits more from flexibility than from overly restrictive guidance.

\paragraph{(2) Existing methods struggle in the adaptive setting and may harm instruction-tuned base models.}
Adaptive role-playing requires integrating \texttt{Thought--Action--Speech--Environment} and handling dynamic scene/role shifts.
Accordingly, specialized models (e.g., Index-1.9B-Character) perform poorly, barely above the neutral baseline.
More strikingly, fine-tuning a strong Qwen2.5-7B-Instruct base (8.37) on Beyond, Crab, or CoSER \emph{reduces} performance (to 6.61, 6.88, and 8.15).
We attribute this to (i) \textit{format mismatch} (missing thought/environment signals), (ii) \textit{interaction mismatch} (dyadic chat vs.\ multi-role coordination), and (iii) \textit{distribution mismatch} (few scene transitions or role reassignments), which together make models brittle under dynamic trajectories.

\paragraph{(3) AdaRPSet yields substantial and robust improvements across model scales and protocols.}
Fine-tuning with AdaRPSet consistently strengthens the Actor Model under AdaptiveBench, yielding broad improvements across sub-metrics (\textbb{CC}/\textbb{EG}/\textbb{II}/\textbb{NP}/\textbb{IC}) and generalizing across model scales (7B--72B) and backbones (Qwen2.5 vs.\ Llama-3.1). Concretely, Qwen2.5-7B-Instruct-Ours improves the overall score from 8.37 to 8.72 (+4.2\%), while Qwen2.5-14B-Instruct-Ours and Qwen2.5-72B-Instruct-Ours rise to 8.84 (+1.6\%) and 8.81 (+1.1\%), respectively. Notably, Llama-3.1-8B-Ours achieves 8.89, outperforming Qwen3-14B (8.88), GPT-4o-mini (8.78), and several 70B/72B baselines, demonstrating that our training recipe elicits strong adaptive capabilities even at smaller scales. 
Importantly, these improvements are also robust across evaluation protocols: beyond automatic judging, human evaluation results (Appendix~\ref{app:subsec:human_evaluation}) confirm the same trend, and our method \emph{mostly} outperforms baselines on CharacterArena and CharacterBench (Appendix~\ref{app:subsec:characterarena_results}). We further provide qualitative case studies in Appendix~\ref{app:sec:case_study} to illustrate these gains in practice.

\subsubsection{Scene Manager}
\label{subsubsec:scene_manager_main}
We evaluate the \textbf{Scene Manager} on AdaptiveBench by fixing the User and Actor models (Doubao-1.5-Pro-Character) and varying only the manager. We use the \texttt{Enhance} system prompt and GPT-5-Chat as the judge (ablations in Appendix~\ref{app:scene_judge_ablation}). Results in Table~\ref{tab:scene_manager_results} (where 5 point denotes neutral performance) reveal the following:

\begin{table*}[t]
\centering
\small 
\setlength{\tabcolsep}{4pt}      
\renewcommand{\arraystretch}{0.95} 
\begin{adjustbox}{width=0.8\textwidth}
\begin{tabular}{l c c c c}
\toprule
Models & Scene Understanding & Speaker Discipline & Role Introduction Judgment & Overall Assessment \\
\midrule
\multicolumn{5}{l}{\textbf{Close-source LLMs}} \\
GPT-4o-mini & 7.64$\pm$1.13 & \underline{8.55$\pm$0.87} & 7.18$\pm$1.26 & 7.64$\pm$1.01 \\
GPT-5-Chat & \underline{8.03$\pm$0.75} & 8.15$\pm$1.30 & 7.78$\pm$1.02 & \underline{7.90$\pm$0.90} \\
Claude Sonnet 4.5 & \textbf{8.21}$\pm$0.57 & \textbf{8.62}$\pm$0.83 & \textbf{8.05}$\pm$0.85 & \textbf{8.17}$\pm$0.57 \\
Doubao-1-5-Pro-Character & 7.67$\pm$0.68 & 7.39$\pm$1.35 & \underline{7.95$\pm$1.12} & 7.51$\pm$0.91 \\
\midrule
\multicolumn{5}{l}{\textbf{Open-source LLMs}} \\
Qwen2.5-7B-Instruct & 7.55$\pm$0.89 & 8.35$\pm$1.08 & 7.25$\pm$1.17 & 7.52$\pm$0.87 \\
\textbf{Qwen2.5-7B-Instruct-Ours} & 7.96$\pm$0.71 & 8.21$\pm$1.11 & 8.00$\pm$1.04 & 7.93$\pm$0.71 \\
Qwen2.5-14B-Instruct & 7.65$\pm$0.86 & 8.42$\pm$1.12 & 7.34$\pm$1.16 & 7.63$\pm$0.84 \\
\textbf{Qwen2.5-14B-Instruct-Ours} & \underline{8.23$\pm$0.44} & 8.25$\pm$0.84 & \textbf{8.64$\pm$0.59} & \underline{8.37$\pm$0.52} \\
Qwen3-14B & 7.81$\pm$0.66 & 8.49$\pm$0.84 & 7.60$\pm$1.10 & 7.80$\pm$0.69 \\
Qwen2.5-72B-Instruct & 7.86$\pm$0.53 & \underline{8.51$\pm$0.76} & 7.65$\pm$0.99 & 7.82$\pm$0.61 \\
Llama-3.1-70B-Instruct & 7.75$\pm$0.75 & 8.33$\pm$0.99 & 7.85$\pm$1.19 & 7.79$\pm$0.84 \\
QwQ-32B & \textbf{8.48$\pm$0.54} & \textbf{8.92$\pm$0.70} & 7.97$\pm$0.71 & \textbf{8.41$\pm$0.55} \\
\bottomrule
\end{tabular}
\end{adjustbox}
\caption{Scene Manager evaluation results on \textbf{AdaptiveBench}, with GPT-5-Chat as the judge model. \textbf{Bold} indicates the best performance within the same model scale, and \underline{underline} indicates the second-best performance within the same scale. $\pm$ denotes the standard deviation computed over all samples.}
\label{tab:scene_manager_results}
\end{table*}

\paragraph{(1) Standard instruction tuning yields compressed performance, while reasoning models break the ceiling.}
Most instruction-tuned models cluster in a narrow Overall Assessment range (7.5--7.9), suggesting that general-purpose tuning does not explicitly optimize scene coordination skills (e.g., state tracking and turn-taking). For instance, Qwen2.5-7B/14B/72B and Llama-3.1-70B improve only marginally and plateau around 7.8, forming a practical ceiling for this class. In contrast, stronger proprietary models (Claude Sonnet 4.5, GPT-5-Chat) perform consistently better, and the reasoning-oriented QwQ-32B reaches 8.41, indicating that reasoning-centric training better supports high-level orchestration than standard instruction tuning alone. Moreover, Appendix~\ref{app:scene_prompting_analysis} shows that \texttt{Enhance} prompts (with stronger constraints) improve Scene Manager performance, opposite to the trend for the Actor Model.

\paragraph{(2) AdaSMSet substantially improves scene management and approaches the best open-source reasoning model.}
Training on AdaSMSet yields clear gains for the Scene Manager. Qwen2.5-14B-Instruct-Ours reaches 8.37 overall, outperforming Claude Sonnet 4.5 (8.17) and improving over the base Qwen2.5-14B-Instruct (7.63) by +0.74 (+9.7\%). \textit{Speaker Discipline} drops slightly, which is expected since AdaSMSet is derived from AdaRPSet-Synthesis and partially inherits GPT-5-Chat’s turn-taking patterns (with \texttt{pick\_speaker} as the most frequent action). Importantly, \textit{Scene Understanding} and \textit{Role Introduction Judgment} improve substantially, yielding a strong net gain and bringing performance close to the best open-source model here (QwQ-32B, 8.41).  A similar trend holds for Qwen2.5-7B-Instruct: the overall score rises from 7.52 to 7.93 (+0.41), with consistent gains on \textit{Scene Understanding} and \textit{Role Introduction Judgment}, while \textit{Speaker Discipline} remains comparable.

\subsection{Ablation Study}
\label{subsec:ablation_study}

To quantify the contribution of each AdaRPSet component, we train Actor Models with (i) \textbf{AdaRPSet-Extracted} and (ii) the full AdaRPSet (\textit{Extracted}+\textit{Synthesis}). We evaluate our method on AdaptiveBench using GPT-5-Chat as the judge, and the results are shown in Table~\ref{tab:actor_model_adaptivebench_ablation}.

\paragraph{AdaRPSet-Extracted alone provides consistent gains and improves stability.}
On Qwen2.5-7B-Instruct, training with extracted data improves the overall score from 8.37 to 8.49 (+0.12). The gains are broad-based across most sub-metrics (e.g., \textbb{CC} and \textbb{IC}), indicating that AdaRPSet-Extracted effectively teaches the model to follow the unified \texttt{Thought--Action--Speech--Env} protocol and to better maintain character consistency during interaction. For the few dimensions where the mean does not increase (e.g., \textbb{EA}/\textbb{RA}/\textbb{STB}), we observe noticeably smaller variance, suggesting more stable behavior across different trajectories.

\paragraph{AdaRPSet-Synthesis yields further improvements and is crucial for generalization.}
When we incorporate the synthesis data, Qwen2.5-7B-Instruct improves to 8.72 overall, i.e., +0.23 over \textit{Extracted}-only and +0.35 over the base model (8.37 $\rightarrow$ 8.49 $\rightarrow$ 8.72). Gains are more pronounced on interaction- and narrative-related dimensions (e.g., \textbb{II}/\textbb{NP}), consistent with AdaRPSet-Synthesis containing more adaptive trajectories with scene transitions and role re-assignment. This highlights the complementarity of the two subsets: extracted data strengthens format alignment and core role-playing skills, while synthesis data improves adaptivity under dynamic trajectories. This trend generalizes to other model families. For Llama-3.1-8B, we omit the base result because the vanilla model does not reliably follow our structured protocol. Training with \textit{AdaRPSet-Extracted} alone is insufficient (7.70 overall; \textbb{IC}: 6.45), whereas adding \textit{AdaRPSet-Synthesis} substantially boosts performance to 8.89 overall (+1.19) and \textbb{IC} to 9.49, demonstrating strong generalization beyond the Qwen family.

\begin{table*}[t]
\centering
\resizebox{\textwidth}{!}{
\begin{tabular}{l c c c c c c c c c c c c c}
\hline
\multicolumn{14}{l}{\makecell[l]{
\textbb{ICoh}: Internal Coherence ~~~
\textbb{SSF}: Speaking Style Fidelity ~~~
\textbb{LFH}: Language Fluency \& Human-likeness ~~~
\textbb{IPF}: Identity \& Profile Fidelity ~~~
\textbb{MVS}: Motivation \& Value Stability \\
\textbb{EA}: Environmental Awareness ~~~
\textbb{EU}: Environmental Utilization ~~~
\textbb{CR}: Contextual Responsiveness ~~~
\textbb{RA}: Relationship Awareness ~~~
\textbb{ATT}: Attractiveness ~~~
\textbb{STB}: Stability ~~~
\textbb{IC}: Instruction Compliance \\
\textbb{CC}: Character Consistency ~~~
\textbb{EG}: Environmental Grounding ~~~
\textbb{II}: Interpersonal Interaction ~~~
\textbb{NP}: Narrative Progression
}} \\
\hline
\multirow{2}{*}{\textbb{Models}}
& \multicolumn{5}{c}{\textbb{CC}}
& \multicolumn{2}{c}{\textbb{EG}}
& \multicolumn{2}{c}{\textbb{II}}
& \multicolumn{2}{c}{\textbb{NP}}
& \multicolumn{1}{c}{\textbb{IC}}
& \multirow{2}{*}{\textbb{Average}} \\
\cmidrule(lr){2-6}
\cmidrule(lr){7-8}
\cmidrule(lr){9-10}
\cmidrule(lr){11-12}
\cmidrule(lr){13-13}
& \textbb{ICoh} & \textbb{SSF} & \textbb{LFH} & \textbb{IPF} & \textbb{MVS}
& \textbb{EA} & \textbb{EU}
& \textbb{CR} & \textbb{RA}
& \textbb{ATT} & \textbb{STB}
& \textbb{IC}
&  \\
\hline
\multicolumn{14}{l}{\textbf{Qwen2.5-7B-Instruct}} \\
\hline
Base
& 8.61$\pm$0.99 & 7.70$\pm$0.91 & 8.70$\pm$0.96 & 9.03$\pm$1.02 & 8.35$\pm$1.06
& 8.24$\pm$1.06 & 7.33$\pm$0.95
& 8.81$\pm$0.97 & 7.80$\pm$0.89
& 7.73$\pm$0.96 & 8.89$\pm$0.97
& 9.23$\pm$0.72
& 8.37 \\
+Extracted
& 8.67$\pm$0.47 & 8.00$\pm$0.42 & 8.82$\pm$0.38 & 9.29$\pm$0.45 & 8.40$\pm$0.61
& 8.24$\pm$0.56 & 7.36$\pm$0.64
& 8.89$\pm$0.38 & 7.96$\pm$0.51
& 7.96$\pm$0.51 & 8.87$\pm$0.45
& 9.42$\pm$0.49
& 8.49 \\
+Extracted+Synthesis
& 8.86$\pm$0.92 & 8.28$\pm$0.97 & 8.83$\pm$0.94 & 9.51$\pm$1.08 & 8.81$\pm$1.07
& 8.54$\pm$1.01 & 7.65$\pm$0.90
& 8.99$\pm$0.94 & 8.11$\pm$0.94
& 8.38$\pm$0.98 & 9.04$\pm$0.99
& 9.70$\pm$0.46
& 8.72 \\
\hdashline
\multicolumn{14}{l}{\textbf{Llama-3.1-8B}} \\
\hline
+Extracted
& 7.97$\pm$1.24 & 7.49$\pm$0.98 & 7.50$\pm$1.57 & 8.68$\pm$1.14 & 8.30$\pm$1.12
& 7.75$\pm$1.13 & 6.83$\pm$1.08
& 8.36$\pm$1.14 & 7.55$\pm$1.08
& 7.48$\pm$1.13 & 8.07$\pm$1.25
& 6.45$\pm$1.75
& 7.70 \\
+Extracted+Synthesis
& 9.00$\pm$0.00 & 8.39$\pm$0.49 & 8.93$\pm$0.26 & 9.69$\pm$0.48 & 9.08$\pm$0.63
& 8.77$\pm$0.49 & 7.84$\pm$0.44
& 9.26$\pm$0.46 & 8.36$\pm$0.52
& 8.52$\pm$0.50 & 9.39$\pm$0.49
& 9.49$\pm$0.54
& 8.89 \\
\hline
\end{tabular}}
\caption{Actor model ablation results on \textbf{AdaptiveBench} (judge: GPT-5-Chat). ``Extracted'' denotes training with AdaRPSet-Extracted, and ``Extracted+Synthesis'' denotes training with full AdaRPSet (AdaRPSet-Extracted + AdaRPSet-Synthesis). $\pm$ denotes the standard deviation computed over all samples.}
\label{tab:actor_model_adaptivebench_ablation}
\end{table*}

\section{Conclusion}
\label{sec:conclusion}
We present \textbf{AdaMARP}, a general role-playing framework that boosts immersion and adaptability with (i) an environment-aware format interleaving thought, action, environment, and speech, and (ii) a discrete-action Scene Manager for multi-character coordination, scene transitions, and dynamic role addition. We release AdaRPSet, AdaSMSet, and AdaptiveBench for trajectory-level evaluation. Experiments show consistent improvements in role consistency, narrative coherence, and environmental grounding across model scales; our 8B Actor and 14B Scene Manager outperform several larger proprietary systems, pointing toward agentic, open-ended role-playing beyond fixed response patterns.


\section*{Ethical Considerations}
\label{sec:ethical_considerations}

The extracted subset of AdaRPSet is derived from existing literary works. We emphasize that the dataset is used \emph{exclusively for scientific research} and not for any commercial purpose. The data consists of transformed and abstracted representations rather than verbatim copyrighted text, and all rights to the original works remain with their respective copyright holders. We provide this dataset solely for academic use and disclaim responsibility for any misuse beyond its intended research scope.

\setcitestyle{numbers,square}
\bibliography{yoUTU_bib}

\appendix

\section{Additional Clarifications on Dataset Comparison}
\label{app:dataset_comparison_details}

This appendix provides detailed clarifications for several datasets summarized in Table~\ref{tab:dataset_comparison}, particularly where profile completeness or annotation symbols (e.g., \pmark) may otherwise be ambiguous.

\paragraph{RoleLLM.}
The released RoleLLM dataset contains only character names, instruction-style queries, and the corresponding model outputs. Although the paper states that character profiles are incorporated during data construction, the released dataset itself does not include these profiles, nor are they available in the accompanying open-source repository. As a result, character profile fields are marked as unavailable in Table~\ref{tab:dataset_comparison}.

\paragraph{CharacterLLM.}
For CharacterLLM, the \pmark symbol indicates a \emph{think-then-speak} paradigm rather than continuously interleaved reasoning and dialogue. Its data construction strategy segments a character’s life experience into multiple independent episodes, each treated as a separate scene for interaction. Consequently, individual dialogue samples do not contain a complete or unified character profile, motivating the partial designation in the profile-related columns.

\paragraph{CharacterGLM.}
In CharacterGLM, profile information is inconsistently provided: a small subset of characters includes partial profiles, while the majority of instances omit them. Accordingly, relevant profile attributes are marked with \pmark in Table~\ref{tab:dataset_comparison} to reflect this incomplete and non-uniform availability.

\paragraph{ROLEPERSONALITY.}
The final released ROLEPERSONALITY dataset consists solely of instruction prompts and corresponding outputs. Background information is referenced internally during LLM generation but is not released as structured profile data. The reported sizes of 32{,}089 and 32{,}767 correspond to single-turn and five-turn dialogues, respectively. Even when background information is concatenated into the system prompt, it lacks a fixed schema: different attributes may appear arbitrarily, preventing a consistent profile structure.

\paragraph{SimChat.}
For SimChat, the \pmark annotation in the \textit{Thought} column likewise denotes a \emph{think-then-speak} format, where reasoning precedes the response rather than being interleaved with natural language output.

\paragraph{TAILORGEN.}
TAILORGEN has not been publicly released at the time of writing, making it impossible to directly verify the details of its character profiles. The authors state that profile attributes are organized with reference to CharacterLLM and RoleLLM. The ``Fixed~2$^\ast$'' dialogue length reported in Table~\ref{tab:dataset_comparison} is inferred from the described pipeline: the method primarily constructs a query and then generates a corresponding response, which suggests a two-turn interaction.

\section{Details of Comprehensive Character Information}
\label{app:sec:character_profile_details}
This appendix elaborates the seven dimensions used in the main character profile in AdaMARP.

\textbf{(I) Identity and Appearance} captures fundamental attributes such as name, age, gender, and occupation, as well as a detailed description of key physical traits.
\textbf{(II) Personality and Psychology} specifies the character’s behavioral tendencies, typical emotional reaction patterns, and underlying values or preferences.
To ensure linguistic and interactional authenticity, \textbf{(III) Speaking Style} defines the character’s verbal rhythm, tone, and habitual lexical choices, ranging from formal discourse to sarcastic or indirect expression.

Beyond these intrinsic traits, \textbf{(IV) Abilities, Interests, and Achievements} represent the character’s hard and soft skills, personal hobbies, and representative accomplishments.
The character is situated within a broader context through \textbf{(V) Social and Historical Context}, which delineates the social environment, era, family background, and cultural or class positioning.
A central narrative component is \textbf{(VI) Personal History Arc}, which encodes significant past experiences and clarifies the current stage of the character’s development within an ongoing narrative.
Finally, \textbf{(VII) Relationships} provides natural-language descriptions of the character’s connections with other entities, ensuring that interpersonal interactions remain consistent with established social dynamics.

Collectively, these seven dimensions enable AdaMARP to support coherent, immersive, and evolution-aware role-playing behavior.

\section{Details of Adaptive Role-Playing Framework}
\label{app:sec:details_of_adaptive_role-playing_framework}

This appendix provides the implementation-level details of the AdaMARP framework, including its execution flow, key variables, and the pseudo-code specification. The goal is to clarify how adaptive role selection, scene transitions, and dynamic character introduction are realized during role-playing interactions.

\subsection{Framework Components and State Definition}
At each interaction step $t$, the framework maintains a global interaction state $\mathcal{G}_t$, which consists of the active role set $\mathcal{R}_t = \{r_1, r_2, \dots, r_{n_t}\}$ and the dialogue history $\mathcal{H}_t$. 
Unlike conventional designs, the current scene description is \emph{not} explicitly stored as a state variable. Instead, scene information is implicitly encoded in the outputs of the Scene Manager and preserved within the interaction history. Each role $r_i \in \mathcal{R}_t$, including the user role, is associated with a structured profile $\mathcal{P}_i$ and a scene-dependent motivation $\mathcal{M}_i^t$.

\subsection{Scene Manager Actions}
The Scene Manager $\mathcal{S}$ selects an action $m_t$ from the predefined action space

\begin{equation}
\begin{aligned}
\mathcal{M} = \{&
\texttt{init\_scene},\;
\texttt{pick\_speaker},\;
\texttt{switch\_scene},\;
\texttt{add\_role},\;
\texttt{end}
\}.
\end{aligned}
\end{equation}

The first action issued by the Scene Manager $\mathcal{S}$ is always \texttt{init\_scene}, which initializes the role-playing interaction by providing an initial scene description. 
During the interaction, when $m_t=\texttt{switch\_scene}$, $\mathcal{S}$ outputs a new scene description without overwriting previous scenes. Instead, all scene transitions are recorded in the dialogue history, allowing past scenes to remain accessible as contextual evidence.

For each action $m_t \in \mathcal{M}$, $\mathcal{S}$ outputs both the selected action and a natural-language rationale explaining the decision. 
When $m_t=\texttt{pick\_speaker}$, the action specifies the role selected to generate the next in-character response.
When $m_t=\texttt{add\_role}$, the action output additionally includes the new character’s name, profile, and initial motivation.
When $m_t=\texttt{end}$, the interaction is explicitly terminated.

\subsection{Overall Execution Flow}
Algorithm~\ref{alg:adaptive_marps} illustrates the execution procedure of AdaMARP. 
The interaction is initialized by an explicit scene initialization action, after which the framework alternates between high-level control decisions and in-character response generation.

\subsection{Prompt Realization}
The above framework is instantiated through three structured prompt templates corresponding to the Actor Model $\mathcal{A}$, the User Model $\mathcal{U}$, and the Scene Manager $\mathcal{S}$. Each prompt explicitly encodes the symbolic constraints, behavioral rules, and decision logic defined in the main text, thereby bridging high-level narrative control with low-level in-character generation.

The \textbf{User prompt} constrains the model to behave as a realistic human participant immersed in the story world, emphasizing first-person dialogue, limited utterance length, and proactive narrative momentum. The \textbf{Actor prompt} governs individual non-user characters, enabling explicit separation of internal thoughts, visible actions, spoken dialogue, and environmental effects. The \textbf{Scene Manager prompt} implements global orchestration logic, including speaker rotation, scene transitions, and dynamic role introduction, while remaining strictly output-constrained to structured JSON decisions. For completeness and reproducibility, the full prompt templates used in our implementation are provided in Tables~\ref{tab:prompts_user}, \ref{tab:actor_model_prompt}, and \ref{tab:scene_manager_prompt_basic}.

\section{Details of AdaRPSet Construction}
\label{app:sec:details_of_adarpset_construction}

\subsection{AdaRPSet-Extracted}
\label{app:subsec:adarpset_extracted}

AdaRPSet-Extracted is constructed from full-length literary works and is designed to provide supervision for generating role-playing messages that interleave \textit{thought}, \textit{action}, \textit{speech}, and \textit{environment} (Section~\ref{subsubsec:immersive_message}). The construction pipeline consists of three stages: \textbf{Chunking}, \textbf{LLM-based Extraction}, and \textbf{LLM-based Profile Generation}. The overall construction pipeline is summarized in Algorithm~\ref{alg:adarpset_extracted}.

\paragraph{Chunking.}
Given the full text of a book, we first identify candidate chapter boundaries using a set of predefined regular-expression rules that capture common chapter-heading patterns, such as \texttt{Chapter 1}, \texttt{CHAPTER I}, roman numerals, or markdown-style headers (e.g., \texttt{\# Chapter Name}). This step produces initial chapter-level segments.
We then merge consecutive chapters into larger chunks while keeping chapters as intact as possible, until reaching a target chunk size determined by the context window of the extraction LLM.
To balance extraction accuracy and coverage, we set the default chunk size to \textbf{8,192 tokens}, based on multiple pilot runs with GPT-5-Chat. If chapter-title matching fails for a given book, we fall back to a fixed-size splitting strategy using the same target chunk size. 

\paragraph{LLM-based Extraction.}
For each chunk, we use GPT-5-Chat to perform structured information extraction. The model is prompted to: (i) recover chapter beginnings if present; (ii) identify several salient plot units within the chunk (each plot is delimited by its first and last sentence as they appear in the chunk); (iii) summarize each plot and assign a prominence score; (iv) identify key characters in each plot and provide lightweight character descriptors (\emph{not} the full profile) and plot-specific experiences; and (v) extract a conversation trajectory for each plot.

A key requirement is that the LLM produces the extracted conversations \emph{directly} in our unified message format: each dialogue turn is a single string that may combine spoken words with explicit \textit{thought} in \texttt{[\,]}, \textit{action} in \texttt{(\,)}, and \textit{environment} in \texttt{<\,>}. We explicitly instruct the LLM to (a) strictly distinguish actions (character-originated behaviors) from environment (external sensory cues or setting changes), (b) render thoughts in first-person perspective, and (c) incorporate relevant narrative descriptions surrounding the dialogue into action/environment tags to improve immersion. The full extraction prompt and JSON schema are provided in Table~\ref{tab:book_prompt_part1},~\ref{tab:book_prompt_part2} and \ref{tab:book_prompt_part3}. Additional engineering details (e.g., handling cases where a plot spans the chunk boundary but the chapter boundary is ambiguous) are implemented in our released codebase.

\begin{algorithm}[htbp]
\caption{Construction of AdaRPSet-Extracted}
\label{alg:adarpset_extracted}
\footnotesize
\begin{algorithmic}[1]
\REQUIRE Book text $B$; chapter-title regex set $\mathcal{R}$; target chunk size $T$ (tokens, default 8192); extraction LLM $\mathcal{M}_{ext}$; profiling LLM $\mathcal{M}_{prof}$
\ENSURE Plot/conversation records $\mathcal{D}$ and character profiles $\mathcal{P}$

\STATE $\mathcal{D}\leftarrow \emptyset$; $\mathcal{P}\leftarrow \emptyset$
\vspace{2pt}
\STATE \textbf{// Stage 1: Chunking}
\STATE $\mathcal{S}\leftarrow \textsc{SplitByChapterTitles}(B,\mathcal{R})$
\IF{$\mathcal{S}=\emptyset$}
    \STATE $\mathcal{C}\leftarrow \textsc{FixedSplit}(B, T)$ \COMMENT{fallback: fixed-size chunking}
\ELSE
    \STATE $\mathcal{C}\leftarrow \textsc{MergeSegmentsToLimit}(\mathcal{S}, T)$
    \COMMENT{keep chapters intact when possible}
\ENDIF

\vspace{2pt}
\STATE \textbf{// Stage 2: LLM-based Extraction}
\FOR{each chunk $c \in \mathcal{C}$}
    \STATE $J \leftarrow \mathcal{M}_{ext}(c)$
    \STATE $\mathcal{D}\leftarrow \mathcal{D}\cup \textsc{Parse}(J)$
\ENDFOR

\vspace{2pt}
\STATE \textbf{// Stage 3: LLM-based Profile Generation}
\STATE $\mathcal{K}\leftarrow \textsc{CollectAllCharacterNames}(\mathcal{D})$
\FOR{each character $k \in \mathcal{K}$}
    \STATE $E_k \leftarrow \textsc{AggregateEvidence}(\mathcal{D}, k)$
    \COMMENT{plot summaries, character experiences, and all extracted messages}
    \STATE $P_k \leftarrow \mathcal{M}_{prof}(E_k)$ \COMMENT{7-dim profile + relationships}
    \STATE $\mathcal{P}\leftarrow \mathcal{P}\cup \{(k, P_k)\}$
\ENDFOR
\RETURN $\mathcal{D}, \mathcal{P}$
\end{algorithmic}
\end{algorithm}

\paragraph{LLM-based Profile Generation.}
After plot- and conversation-level extraction, we construct character profiles by aggregating evidence across the entire book. For each character, we collect: (i) summaries of all plots in which the character appears, (ii) the character’s plot-specific experience fields, and (iii) all extracted dialogue turns attributed to the character. Conditioned on this aggregated evidence, we prompt the LLM to synthesize a comprehensive profile following the seven-dimensional schema in Section~\ref{subsubsec:character_info}, including relationship descriptions derived from recurring interactions with other characters. The full profiling prompt template is reported in Table~\ref{tab:prompts_character_profile}.

\subsection{AdaRPSet-Synthesis}
\label{app:subsec:adarpset_synthesis}

While AdaRPSet-Extracted enables the Actor Model to produce well-formed messages conditioned on character profiles and local context, we find that models still struggle with \emph{dynamic} role-playing behaviors, especially (i) \textbf{scene transitions} and (ii) \textbf{introducing new characters} on the fly. In addition, literary narratives can be relatively conservative in interaction patterns. To better supervise these capabilities, we further construct \textbf{AdaRPSet-Synthesis} by prompting an LLM to generate synthetic, plot-driven role-playing trajectories with explicit environment control signals.

\paragraph{Plot specification and trajectory format.}
Each synthesized record corresponds to a \textbf{plot} that contains: (1) an \textbf{initial scenario} describing the starting situation; (2) a set of characters including a \textbf{main character} and \textbf{multiple other characters}, where each character is assigned a \textbf{Profile} and an \textbf{initial Motivation}; and (3) a multi-turn \textbf{dialogue trajectory} written \emph{directly} in our unified messaging format. Besides character utterances (optionally with explicit \textit{thought}, \textit{action}, and \textit{environment} fields), the trajectory includes special messages from a \textbf{scene manager} that perform state-changing control actions such as \texttt{Action=add\_role} (introducing a new character) and \texttt{Action=switch\_scene} (transitioning to a new scene). We intentionally exclude \texttt{pick-speaker} messages so that the data focuses on the core state changes rather than turn-selection artifacts. We require every trajectory to include \emph{at least one} scene switch and \emph{at least one} role addition.

\paragraph{Themes and descriptions.}
To diversify event structures and interaction patterns, we generate plots under 20 themes, each with an explicit high-level guideline:
\begin{itemize}[leftmargin=1.2em, labelsep=0.4em,
                itemsep=0pt, parsep=0pt, topsep=2pt, partopsep=0pt]
    \item \textbf{Adventure}: Characters embark on a journey, facing challenges and discovering new places.
    \item \textbf{Quest}: A specific mission or goal drives the characters' actions.
    \item \textbf{Rescue}: Characters must save someone or something in danger.
    \item \textbf{Battle}: Conflict escalates into a physical or magical confrontation.
    \item \textbf{Escape}: Characters attempt to flee from a dangerous situation.
    \item \textbf{Exploration}: Discovering unknown territories, objects, or secrets.
    \item \textbf{Mystery}: Unexplained phenomena or events spark investigation.
    \item \textbf{Investigation}: Characters collect clues and analyze information to solve a case.
    \item \textbf{Crime-solving}: Characters work together to uncover a criminal or culprit.
    \item \textbf{Puzzle-solving}: Solving riddles, codes, or logical challenges.
    \item \textbf{Conspiracy}: Hidden schemes or secrets are uncovered gradually.
    \item \textbf{Romance}: Characters explore feelings of love, attraction, or affection.
    \item \textbf{Friendship}: Building trust, bonds, and camaraderie between characters.
    \item \textbf{Rivalry}: Competing interests or personalities create tension.
    \item \textbf{Betrayal}: Trust is broken, and hidden motives are revealed.
    \item \textbf{Reconciliation}: Conflicts are resolved, misunderstandings cleared, relationships repaired.
    \item \textbf{Negotiation}: Characters attempt to reach agreements or compromises.
    \item \textbf{Strategy}: Planning, scheming, or discussing complex plans for a goal.
    \item \textbf{Magic}: Supernatural powers or magical phenomena influence events.
    \item \textbf{Apocalypse}: Characters face large-scale disasters or catastrophic events.
\end{itemize}

\paragraph{De-duplication and quality control.}
During generation, we observe that the LLM may repeatedly produce common character names and similar profiles. To mitigate this, we (i) impose explicit constraints in the prompt to avoid overly common names and discourage reusing names across records, and (ii) perform an automatic \textbf{double check} to filter duplicates: if the \textbf{main character name} of a newly generated plot has already appeared in the current collection, we discard the plot and do not add it to the dataset. In addition, we conduct a \textbf{topic-wise manual second-pass review}: for each theme, we manually verify that the \textbf{initial scenario} is distinct across records and that required control actions (\texttt{add\_role}, \texttt{switch\_scene}) occur as specified.

\paragraph{Prompts and examples.}
The complete prompt templates are relatively lengthy and are therefore provided in our open-source code repository to ensure readability.
A qualitative example under the \emph{Adventure} theme is shown in
Tables~\ref{tab:adventure_example_part1}, \ref{tab:adventure_example_part2}, and \ref{tab:adventure_example_part3}.

\subsection{Unified Training Sample Format}
\label{app:subsec:training_sample_format}

To train the Actor Model with consistent supervision across different data sources, we convert both AdaRPSet-Extracted and AdaRPSet-Synthesis into a unified message-based training format (Table~\ref{tab:training_sample_formatter}). Concretely, each training instance is \emph{character-centric}: we designate one character as the \textbf{main character} to be played by the \texttt{assistant}, while all remaining characters (as well as scene-manager control signals) are provided on the \texttt{user} side as context.

\paragraph{Formatting AdaRPSet-Extracted.}
For the extracted subset, we iterate over every extracted plot and then iterate over every character appearing in that plot. For a character $c$, we build one training instance where:
(i) the \textbf{system} message instantiates the role-play prompt by injecting $c$'s seven-dimensional profile and its scene-specific motivation, and also lists the other characters' profiles/motivations (as shown in Table~\ref{tab:training_sample_formatter});
(ii) we prepend an initialization message as a \texttt{user} turn in the form \texttt{scene\_manager: action: init\_scene | initial\_scene: \{...\}}, using the extracted scene description as \texttt{initial\_scene};
(iii) within the dialogue trajectory, all non-$c$ utterances are mapped to \texttt{user} turns in the canonical form \texttt{\{other\_character\_name\}: \{message\_content\}}, whereas $c$'s utterances are mapped to \texttt{assistant} turns in the form \texttt{\{main\_profile\_name\}: \{message\_content\}}. During training, only the \texttt{assistant} message contents are used for loss computation, and the \texttt{user} side is treated as conditioning context.

\paragraph{Formatting AdaRPSet-Synthesis.}
For the synthetic subset, the generated trajectories already follow our required role-playing message structure. We therefore mainly normalize them to the unified wrapper in Table~\ref{tab:training_sample_formatter}: we fill the \textbf{system} message with the selected main character profile/motivation and the other characters' profiles/motivations, keep the initial \texttt{init\_scene} user message, and serialize subsequent turns as \texttt{user}/\texttt{assistant} messages with the same \texttt{Name: Content} convention. When present, additional scene-manager actions (e.g., \texttt{switch\_scene}, \texttt{add\_role}) are kept as \texttt{user} messages with the \texttt{scene\_manager: action: ...} prefix to preserve explicit state changes.

Overall, this conversion ensures that both extracted and synthetic data are presented to the Actor Model in an identical supervision format before training.

\subsection{AdaSMSet Construction}
\label{app:subsec:adasmset}

AdaSMSet is a supervised dataset designed to train the Scene Manager $\mathcal{S}$ to perform high-level control decisions in AdaMARP. We construct AdaSMSet on top of AdaRPSet-Synthesis, as the synthesized trajectories already include explicit scene-manager actions (e.g., \texttt{add\_role}, \texttt{switch\_scene}, and \texttt{end}). The remaining key supervision signal is \textbf{speaker selection}. Specifically, we insert \texttt{pick\_speaker} actions between consecutive character turns, where the target \texttt{speaker} is determined by the next ground-truth turn in the original trajectory. We further generate a natural-language \texttt{reason} field to justify each selection using a strong instruction-following language model; in this work, we instantiate this model with GPT-5-Chat.

\paragraph{System prompt construction.}
For each trajectory, we provide $\mathcal{S}$ with the full story state as the system context (character profiles, motivations, and scene information), and attach a dedicated Scene-Manager instruction that specifies the available actions and the required JSON schema. The resulting training formatter is summarized in Table~\ref{tab:training_scene_manager}.

\paragraph{Handling existing scene-manager actions.}
AdaRPSet-Synthesis already contains explicit scene-manager decisions (e.g., \texttt{init\_scene}, \texttt{switch\_scene}, \texttt{add\_role}, \texttt{end}). We retain these decisions as supervised targets and normalize them into JSON action objects so that $\mathcal{S}$ is trained to output a unified, structured control signal.

\paragraph{Inserting \texttt{pick\_speaker} actions.}
Between any two consecutive turns where the next turn is a \emph{character} message (i.e., not a \texttt{scene\_manager} message), we insert an \texttt{assistant} message:
\texttt{\{"action":"pick\_speaker", "speaker": $\cdot$, "reason": $\cdot$\}}.
The \texttt{speaker} field is derived from the next message by extracting the name prefix before the first colon (``\texttt{Name: ...}''); if unavailable, we fall back to the message role field. We additionally normalize names by stripping an optional \texttt{(user)} suffix for consistency.

\paragraph{LLM-generated \texttt{reason}.}
For each inserted \texttt{pick\_speaker}, we prompt an LLM to generate a single-sentence explanation of why this speaker should act next, conditioned on (i) the cleaned system prompt, (ii) the recent conversation history, and (iii) the pending speaker name. The prompt template is given in Table~\ref{tab:pick_speaker_reason_prompt}. We instruct the LLM to avoid formulaic patterns (e.g., ``X is chosen to speak next'') and to use varied, context-grounded phrasing.

Overall, AdaSMSet provides supervision covering all action types of the Scene Manager. It preserves the original control actions from AdaRPSet-Synthesis—such as scene switching, role addition, and episode termination—and further augments the data with \texttt{pick\_speaker} actions equipped with explicit rationales.

\begin{table*}[t]
\centering
\begin{tabular}{l c c c c c c}
\toprule
\textbf{Dataset} & \textbf{Source} & \textbf{Plots} & \textbf{Roles} & \textbf{Convs.} & \textbf{Utterances} & \textbf{Avg. Turns} \\
\midrule
\textbf{AdaRPSet-Extracted} & 81 Books & 4,443 & 2,608 & 12,525 & 177,157 & 14.14 \\
\textbf{AdaRPSet-Synthesis} & 20 Topics & 9,900 & 29,701 & 9,900 & 273,078 & 27.58 \\
\midrule
\textbf{AdaRPSet (Total)} & -- & 14,343 & 32,309 & 22,425 & 450,235 & 20.08 \\
\bottomrule
\end{tabular}
\caption{Detailed statistics of the \textbf{AdaRPSet} dataset. \textbf{Source} indicates the origin domain (books vs. topics); \textbf{Plots} refers to distinct narrative segments; \textbf{Roles} denotes unique character profiles; \textbf{Convs.} represents the number of training samples (dialogue sessions); and \textbf{Avg. Turns} is the average number of utterances per conversation.}
\label{tab:adarpset_statistics}
\end{table*}

\begin{figure*}[t]
    \centering
    \begin{subfigure}[b]{0.32\textwidth}
        \centering
        \includegraphics[width=\linewidth]{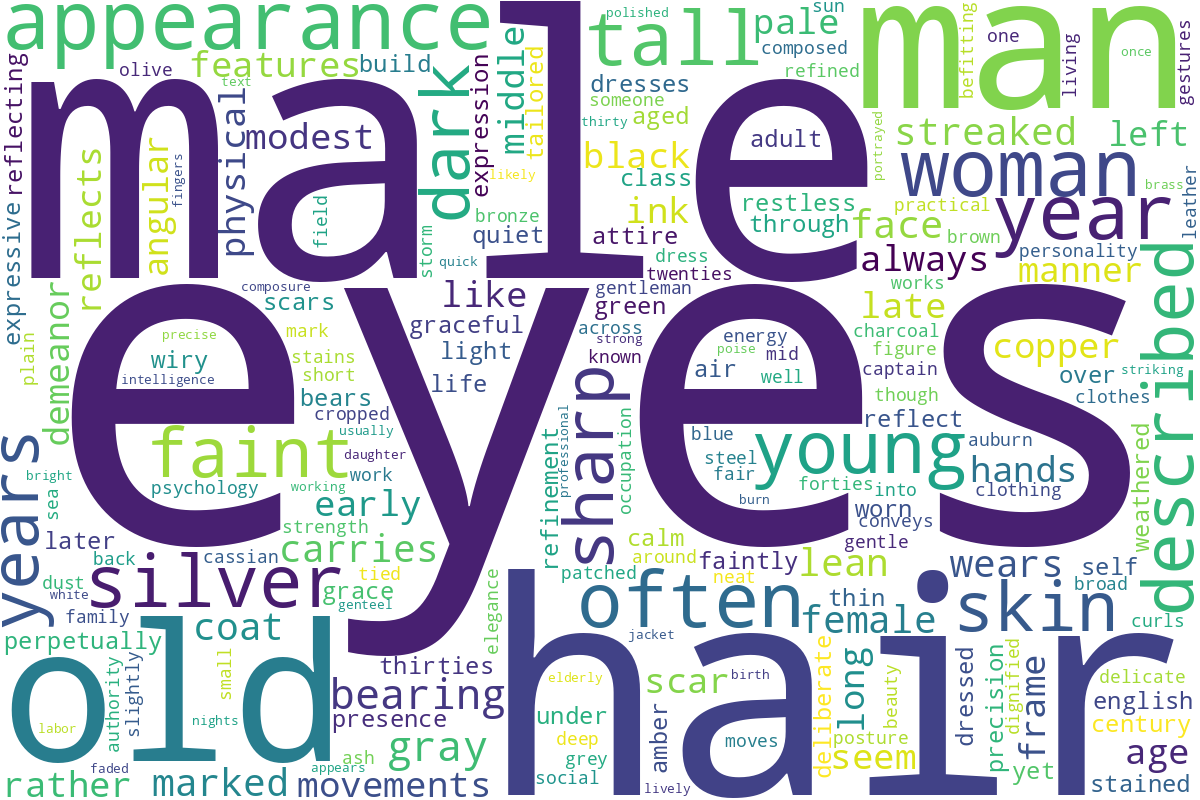} 
        \caption{Identity \& Appearance}
        \label{fig:wordcloud_identity}
    \end{subfigure}
    \hfill
    \begin{subfigure}[b]{0.32\textwidth}
        \centering
        \includegraphics[width=\linewidth]{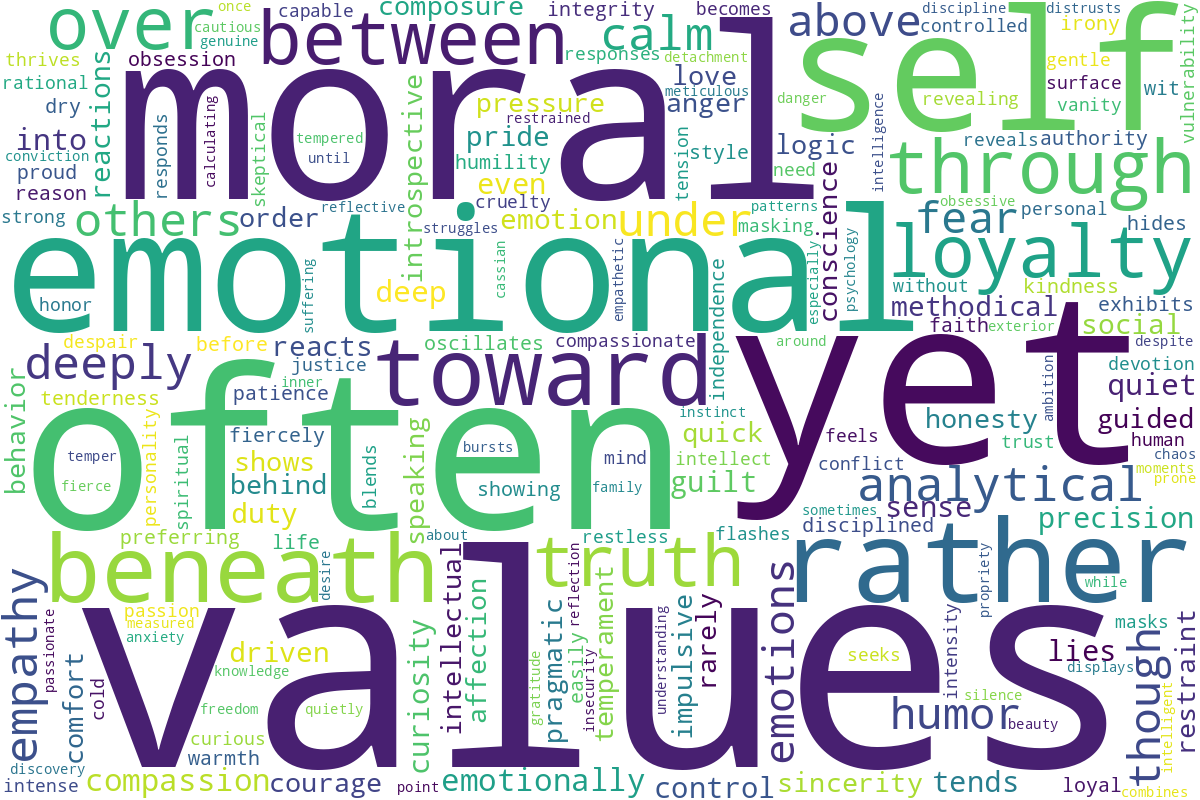}
        \caption{Personality \& Psychology}
        \label{fig:wordcloud_personality}
    \end{subfigure}
    \hfill
    \begin{subfigure}[b]{0.32\textwidth}
        \centering
        \includegraphics[width=\linewidth]{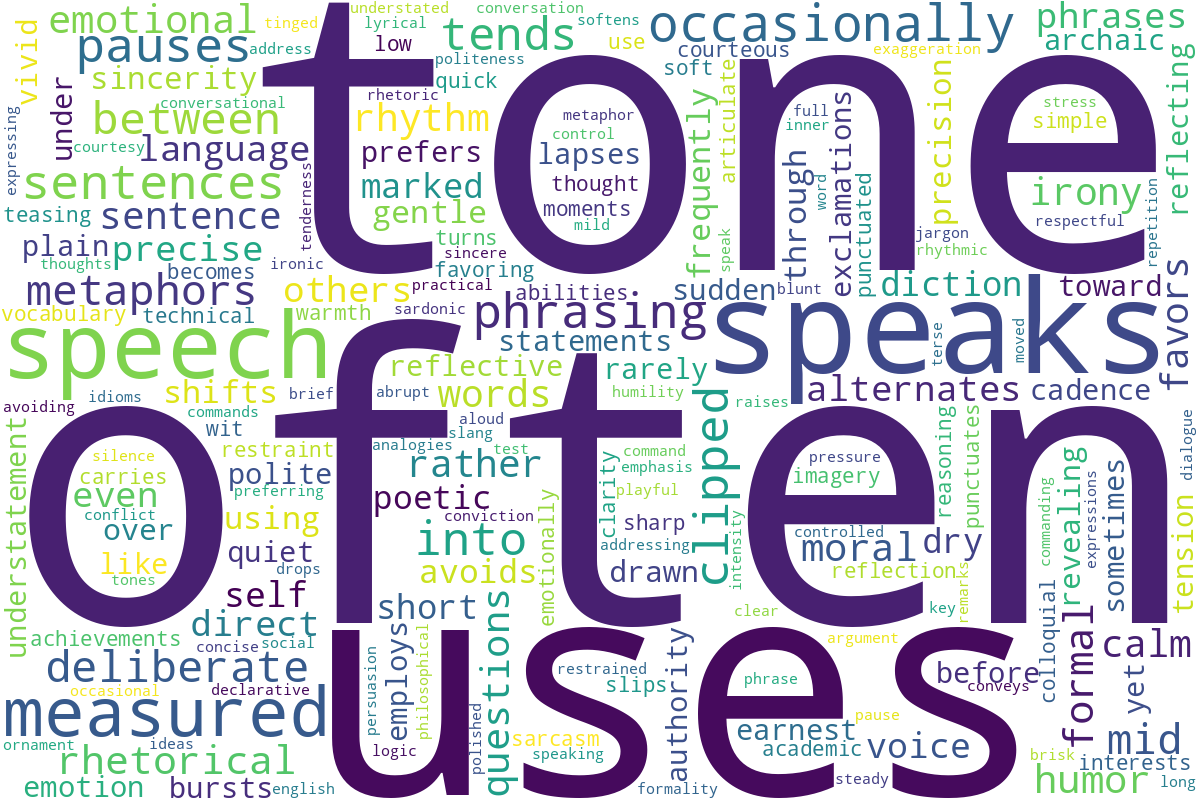}
        \caption{Speaking Style}
        \label{fig:wordcloud_speaking}
    \end{subfigure}
    
    \vspace{0.8em} 
    
    \begin{subfigure}[b]{0.32\textwidth}
        \centering
        \includegraphics[width=\linewidth]{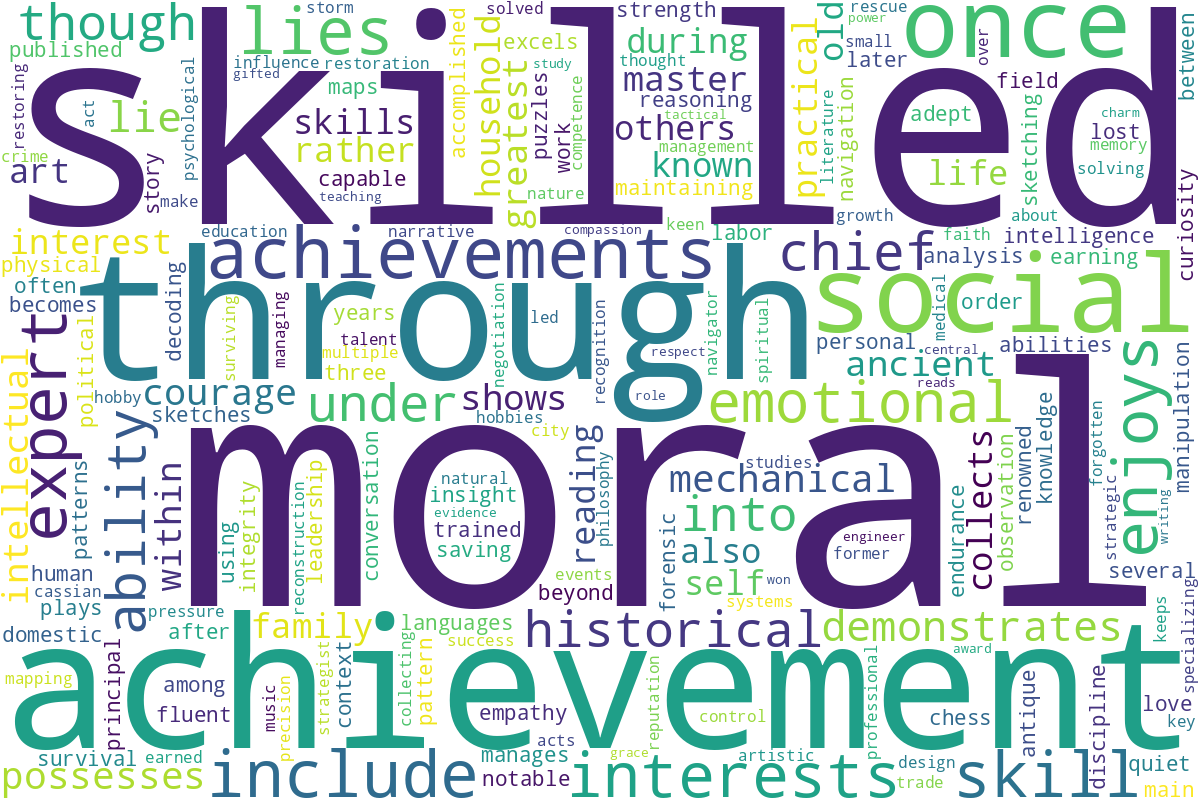}
        \caption{Abilities \& Achievements}
        \label{fig:wordcloud_abilities}
    \end{subfigure}
    \hfill
    \begin{subfigure}[b]{0.32\textwidth}
        \centering
        \includegraphics[width=\linewidth]{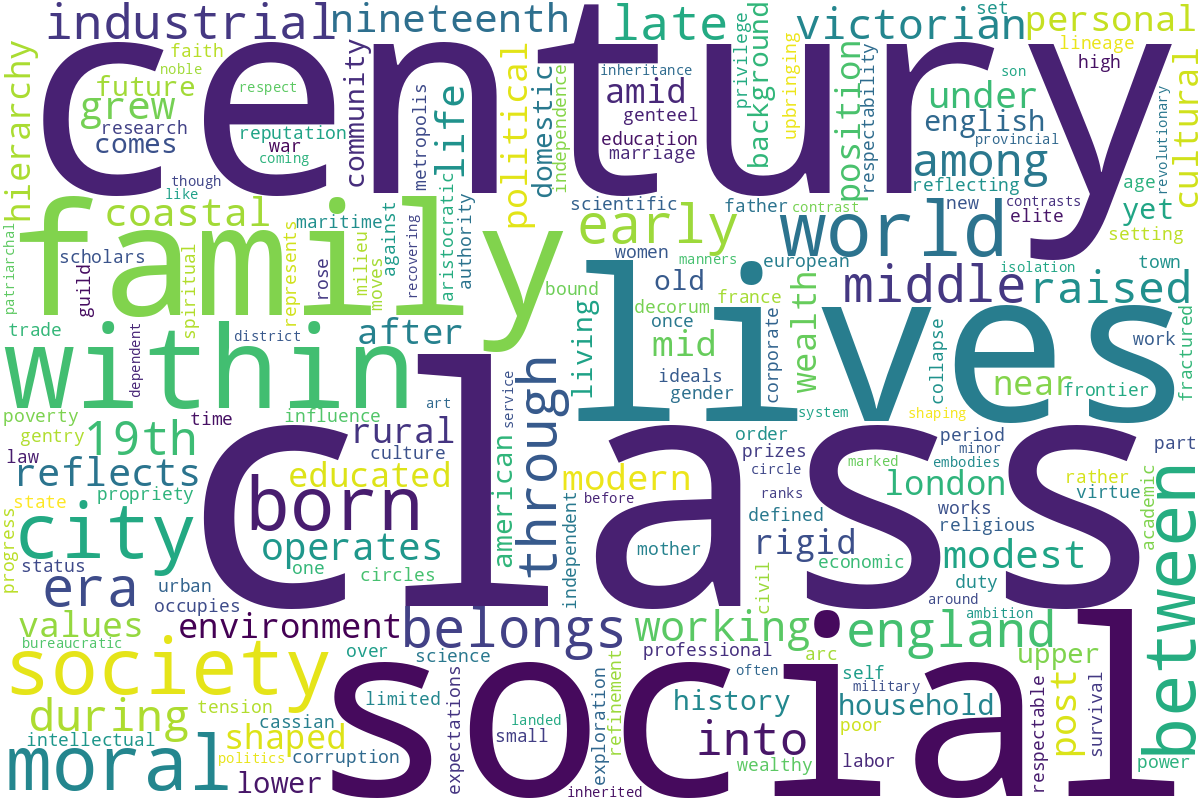}
        \caption{Social \& Historical Context}
        \label{fig:wordcloud_social}
    \end{subfigure}
    \hfill
    \begin{subfigure}[b]{0.32\textwidth}
        \centering
        \includegraphics[width=\linewidth]{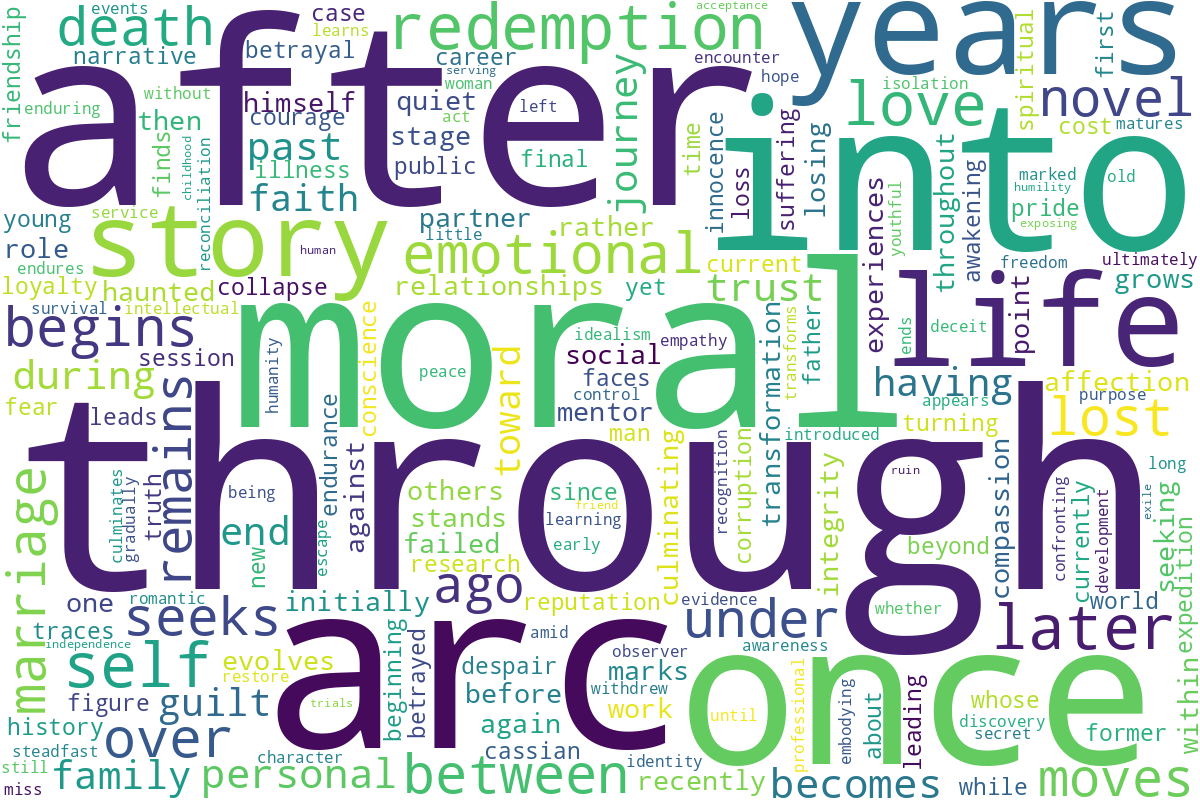}
        \caption{Personal History Arc}
        \label{fig:wordcloud_history}
    \end{subfigure}
    
    \caption{Word cloud visualizations of six key profile dimensions in \textbf{AdaRPSet}. The layout displays: (a) Identity \& Appearance, (b) Personality \& Psychology, (c) Speaking Style, (d) Abilities, Interests \& Achievements, (e) Social \& Historical Context, and (f) Personal History Arc. The diversity of terms indicates a broad coverage of character archetypes and backgrounds.}
    \label{fig:adarpset_profile_wordclouds}
\end{figure*}

\begin{figure*}[t]
    \centering
    \includegraphics[width=0.75\textwidth]{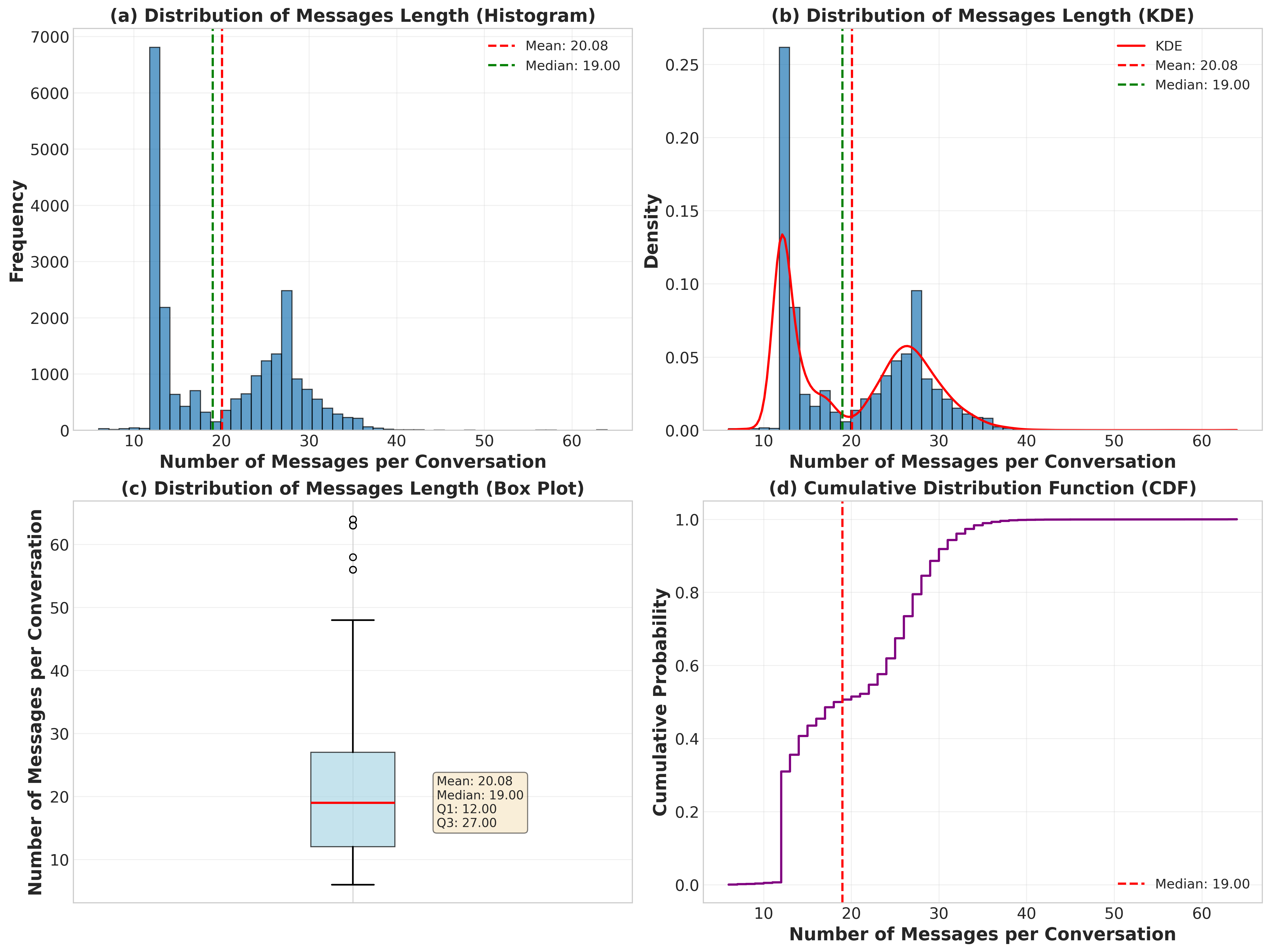}
    \caption{Distribution analysis of message counts per conversation in \textbf{AdaRPSet}. The figure aggregates (a) Histogram, (b) KDE, (c) Boxplot, and (d) CDF. The bimodal nature of the distribution reflects the complementary characteristics of the Extracted and Synthesis subsets.}
    \label{fig:message_distribution}
\end{figure*}

\section{Statistical Analysis of AdaRPSet}
\label{app:sec:statistical_analysis_adarpset}

Table~\ref{tab:adarpset_statistics} provides a detailed statistical summary of AdaRPSet, which comprises two subsets: \textbf{AdaRPSet-Extracted} and \textbf{AdaRPSet-Synthesis}.

\subsection{Basic Statistics}
\label{app:subsec:basic_statistics}

\paragraph{AdaRPSet-Extracted.}
This subset is grounded in literary narratives derived from 81 representative books (listed in Table~\ref{tab:selected_books}). We extracted 4,443 distinct \textbf{Plots} (coherent narrative segments) involving 2,608 unique main and supporting \textbf{Roles}. These plots yield a total of 12,525 \textbf{Conversations}. The number of conversations exceeds the number of plots because we employ a multi-perspective augmentation strategy: for a single multi-character plot, we create separate training samples by designating different participants as the "main character" (the model's role) while treating others as NPCs. This subset contains 177,157 utterances with an average length of 14.14 turns per conversation.

\paragraph{AdaRPSet-Synthesis.}
This subset is generated entirely by LLMs to cover dynamic interaction patterns. It spans 20 diverse \textbf{Topics} (detailed in Appendix~\ref{app:subsec:adasmset}) and contains 9,900 plots. Unlike the extracted set, each synthesized plot corresponds to exactly one conversation sample, involving a total of 29,701 generated roles. The synthetic trajectories differ significantly in complexity, containing 273,078 utterances with a much higher average density of 27.58 turns per conversation. This reflects the deliberate inclusion of extended interactions, scene transitions, and role additions.

In total, \textbf{AdaRPSet} combines these complementary sources to provide over 450k utterances across more than 22k dialogue sessions, offering a robust foundation for training adaptive role-playing agents.

\subsection{Profile Diversity Analysis}
\label{app:subsec:profile_diversity}

To assess the semantic richness of the characters in AdaRPSet, we conduct a granular analysis of the main character profiles across the dataset. We aggregate the textual descriptions from six core dimensions—\textit{Identity \& Appearance}, \textit{Personality \& Psychology}, \textit{Speaking Style}, \textit{Abilities, Interests \& Achievements}, \textit{Social \& Historical Context}, and \textit{Personal History Arc}—and generate word clouds to visualize the distribution of high-frequency attributes.

As illustrated in Figure~\ref{fig:adarpset_profile_wordclouds}, the word-clouds corroborate the breadth and diversity of AdaRPSet across six profile dimensions. In \textit{Identity \& Appearance}, frequent descriptors span physical attributes and demographics (e.g., ``old,'' ``man,'' ``woman,'' ``hair,'' ``tall''). \textit{Personality \& Psychology} captures a wide range of inner states and traits such as ``emotional,'' ``calm,'' ``loyalty,'' and ``analytical.'' \textit{Speaking Style} highlights heterogeneous linguistic manners, including cues like ``speaks,'' ``phrases,'' ``tone,'' and ``rhythm.'' Beyond these, \textit{Abilities, Interests \& Achievements} emphasizes competence- and interest-related terms (e.g., ``master,'' ``skill,'' ``mechanical,'' ``achievements''), while \textit{Social \& Historical Context} reflects varied societal settings (e.g., ``society,'' ``city,'' ``industrial,'' ``Victorian,'' ``world''). Finally, \textit{Personal History Arc} foregrounds life-course events and temporal progression (e.g., ``childhood,'' ``years,'' ``life,'' ``love,'' ``later''). Collectively, this broad lexical coverage suggests that models trained on AdaRPSet can be exposed to diverse character archetypes and rich social backdrops.

\subsection{Message Distribution Analysis}
\label{app:subsec:message_distribution}

To understand the structural complexity and interaction depth of the dataset, we analyze the distribution of message counts (i.e., conversation length) across all dialogue sessions. Figure~\ref{fig:message_distribution} presents a comprehensive visualization using four complementary statistical views: (a) a histogram of message counts, (b) the Kernel Density Estimation (KDE) curve, (c) a boxplot summarizing central tendency and dispersion, and (d) the Cumulative Distribution Function (CDF).

As shown in panels (a) and (b), the dataset exhibits a distinct bimodal distribution with significant peaks concentrated around 10 turns and 30 turns. This pattern aligns with the composition of AdaRPSet: the peak near 10 turns corresponds to the naturally shorter, book-extracted scenes from \textbf{AdaRPSet-Extracted} (avg.\ 14.14 turns), while the peak near 30 turns reflects the extended, dynamic trajectories from \textbf{AdaRPSet-Synthesis} (avg.\ 27.58 turns). 

The boxplot in panel (c) further quantifies these statistics, revealing a median conversation length of 19 turns, with the first quartile ($Q1$) at 12 turns and the third quartile ($Q3$) at 27 turns. Finally, the CDF in panel (d) indicates that the majority of conversations fall within the 10-to-40 turn range. This distribution confirms that AdaRPSet successfully balances concise, plot-driven interactions with longer, evolution-heavy role-playing sessions.

\subsection{Statistics of AdaSMSet}
\label{app:subsec:statistics_adasmset}

\textbf{AdaSMSet} is constructed directly from the \textbf{AdaRPSet-Synthesis} subset to provide supervised training data for the Scene Manager. Consequently, it shares the same foundational structure—9,900 plots and 9,900 conversations—as its source. However, because we explicitly insert Scene Manager control messages (most notably \texttt{pick\_speaker}) before every character turn, the total message count increases substantially. Specifically, AdaSMSet contains \textbf{496,493 utterances} in total, with an average length of \textbf{50.15 messages per conversation} (Median: 49.00, Std: 6.47), ranging from a minimum of 35 to a maximum of 85. This roughly doubles the density compared to the raw synthesis data, reflecting the fine-grained orchestration steps.

We further analyze the distribution of the five action types in the Scene Manager's output space. As designed, every conversation contains exactly one \texttt{init\_scene} at the start and one \texttt{end} action at the conclusion (9,900 instances each). The core dynamic actions show the following statistics:
\begin{itemize}[leftmargin=1.2em, labelsep=0.4em,
                itemsep=0pt, parsep=0pt, topsep=2pt, partopsep=0pt]
    \item \textbf{\texttt{pick\_speaker}}: By far the most frequent action, appearing \textbf{223,415 times} in total. On average, the manager performs speaker selection \textbf{22.57 times} per conversation (Median: 22.00, Range: 15--40), ensuring continuous turn-taking supervision.
    \item \textbf{\texttt{switch\_scene}}: There are \textbf{10,101 instances} of scene transitions, averaging \textbf{1.02 times} per conversation (Median: 1.00, Range: 1--2). This confirms that every dialogue involves at least one scene shift, with some containing multiple transitions.
    \item \textbf{\texttt{add\_role}}: Role introduction occurs \textbf{9,862 times}, averaging \textbf{1.00 times} per conversation (Median: 1.00, Range: 0--2). Approximately 99.5\% of conversations (9,847/9,900) feature at least one dynamic character entry, enforcing the requirement for adaptable cast management.
\end{itemize}
These statistics validate that AdaSMSet provides dense and balanced supervision for all critical adaptive behaviors—routine speaker selection, plot-driven scene changes, and dynamic cast expansion.

\section{Details of Evaluation Framework}
\label{app:sec:details_of_evaluation_framework}

\subsection{Actor Model Evaluation}
\label{app:subsec:actor_model_evaluation}

\paragraph{Scope and inputs.}
We evaluate the Actor Model $\mathcal{A}$ at the \emph{trajectory level} rather than the sentence level. Given a simulated dialogue trajectory $\tau=\{m_t\}_{t=1}^{T}$, the judge receives (i) the \textbf{Main Character Profile} (the target persona enacted by $\mathcal{A}$), (ii) \textbf{Other Characters} (profiles of the User and NPCs), and (iii) the full \textbf{Dialogue History} including Scene Manager messages (e.g., \texttt{init\_scene}, \texttt{switch\_scene}, \texttt{add\_role}). Importantly, the judge is instructed to score \textbf{only the Main Character turns generated by $\mathcal{A}$}; the writing quality of other speakers is ignored, and Scene Manager decisions are not judged except insofar as $\mathcal{A}$ fails to react to state changes.

\paragraph{Scoring philosophy.}
To make grading intentionally strict, we use a conservative, evidence-based $0$--$10$ rubric. The judge starts from \textbf{5 (baseline)} and raises the score only with \emph{explicit} textual evidence; partial or ambiguous evidence should not receive high scores, and ties are resolved by choosing the lower score. For every metric, the judge also provides brief, excerpt-grounded justifications from $\tau$. The full judge prompt is listed in Table~\ref{tab:actor_eval_prompt_part1} and ~\ref{tab:actor_eval_prompt_part2}.

\subsubsection*{I. Character Consistency}
\textbf{What this dimension evaluates.} Character Consistency assesses whether $\mathcal{A}$ consistently embodies the specified persona throughout the interaction, including alignment among internal thoughts, external actions, and spoken lines, as well as adherence to identity constraints and motivational drivers. It addresses the question: \emph{If character names were removed, would the output remain recognizably the same character?}

\begin{itemize}[leftmargin=1.2em, labelsep=0.4em,
                itemsep=0pt, parsep=0pt, topsep=2pt, partopsep=0pt]
    \item \textbf{Internal Coherence.} Assesses whether \texttt{[Thought]}, \texttt{(Action)}, and speech form a coherent causal loop, with thoughts motivating actions and actions supporting or anticipating speech, without unexplained contradictions.
    \item \textbf{Speaking Style Fidelity.} Assesses whether lexical choice, rhythm, tone, and stylistic markers remain faithful to the specified \texttt{speaking\_style}, avoiding drift into generic assistant-like language.
    \item \textbf{Language Fluency \& Human-likeness.} Assesses whether language is natural, context-appropriate, and non-repetitive, with utterance length and flow resembling human dialogue.
    \item \textbf{Identity \& Profile Fidelity.} Assesses whether knowledge, skills, and behaviors remain consistent with the character profile, penalizing out-of-profile expertise or background-inconsistent actions.
    \item \textbf{Motivation \& Value Stability.} Assesses whether stated motivations and core values persist and causally guide decisions across $\tau$, without abrupt or unmotivated shifts.
\end{itemize}

\subsubsection*{II. Environmental Grounding}
\textbf{What this dimension evaluates.} Environmental Grounding assesses whether $\mathcal{A}$ treats the scene as a binding world state rather than a passive backdrop, with correct perception of and interaction with the environment, including updates introduced by Scene Manager messages (e.g., initialization, scene switches, and newly introduced world facts).

\begin{itemize}[leftmargin=1.2em, labelsep=0.4em,
                itemsep=0pt, parsep=0pt, topsep=2pt, partopsep=0pt]
    \item \textbf{Environmental Awareness.} Assesses whether actions and perceptions are constrained by the current environment and its history, including prior events and \texttt{switch\_scene} updates, without violating physical or setting constraints.
    \item \textbf{Environmental Utilization.} Assesses whether environmental elements (objects, spatial relations, sensory cues) are meaningfully incorporated to support believable actions and advance the interaction, rather than serving as a mere backdrop.
\end{itemize}

\subsubsection*{III. Interpersonal Interaction}
\textbf{What this dimension evaluates.} Interpersonal Interaction assesses whether $\mathcal{A}$ engages in genuine turn-by-turn interaction by understanding others’ utterances and actions, maintaining appropriate relational stance, and adapting to newly introduced characters. It emphasizes \emph{listening}, \emph{responsive reply}, and \emph{social coherence}.

\begin{itemize}[leftmargin=1.2em, labelsep=0.4em,
                itemsep=0pt, parsep=0pt, topsep=2pt, partopsep=0pt]
    \item \textbf{Contextual Responsiveness.} Assesses how directly the Main Character’s reply builds on preceding turns, including content, implied intent, actions, and subtext, penalizing ignored questions, abrupt topic shifts, and non sequiturs.
    \item \textbf{Relationship Awareness.} Assesses whether attitudes and behaviors align with predefined relationships and updates appropriately as events unfold, including the correct recognition of new roles introduced by the Scene Manager.
\end{itemize}

\subsubsection*{IV. Narrative Progression}
\textbf{What this dimension evaluates.} Narrative Progression assesses long-context role-playing quality, focusing on whether the model advances the plot in an engaging way while maintaining continuity across many turns. Unlike Character Consistency, this dimension emphasizes \emph{trajectory-level dynamics} and \emph{long-horizon stability}.

\begin{itemize}[leftmargin=1.2em, labelsep=0.4em,
                itemsep=0pt, parsep=0pt, topsep=2pt, partopsep=0pt]
    \item \textbf{Narrative Attractiveness.} Assesses whether each Main Character turn contributes forward momentum through new information, actions, emotional development, or interaction hooks, penalizing repetition and static confirmations.
    \item \textbf{Stability Over Time.} Assesses whether the model maintains coherence over long interactions (e.g., $10+$ turns), including retention of established facts, avoidance of fabricated history, and resistance to style or persona drift.
\end{itemize}

\subsubsection*{V. Instruction Compliance}
\textbf{What this dimension evaluates.} Instruction Compliance serves as a strict technical gatekeeper, assessing whether $\mathcal{A}$ adheres to required output constraints and prohibitions in the role-playing format. Even minor violations (e.g., impersonating the User) may invalidate a trajectory for downstream use.

\begin{itemize}[leftmargin=1.2em, labelsep=0.4em,
                itemsep=0pt, parsep=0pt, topsep=2pt, partopsep=0pt]
    \item \textbf{Compliance \& Formatting.} Assesses adherence to all formatting and structural rules, including correct use of \texttt{[Thought]}, \texttt{(Action)}, and \texttt{<Environment>} tags, punctuation and length constraints, and the strict prohibition against generating content for other speakers (User, NPCs, or the Scene Manager).
\end{itemize}

\subsection{Scene Manager Evaluation}
\label{app:subsec:scene_manager_evaluation}

\paragraph{Scope and inputs.}
We evaluate the \textbf{Scene Manager} as a system-level orchestrator (scene/turn/role control), \textbf{not} as a writer. The judge must assess \emph{system decisions only} and ignore prose/dialogue quality, creativity, emotional impact, and character acting. For each trajectory $\tau=\{m_t\}_{t=1}^{T}$, the LLM-as-Judge is given $\mathcal{R}$ (profiles), motivations $\mathcal{M}$, the initial scene $\mathcal{E}_0$, and the full trajectory $\tau$ (including \texttt{init\_scene}, \texttt{switch\_scene}, \texttt{add\_role}).

\paragraph{Scoring philosophy.}
All axes use a strict, conservative $0$--$10$ scale where \textbf{5 is merely acceptable}; $7$--$8$ requires consistently good judgment with minor issues; $9$--$10$ requires exceptional discipline with no meaningful errors. Scores must be justified with concise, criterion-tied evidence. The full judge prompt and rubric are in Table~\ref{tab:scene_manager_rubric_core} and \ref{tab:scene_manager_rubric_output}.

\subsubsection*{Axis I: Scene Understanding (0--10)}
Evaluates correct management of scene state and transitions (this is the \textbf{core} axis).
Key checks: (i) track current scene theme/stakes/goals; (ii) distinguish major transitions vs. minor in-scene shifts; (iii) avoid premature \texttt{switch\_scene} when characters only propose/discuss moving; (iv) detect natural scene closure; (v) when switching, the new scene is clear, causally justified, and well-timed.

\subsubsection*{Axis II: Turn \& Speaker Selection Discipline (0--10)}
Evaluates turn-order fairness and structural discipline (focus on \textbf{user agency}, not dramatic effect).
Key checks: (i) no same speaker twice consecutively; (ii) avoid long NPC-only stretches; (iii) re-include the user within $\approx3$--$4$ turns; (iv) prevent role monopolization; (v) selected speaker is a valid role; (vi) speaker-selection reasons (if provided) are coherent and context-grounded.

\subsubsection*{Axis III: Role Introduction \& Utilization Judgment (0--10)}
Evaluates whether/when/why new roles are introduced via \texttt{add\_role}.
Key checks: (i) add roles only when interaction is needed; (ii) timing is appropriate; (iii) role function/profile/motivation meaningfully serves the scene/plot; (iv) rationale is specific and grounded; (v) penalize missed necessary introductions; (vi) avoid redundant or decorative roles.

\subsubsection*{Axis IV: Overall Assessment (0--10)}
A holistic score of orchestration quality; \textbf{not} a simple average. It reflects whether the three axes jointly support coherent pacing and user agency. Major failure in any axis should cap the overall score; no extra criteria should be introduced.

\section{Training Details}
\label{app:sec:training_details}


\paragraph{Data split.}
We hold out 5\% of the training data as a validation set and use the checkpoint with the lowest validation loss for all reported results.

\paragraph{Batching and parallelism.}
For 7B/8B models, we train on a single GPU; for 14B models, we use 4 GPUs; for 70B/72B models, we use 8 GPUs for parallel training. We set:
(i) for 7B/8B and 14B models: \texttt{micro\_batch\_size}=24, \texttt{global\_batch\_size}=48;
(ii) for 70B/72B models: \texttt{micro\_batch\_size}=1, \texttt{global\_batch\_size}=32.

\paragraph{Optimization setup.}
We train for 8 epochs with a fixed learning rate of $1\mathrm{e}{-6}$. We use a warm-up schedule over the first 5\% of total steps, linearly increasing from \texttt{min\_lr}$=1\mathrm{e}{-7}$ to the target learning rate.

\paragraph{Sequence length and truncation.}
We use a maximum sequence length of 16K tokens and apply left truncation when inputs exceed the context limit.

\section{\texttt{Basic} vs \texttt{Enhance} for Actor and Scene Manager}
\label{app:sec:prompt_templates}

This appendix reports the system prompts used for the Actor Model and the Scene Manager. We provide both \texttt{Basic} and \texttt{Enhance} variants; the latter introduces more detailed constraints and guidance.

For the Actor Model, the differences between \texttt{Basic} and \texttt{Enhance} prompts—particularly regarding the usage of role-playing annotations and adaptive constraints—are shown in Table~\ref{tab:actor_model_prompt}.  
For the Scene Manager, the \texttt{Basic} and \texttt{Enhance} prompts, which differ mainly in the level of action-selection guidance, are listed in Table~\ref{tab:scene_manager_prompt_basic} and Table~\ref{tab:scene_manager_prompt_enhanced}, respectively.

\section{Additional Evaluation Frameworks}
\label{app:sec:additional_evaluation_frameworks}
In addition to AdaptiveBench, we employ two established evaluation protocols, CharacterArena and CharacterBench, to assess model performance under alternative rubrics and prompt designs. 

\subsection{CharacterArena Evaluation}
\label{app:subsec:characterarena}
CharacterArena is a trajectory-level evaluation framework introduced by CPO~\citep{yeCPOAddressingReward2025}, which assesses role-playing quality via pairwise comparison (win-rate). Given a shared context (initial scene and character profiles) and two generated trajectories from different models (Model A vs. Model B), an LLM judge acts as an adjudicator to analyze the strengths and weaknesses of each trajectory, determine a winner, and provide a rationale.

We apply this protocol to the 100 evaluation seeds from our AdaptiveBench set. For each seed, we compare the trajectory generated by our Actor Model against those produced by baseline methods. The evaluation rubric consists of the following six dimensions:
\begin{itemize}[leftmargin=1.2em, labelsep=0.4em,
                itemsep=0pt, parsep=0pt, topsep=2pt, partopsep=0pt]
    \item \textbf{Plot Development.} Evaluates pacing and novelty, penalizing stagnation (lack of plot progression), excessive rushing (skipping key developments), or overreliance on clichés.
    \item \textbf{Dialogue Information Density.} Penalizes vague, abstract, or overly generalized responses, as well as preachy content that fails to contribute substantively to the narrative.
    \item \textbf{Dialogue Immersion.} Assesses the ability to construct vivid, three-dimensional scenes through sensory details (e.g., sight, sound) and fine-grained actions or micro-expressions that evoke emotional or physiological responses.
    \item \textbf{Storyline.} Values advanced narrative techniques such as foreshadowing, reversals, suspense, and unexpected turns, favoring complex and non-linear developments over predictable progressions.
    \item \textbf{Interactivity.} Measures intent understanding (avoiding misinterpretation or talking past the user), avoidance of repetition, and proactive engagement strategies (e.g., asking questions or guiding the interaction).
    \item \textbf{Dialogue Coherence.} Checks linguistic fluency and logical consistency, penalizing contradictions in time, location, facts, or character perspective, as well as abrupt topic shifts or out-of-character behavior.
\end{itemize}

For the exact prompts and detailed scoring criteria used in CharacterArena, we refer readers to the original paper~\citep{yeCPOAddressingReward2025}.

\subsection{CharacterBench Evaluation}
\label{app:subsec:characterbench}
CharacterBench~\citep{zhouCharacterBenchBenchmarkingCharacter2024} is a comprehensive evaluation framework that assesses character authenticity grounded in interpersonal interaction theory. Unlike our trajectory-level AdaptiveBench, CharacterBench focuses on single-turn response quality given a specific profile $\mathcal{P}$, context $\mathcal{C}$, and user query $u_n$.

We utilize the official open-source implementation and evaluator models of CharacterBench to verify whether our AdaMARP framework—trained on our proposed AdaRPSet—generalizes well to external benchmarks beyond our own simulation setting. CharacterBench yields 13 specific metrics across 6 aspects:

\begin{itemize}[leftmargin=1.2em, labelsep=0.4em,
                itemsep=0pt, parsep=0pt, topsep=2pt, partopsep=0pt]
    \item \textbf{Memory:} Assessed via \textbf{Memory Consistency} (\textbb{MC}), measuring whether the character’s response aligns with facts and events established in the dialogue history.
    \item \textbf{Knowledge:} Includes \textbf{Fact Accuracy} (\textbb{FA})—the correctness of self-related facts—and \textbf{Boundary Consistency} (\textbb{BC}$_K$), which evaluates adherence to the knowledge constraints of the character’s world.
    \item \textbf{Persona:} Evaluated through \textbf{Attribute Consistency} and \textbf{Behavior Consistency}. These are further divided into consistency with the bot’s own profile (\textbb{AC}$^b$, \textbb{BC}$_P^b$) and consistency with the human interlocutor’s persona or expectations (\textbb{AC}$^h$, \textbb{BC}$_P^h$).
    \item \textbf{Emotion:} Covers \textbf{Emotional Self-regulation} (\textbb{ES}), which reflects the ability to manage the character’s own emotions, and \textbf{Empathetic Responsiveness} (\textbb{ER}), which measures recognizing and responding appropriately to user emotions.
    \item \textbf{Morality:} Examines \textbf{Morality Stability} (\textbb{MS}), i.e., resistance to toxic or adversarial user inputs, and \textbf{Morality Robustness} (\textbb{MR}), i.e., maintaining ethical and safety standards even when the character profile contains toxic traits.
    \item \textbf{Believability:} Split into \textbf{Human-likeness} (\textbb{HL}), reflecting the naturalness of responses, and \textbf{Engagement} (\textbb{EG}), capturing the ability to sustain user interest and emotional connection.
\end{itemize}

By evaluating on CharacterBench, we aim to demonstrate that the Actor Model’s improvements in consistency and immersion are robust and transferable to diverse role-playing scenarios and metric definitions.

\subsection{Human Evaluation}
\label{app:subsec:human_evaluation}

To further mitigate potential evaluation bias introduced by relying solely on AI-based judge models, we additionally conduct a human evaluation on AdaptiveBench. Since it is difficult for human annotators to reliably score all 12 fine-grained dimensions, we simplify the protocol to a pairwise preference setting. After reading the official evaluation guidelines of the judge model, human evaluators are asked to independently compare two trajectories generated under the same Profiles, Motivations, and initial scenario, and decide which actor model better portrays the \emph{main character} (Model A vs.\ Model B).

We recruit three independent human evaluators and treat their judgments as separate samples. Instead of aggregating by majority vote, we compute the win rate for each evaluator individually and then report the mean and variance across evaluators. To avoid obvious identification bias, we do not include Crab, CoSER, or BeyondDialogue in this study: Crab and CoSER outputs lack either \emph{thought} or \emph{environment} fields, while BeyondDialogue only contains \emph{speech}, making the generating model easily identifiable. Therefore, we focus on a controlled comparison between \textbf{Qwen2.5-7B-Instruct} before and after training with AdaRPSet.

The results show that the post-training model achieves an average win rate of 80.33\% with a variance of 6.22, indicating a consistent human-perceived improvement after training. Although this win rate is lower than that obtained from AI-based evaluation on CharacterArena (94\%; Table~\ref{tab:characterarena_qwen}) using their official prompts, the overall trend is consistent: the model trained with AdaRPSet is clearly preferred by human evaluators.

\section{Results on Additional Evaluation Frameworks}
\label{app:sec:results_on_additional_frameworks}

\subsection{CharacterArena Results}
\label{app:subsec:characterarena_results}
We conduct pairwise win-rate evaluations using the CharacterArena protocol (Appendix~\ref{app:subsec:characterarena}) on two model families: Qwen2.5-7B-Instruct and Llama-3.1-8B-Instruct.

\paragraph{Qwen2.5-7B-Instruct Series.}
Table~\ref{tab:characterarena_qwen} presents the win-rate matrix for the Qwen2.5-7B series. Here, ``Base'' refers to the off-the-shelf Qwen2.5-7B-Instruct model. Our method (\textbf{Ours}) consistently dominates other baselines, achieving over 80\% win-rate against Base (94\%), BeyondDialogue (100\%), CoSER (99\%), and Crab (84\%). These results confirm that our training data and framework yield superior trajectory-level performance even under an independent pairwise judging mechanism.

\paragraph{Llama-3.1-8B-Instruct Series.}
Table~\ref{tab:characterarena_llama} shows the results for the Llama-3.1-8B series. We exclude the BeyondDialogue variant here, as the model trained on its relatively small dataset ($\sim$3k dialogues, $\sim$800 English) struggled to follow instructions reliably. Our method again significantly outperforms both Crab and CoSER, further validating the robustness of our approach across different backbone architectures.

\begin{table}[h]
\centering
\small
\setlength{\tabcolsep}{4pt}
\begin{tabular}{l c c c c c}
\toprule
Model A $\downarrow$ vs B $\rightarrow$ & Base & Beyond & CoSER & Crab & \textbf{Ours} \\
\midrule
Base   & -- & 93\% & 77\% & 88\% & 6\% \\
Beyond & 7\% & -- & 28\% & 48\% & 0\% \\
CoSER  & 23\% & 72\% & -- & 65\% & 1\% \\
Crab   & 12\% & 52\% & 35\% & -- & 16\% \\
\textbf{Ours}   & \textbf{94\%} & \textbf{100\%} & \textbf{99\%} & \textbf{84\%} & -- \\
\bottomrule
\end{tabular}
\caption{Pairwise win-rates on \textbf{CharacterArena} (Model A wins against Model B) for the\textit{ Qwen2.5-7B series}. ``Base'' denotes Qwen2.5-7B-Instruct.}
\label{tab:characterarena_qwen}
\end{table}

\begin{table}[h]
\centering
\small
\begin{tabular}{l c c c}
\toprule
Model A $\downarrow$ vs B $\rightarrow$ & Crab & CoSER & \textbf{Ours} \\
\midrule
Crab  & -- & 52\% & 2\% \\
CoSER & 48\% & -- & 1\% \\
\textbf{Ours}  & \textbf{98\%} & \textbf{99\%} & -- \\
\bottomrule
\end{tabular}
\caption{Pairwise win-rates on \textbf{CharacterArena} for the \textit{Llama-3.1-8B} series.}
\label{tab:characterarena_llama}
\end{table}

\subsection{CharacterBench Results}
\label{app:subsec:characterbench_results}

We evaluate model performance on CharacterBench (Appendix~\ref{app:subsec:characterbench}) using the official judge model provided by the benchmark. Table~\ref{tab:actor_model_characterbench_result_main} reports the results for the Qwen2.5-7B-Instruct series and the Llama-3.1-8B base models.

For Qwen2.5-7B-Instruct, training with existing role-playing baselines such as Beyond, Crab, or CoSER generally degrades overall role-playing quality compared to the base instruct model. Unlike the results on AdaptiveBench, Crab performs notably better on CharacterBench than the other baselines, achieving top or near-top scores on several sub-metrics, while Beyond and CoSER exhibit improvements only in isolated dimensions. Nevertheless, Crab still falls short of our approach. Our method (\textbf{Ours}) achieves the highest average score (3.68), demonstrating that the capabilities learned from AdaRPSet transfer effectively to an external evaluation framework. In particular, our model ranks first or second on most critical dimensions, including Attribute Consistency (Human), Behavior Consistency (Bot), Human-likeness, and Engagement, highlighting its robustness in maintaining persona fidelity and interaction quality beyond our simulated environment.

For Llama-3.1-8B, however, directly training the base model with AdaRPSet results in slightly worse overall performance than CoSER (3.52 vs.\ 3.53), while Crab achieves the best results. We attribute this gap primarily to the presence of large amounts of general instruction-following data in Crab and CoSER, which is absent from AdaRPSet. To verify this hypothesis, we conduct an additional ablation by incorporating general instruction-tuning data as a regularizer. Following CoSER, we randomly sample 22,425 instances from the Tulu3~\citep{lambert2025tulu3pushingfrontiers} dataset (matching the size of AdaRPSet) and jointly train the model. This variant (\textbf{Ours+General}) improves the overall average score by 0.05.

Furthermore, as shown in Table~\ref{tab:adaptivebench_llama_general} (AdaptiveBench results), \textbf{Llama-3.1-8B-Ours+General} exhibits only a minor performance drop on AdaptiveBench (8.89 $\rightarrow$ 8.80), with improvements on the Instruction Compliance metric, and still outperforms all competing methods. These results suggest that, for base models without prior instruction tuning, incorporating a small amount of general-purpose instruction data as regularization is beneficial and recommended.

\begin{table*}[t]
\centering
\resizebox{\textwidth}{!}{
\begin{tabular}{l c c c c c c c c c c c c c c}
\hline
\multicolumn{15}{l}{\makecell[l]{
\textbb{MC}: Memory Consistency ~~~
\textbb{FA}: Fact Accuracy ~~~
\textbb{BC$_K$}: Boundary Consistency ~~~
\textbb{AC$^b$}: Attribute Consistency (Bot) ~~~
\textbb{AC$^h$}: Attribute Consistency (Human) \\
\textbb{BC$_P^b$}: Behavior Consistency (Bot) ~~~
\textbb{BC$_P^h$}: Behavior Consistency (Human) ~~~
\textbb{ES}: Emotional Self-regulation ~~~
\textbb{ER}: Empathetic Responsiveness \\
\textbb{MS}: Morality Stability ~~~
\textbb{MR}: Morality Robustness ~~~
\textbb{HL}: Human-likeness ~~~
\textbb{EG}: Engagement
}}\\
\hline

\multirow{2}{*}{Models}
& \multicolumn{1}{c}{Memory}
& \multicolumn{2}{c}{Knowledge}
& \multicolumn{4}{c}{Persona}
& \multicolumn{2}{c}{Emotion}
& \multicolumn{2}{c}{Morality}
& \multicolumn{2}{c}{Believability}
& \multirow{2}{*}{Average} \\
\cmidrule(lr){2-2}
\cmidrule(lr){3-4}
\cmidrule(lr){5-8}
\cmidrule(lr){9-10}
\cmidrule(lr){11-12}
\cmidrule(lr){13-14}
& \textbb{MC}
& \textbb{FA} & \textbb{BC$_K$}
& \textbb{AC$^b$} & \textbb{AC$^h$} & \textbb{BC$_P^b$} & \textbb{BC$_P^h$}
& \textbb{ES} & \textbb{ER}
& \textbb{MS} & \textbb{MR}
& \textbb{HL} & \textbb{EG}
& \\
\hline

Qwen2-7B-BD-RP*
& 3.43 & 2.42 & 3.37 & 3.33 & 3.44 & 2.97 & 3.16 & 2.80 & 2.65 & 4.84 & 4.81 & 2.25 & 2.33 & 3.22 \\

\hline
\multicolumn{15}{l}{\textbf{Qwen2.5-7B-Instruct-Series}} \\
\hline

Qwen2.5-7B-Instruct
& 3.64 & \textbf{2.37} & 3.53 & 4.57 & 4.01 & 3.66 & \textbf{3.52} & 3.16 & \textbf{3.07} & \textbf{4.89} & 4.85 & \textbf{3.00} & 3.06 & 3.64 \\

Qwen2.5-7B-Instruct-BeyondDialogue
& 3.63 & 2.15 & \textbf{3.82} & 4.21 & 3.82 & 3.28 & 3.22 & 2.93 & 2.69 & 4.62 & 4.66 & 2.72 & 2.75 & 3.43 \\

Qwen2.5-7B-Instruct-Crab
& \textbf{4.13} & 2.25 & 3.11 & \textbf{4.67} & 3.92 & 3.65 & 3.23 & \textbf{3.27} & 2.95 & 4.84 & \textbf{4.87} & 2.82 & 2.93 & 3.59 \\

Qwen2.5-7B-Instruct-CoSER
& 4.06 & 2.28 & 3.74 & 4.52 & 3.93 & 3.42 & 3.23 & 3.21 & 2.79 & 4.76 & 4.63 & 2.68 & 2.78 & 3.54 \\

\textbf{Qwen2.5-7B-Instruct-Ours}
& 4.03 & 2.35 & 3.65 & 4.63 & \textbf{4.14} & \textbf{3.75} & 3.45 & 3.12 & 3.02 & 4.80 & 4.79 & \textbf{3.00} & \textbf{3.07} & \textbf{3.68} \\

\hdashline
\multicolumn{15}{l}{\textbf{Llama-3.1-8B-Series}} \\
\hdashline

Llama-3.1-8B-Crab
& \underline{3.83} & \underline{2.23} & 3.54 & \textbf{4.47} & \textbf{4.17} & \underline{3.53} & \underline{3.20} & \textbf{3.32} & \underline{2.89} & \underline{4.78} & \underline{4.78} & \textbf{3.02} & \underline{3.00} & \textbf{3.60} \\
Llama-3.1-8B-CoSER*
& 3.75 & 2.19 & \underline{3.74} & 4.19 & 4.01 & 3.51 & \textbf{3.39} & 2.97 & \textbf{2.94} & \textbf{4.92} & \textbf{4.97} & 2.70 & 2.62 & 3.53 \\
\textbf{Llama-3.1-8B-Ours}
& 3.86 & 2.15 & \textbf{3.77} & 4.23 & 4.03 & 3.42 & 3.13 & 3.13 & 2.79 & 4.59 & 4.69 & 2.89 & \textbf{3.05} & 3.52 \\
\textbf{Llama-3.1-8B-Ours+General}
& \textbf{3.92} & \textbf{2.30} & 3.69 & \textbf{4.37} & \underline{4.11} & \textbf{3.58} & 3.13 & \textbf{3.21} & 2.88 & 4.74 & 4.73 & \underline{2.92} & 2.84 & \underline{3.57} \\

\hline
\end{tabular}}
\caption{Actor model evaluation results on \textbf{CharacterBench}, where the judge model is their open-sourced evaluation model.
\textbf{Bold} indicates the best performance within the same model scale, and \underline{underline} indicates the second-best performance within the same scale.
Models marked with * are evaluated using their officially released checkpoints without additional training, while unmarked baselines are re-trained under our experimental setup.
\textbf{General} denotes the variant where we additionally incorporate instruction-tuning data of the same scale as AdaRPSet during training.}
\label{tab:actor_model_characterbench_result_main}
\end{table*}

\begin{table*}[t]
\centering
\resizebox{\textwidth}{!}{
\begin{tabular}{l c c c c c c c c c c c c c}
\hline
\multicolumn{14}{l}{\makecell[l]{
\textbb{ICoh}: Internal Coherence ~~~
\textbb{SSF}: Speaking Style Fidelity ~~~
\textbb{LFH}: Language Fluency \& Human-likeness ~~~
\textbb{IPF}: Identity \& Profile Fidelity ~~~
\textbb{MVS}: Motivation \& Value Stability \\
\textbb{EA}: Environmental Awareness ~~~
\textbb{EU}: Environmental Utilization ~~~
\textbb{CR}: Contextual Responsiveness ~~~
\textbb{RA}: Relationship Awareness ~~~
\textbb{ATT}: Attractiveness ~~~
\textbb{STB}: Stability ~~~
\textbb{IC}: Instruction Compliance
}} \\
\hline
\multirow{2}{*}{\textbb{Models}}
& \multicolumn{5}{c}{\textbb{CC}}
& \multicolumn{2}{c}{\textbb{EG}}
& \multicolumn{2}{c}{\textbb{II}}
& \multicolumn{2}{c}{\textbb{NP}}
& \multicolumn{1}{c}{\textbb{IC}}
& \multirow{2}{*}{\textbb{Average}} \\
\cmidrule(lr){2-6}
\cmidrule(lr){7-8}
\cmidrule(lr){9-10}
\cmidrule(lr){11-12}
\cmidrule(lr){13-13}
& \textbb{ICoh} & \textbb{SSF} & \textbb{LFH} & \textbb{IPF} & \textbb{MVS}
& \textbb{EA} & \textbb{EU}
& \textbb{CR} & \textbb{RA}
& \textbb{ATT} & \textbb{STB}
& \textbb{IC}
&  \\
\hline
\textbf{Llama-3.1-8B-Ours}
& 9.00$\pm$0.00 & 8.39$\pm$0.49 & 8.93$\pm$0.26 & 9.69$\pm$0.48 & 9.08$\pm$0.63
& 8.77$\pm$0.49 & 7.84$\pm$0.44
& 9.26$\pm$0.46 & 8.36$\pm$0.52
& 8.52$\pm$0.50 & 9.39$\pm$0.49
& 9.49$\pm$0.54
& 8.89 \\
\textbf{Llama-3.1-8B-Ours+General}
& 8.99$\pm$0.10 & 8.25$\pm$0.45 & 8.93$\pm$0.26 & 9.52$\pm$0.50 & 8.95$\pm$0.54
& 8.58$\pm$0.51 & 7.66$\pm$0.50
& 9.18$\pm$0.41 & 8.26$\pm$0.46
& 8.37$\pm$0.48 & 9.24$\pm$0.43
& \textbf{9.66}$\pm$0.47
& 8.80 \\
\hline
\end{tabular}}
\caption{Comparison between \textbf{Llama-3.1-8B-Ours} and \textbf{Llama-3.1-8B-Ours+General} on \textbf{AdaptiveBench}.
Incorporating general instruction-tuning data improves Instruction Compliance (\textbb{IC}) by +0.17, while several other dimensions exhibit minor and acceptable declines. $\pm$ denotes the standard deviation computed over all samples.}
\label{tab:adaptivebench_llama_general}
\end{table*}

\section{Ablation Analysis of Actor-Model Evaluation}
\label{app:actor_eval_ablation}

We study the robustness of Actor-Model evaluation on AdaptiveBench by varying the \emph{judge model}. Specifically, we use four different LLMs as judges to score trajectories under the same evaluation protocol and prompts: GPT-4o-mini, GPT-5-Chat, Gemini-2.5-Pro, and Doubao-1.5-Pro-Character. Table~\ref{ablation_adaptivebench} reports the detailed metric scores, while Table~\ref{rank_basic} and Table~\ref{rank_enhance} summarize the induced rankings under \texttt{Basic} and \texttt{Enhance} prompting variants, respectively.

Overall, the rankings are highly consistent across judge models. As shown in Table~\ref{rank_basic} and Table~\ref{rank_enhance}, three judges (GPT-5-Chat, Gemini-2.5-Pro, and Doubao-1.5-Pro-Character) produce almost identical ordering of the evaluated models, indicating that our conclusions are not sensitive to a particular judge choice. The main discrepancy comes from GPT-4o-mini, which yields a slightly different ordering in some cases (e.g., swapping the relative positions among middle-tier models). 

Given the above, we adopt GPT-5-Chat as the default judge in the main paper: it provides stable and consistent rankings aligned with other strong judges, while serving as a unified evaluation standard for reporting results.

\begin{table*}[t]
\centering
\resizebox{\textwidth}{!}{
\begin{tabular}{l c c c c c c c c c c c c c}
\hline
\multicolumn{14}{l}{\makecell[l]{
\textbb{ICoh}: Internal Coherence ~~~
\textbb{SSF}: Speaking Style Fidelity ~~~
\textbb{LFH}: Language Fluency \& Human-likeness ~~~
\textbb{IPF}: Identity \& Profile Fidelity ~~~
\textbb{MVS}: Motivation \& Value Stability \\
\textbb{EA}: Environmental Awareness ~~~
\textbb{EU}: Environmental Utilization ~~~
\textbb{CR}: Contextual Responsiveness ~~~
\textbb{RA}: Relationship Awareness ~~~
\textbb{ATT}: Attractiveness ~~~
\textbb{STB}: Stability ~~~
\textbb{IC}: Instruction Compliance \\
\textbb{CC}: Character Consistency ~~~
\textbb{EG}: Environmental Grounding ~~~
\textbb{II}: Interpersonal Interaction ~~~
\textbb{NP}: Narrative Progression
}} \\
\hline
\multirow{2}{*}{\textbb{Models}}
& \multicolumn{5}{c}{\textbb{CC}}
& \multicolumn{2}{c}{\textbb{EG}}
& \multicolumn{2}{c}{\textbb{II}}
& \multicolumn{2}{c}{\textbb{NP}}
& \multicolumn{1}{c}{\textbb{IC}}
& \multirow{2}{*}{\textbb{Average}} \\
\cmidrule(lr){2-6}
\cmidrule(lr){7-8}
\cmidrule(lr){9-10}
\cmidrule(lr){11-12}
\cmidrule(lr){13-13}
& \textbb{ICoh} & \textbb{SSF} & \textbb{LFH} & \textbb{IPF} & \textbb{MVS}
& \textbb{EA} & \textbb{EU}
& \textbb{CR} & \textbb{RA}
& \textbb{ATT} & \textbb{STB}
& \textbb{IC}
&  \\
\hline

\multicolumn{14}{l}{\textbf{Ablation Results}} \\
\hline

\multicolumn{14}{l}{\textbf{Judge: GPT-4o-mini}} \\
\hline
GPT-4o-mini-Basic
& 8.05$\pm$0.26 & 7.55$\pm$0.50 & 7.79$\pm$0.57 & 8.96$\pm$0.24 & 8.00$\pm$0.37
& 8.31$\pm$0.54 & 7.46$\pm$0.54
& 8.44$\pm$0.50 & 7.88$\pm$0.57
& 8.31$\pm$0.46 & 8.24$\pm$0.67
& 9.28$\pm$0.51
& 8.19 \\

GPT-4o-mini
& 8.03$\pm$0.26 & 7.50$\pm$0.54 & 7.82$\pm$0.54 & 8.97$\pm$0.22 & 7.96$\pm$0.37
& 8.25$\pm$0.55 & 7.31$\pm$0.54
& 8.36$\pm$0.48 & 7.84$\pm$0.61
& 8.16$\pm$0.46 & 8.20$\pm$0.69
& 9.19$\pm$0.44
& 8.13 \\

GPT-5-Chat-Basic
& 8.17$\pm$0.40 & 7.86$\pm$0.49 & 7.86$\pm$0.68 & 8.96$\pm$0.34 & 8.15$\pm$0.43
& 8.34$\pm$0.57 & 7.52$\pm$0.59
& 8.52$\pm$0.50 & 7.88$\pm$0.50
& 8.48$\pm$0.50 & 8.18$\pm$0.59
& 9.32$\pm$0.49
& 8.27 \\

GPT-5-Chat
& 8.04$\pm$0.24 & 7.69$\pm$0.59 & 7.68$\pm$0.66 & 8.95$\pm$0.38 & 7.95$\pm$0.41
& 8.33$\pm$0.65 & 7.50$\pm$0.62
& 8.27$\pm$0.51 & 7.72$\pm$0.63
& 8.31$\pm$0.54 & 8.14$\pm$0.71
& 9.28$\pm$0.49
& 8.15 \\

Gemini-2.5-Pro-Basic
& 8.22$\pm$0.41 & 7.59$\pm$0.49 & 7.85$\pm$0.52 & 9.07$\pm$0.26 & 8.17$\pm$0.40
& 8.17$\pm$0.40 & 7.40$\pm$0.60
& 8.47$\pm$0.52 & 8.03$\pm$0.56
& 8.48$\pm$0.50 & 8.02$\pm$0.57
& 9.32$\pm$0.47
& 8.23 \\

Gemini-2.5
& 7.97$\pm$0.41 & 7.57$\pm$0.60 & 7.57$\pm$0.55 & 8.92$\pm$0.46 & 8.02$\pm$0.53
& 8.04$\pm$0.62 & 7.17$\pm$0.68
& 8.30$\pm$0.57 & 7.87$\pm$0.72
& 8.27$\pm$0.58 & 7.88$\pm$0.70
& 9.22$\pm$0.50
& 8.07 \\

Claude Sonnet 4.5-Basic
& 8.32$\pm$0.47 & 7.84$\pm$0.42 & 7.96$\pm$0.68 & 9.04$\pm$0.34 & 8.27$\pm$0.49
& 8.36$\pm$0.54 & 7.62$\pm$0.63
& 8.64$\pm$0.50 & 8.10$\pm$0.54
& 8.75$\pm$0.43 & 8.24$\pm$0.51
& 9.16$\pm$0.39
& 8.36 \\

Claude Sonnet 4.5
& 8.14$\pm$0.40 & 7.68$\pm$0.53 & 7.70$\pm$0.61 & 8.97$\pm$0.36 & 8.12$\pm$0.50
& 8.27$\pm$0.61 & 7.33$\pm$0.65
& 8.39$\pm$0.51 & 7.97$\pm$0.61
& 8.49$\pm$0.57 & 8.11$\pm$0.66
& 9.25$\pm$0.46
& 8.20 \\

Doubao-1-5-Pro-Character-Basic
& 7.87$\pm$0.39 & 7.23$\pm$0.49 & 7.81$\pm$0.46 & 8.85$\pm$0.38 & 7.88$\pm$0.43
& 7.88$\pm$0.60 & 6.92$\pm$0.58
& 8.08$\pm$0.42 & 7.61$\pm$0.68
& 8.03$\pm$0.41 & 7.87$\pm$0.73
& 9.25$\pm$0.43
& 7.94 \\

Doubao-1-5-Pro-Character
& 7.86$\pm$0.37 & 7.09$\pm$0.47 & 7.84$\pm$0.44 & 8.83$\pm$0.43 & 7.82$\pm$0.46
& 7.85$\pm$0.67 & 6.85$\pm$0.75
& 8.13$\pm$0.46 & 7.51$\pm$0.69
& 7.96$\pm$0.47 & 7.78$\pm$0.90
& 9.20$\pm$0.49
& 7.89 \\

\hdashline
\multicolumn{14}{l}{\textbf{Judge: GPT-5-Chat}} \\
\hline
GPT-4o-mini-Basic
& 9.00$\pm$0.14 & 8.29$\pm$0.46 & 8.93$\pm$0.26 & 9.57$\pm$0.50 & 8.99$\pm$0.58
& 8.93$\pm$0.48 & 8.02$\pm$0.45
& 9.35$\pm$0.48 & 8.38$\pm$0.51
& 8.41$\pm$0.49 & 9.42$\pm$0.55
& 9.60$\pm$0.51
& 8.91 \\

GPT-4o-mini
& 8.99$\pm$0.10 & 8.19$\pm$0.39 & 8.89$\pm$0.31 & 9.50$\pm$0.50 & 8.88$\pm$0.57
& 8.71$\pm$0.48 & 7.83$\pm$0.45
& 9.20$\pm$0.40 & 8.15$\pm$0.36
& 8.17$\pm$0.43 & 9.24$\pm$0.43
& 9.64$\pm$0.56
& 8.78 \\

GPT-5-Chat-Basic
& 9.39$\pm$0.51 & 8.97$\pm$0.17 & 9.11$\pm$0.40 & 9.96$\pm$0.20 & 9.17$\pm$0.53
& 9.28$\pm$0.57 & 8.63$\pm$0.66
& 9.66$\pm$0.47 & 8.71$\pm$0.45
& 9.19$\pm$0.42 & 9.69$\pm$0.52
& 9.65$\pm$0.48
& 9.28 \\

GPT-5-Chat
& 9.09$\pm$0.99 & 8.84$\pm$0.92 & 8.94$\pm$0.95 & 9.85$\pm$1.01 & 9.03$\pm$1.04
& 9.03$\pm$1.03 & 8.39$\pm$1.05
& 9.43$\pm$1.07 & 8.52$\pm$1.00
& 8.90$\pm$0.97 & 9.47$\pm$1.08
& 9.70$\pm$0.46
& 9.10 \\

Gemini-2.5-Pro-Basic
& 9.60$\pm$0.49 & 8.95$\pm$0.22 & 9.10$\pm$0.46 & 9.95$\pm$0.22 & 9.41$\pm$0.53
& 9.17$\pm$0.64 & 8.53$\pm$0.69
& 9.77$\pm$0.42 & 8.86$\pm$0.38
& 9.39$\pm$0.51 & 9.76$\pm$0.43
& 9.28$\pm$0.51
& 9.31 \\

Gemini-2.5
& 9.09$\pm$0.63 & 8.76$\pm$0.71 & 8.91$\pm$0.45 & 9.82$\pm$0.75 & 8.99$\pm$0.69
& 8.82$\pm$0.71 & 8.15$\pm$0.79
& 9.40$\pm$0.68 & 8.55$\pm$0.68
& 8.98$\pm$0.63 & 9.33$\pm$0.81
& 9.30$\pm$0.91
& 9.01 \\

Claude Sonnet 4.5-Basic
& 9.59$\pm$1.07 & 8.93$\pm$0.92 & 9.11$\pm$1.01 & 9.85$\pm$1.01 & 9.44$\pm$1.10
& 9.21$\pm$1.05 & 8.60$\pm$1.07
& 9.80$\pm$1.03 & 8.99$\pm$0.97
& 9.49$\pm$1.08 & 9.78$\pm$1.04
& 9.46$\pm$0.50
& 9.35 \\

Claude Sonnet 4.5
& 9.35$\pm$0.54 & 8.97$\pm$0.22 & 8.99$\pm$0.41 & 9.94$\pm$0.28 & 9.27$\pm$0.66
& 9.13$\pm$0.58 & 8.39$\pm$0.63
& 9.78$\pm$0.44 & 8.90$\pm$0.48
& 9.23$\pm$0.53 & 9.65$\pm$0.65
& 9.61$\pm$0.49
& 9.27 \\

Doubao-1-5-Pro-Character-Basic
& 8.58$\pm$0.49 & 7.79$\pm$0.46 & 8.70$\pm$0.46 & 9.12$\pm$0.46 & 8.43$\pm$0.61
& 8.33$\pm$0.62 & 7.32$\pm$0.62
& 8.93$\pm$0.33 & 8.02$\pm$0.40
& 7.89$\pm$0.51 & 8.93$\pm$0.41
& 9.27$\pm$0.75
& 8.44 \\

Doubao-1-5-Pro-Character
& 8.65$\pm$0.48 & 7.82$\pm$0.48 & 8.74$\pm$0.46 & 9.17$\pm$0.53 & 8.42$\pm$0.71
& 8.21$\pm$0.65 & 7.26$\pm$0.67
& 8.91$\pm$0.40 & 7.86$\pm$0.53
& 7.91$\pm$0.49 & 8.86$\pm$0.55
& 9.28$\pm$0.71
& 8.42 \\

\hdashline
\multicolumn{14}{l}{\textbf{Judge: Gemini-2.5-Pro}} \\
\hline
GPT-4o-mini-Basic
& 9.01$\pm$0.71 & 8.35$\pm$1.12 & 8.37$\pm$0.86 & 9.15$\pm$0.79 & 9.29$\pm$0.83
& 8.26$\pm$0.94 & 7.79$\pm$1.19
& 9.23$\pm$0.81 & 8.63$\pm$1.13
& 8.41$\pm$0.97 & 8.97$\pm$1.04
& 5.75$\pm$3.70
& 8.43 \\

GPT-4o-mini
& 9.07$\pm$0.87 & 8.39$\pm$0.97 & 8.24$\pm$0.98 & 9.13$\pm$0.83 & 9.28$\pm$0.80
& 8.36$\pm$1.14 & 7.60$\pm$1.33
& 9.19$\pm$0.91 & 8.48$\pm$1.01
& 8.25$\pm$0.99 & 8.68$\pm$1.48
& 7.66$\pm$3.40
& 8.53 \\

GPT-5-Chat-Basic
& 9.84$\pm$0.37 & 9.63$\pm$0.59 & 9.50$\pm$0.56 & 9.93$\pm$0.32 & 9.81$\pm$0.42
& 9.18$\pm$0.94 & 9.39$\pm$0.80
& 9.75$\pm$0.78 & 9.31$\pm$0.78
& 9.73$\pm$0.47 & 9.59$\pm$0.90
& 6.70$\pm$3.51
& 9.36 \\

GPT-5-Chat
& 9.69$\pm$1.06 & 9.48$\pm$1.14 & 9.39$\pm$1.09 & 9.85$\pm$1.01 & 9.78$\pm$1.04
& 9.20$\pm$1.18 & 9.15$\pm$1.16
& 9.74$\pm$1.05 & 9.28$\pm$1.18
& 9.54$\pm$1.07 & 9.47$\pm$1.16
& 7.33$\pm$3.44
& 9.32 \\

Gemini-2.5-Pro-Basic
& 9.96$\pm$0.20 & 9.82$\pm$0.48 & 9.62$\pm$0.51 & 9.97$\pm$0.17 & 9.91$\pm$0.32
& 9.17$\pm$0.87 & 9.37$\pm$0.73
& 9.91$\pm$0.32 & 9.70$\pm$0.54
& 9.91$\pm$0.29 & 9.82$\pm$0.64
& 6.52$\pm$3.68
& 9.47 \\

Gemini-2.5
& 9.65$\pm$1.09 & 9.44$\pm$1.14 & 9.21$\pm$1.13 & 9.76$\pm$1.04 & 9.65$\pm$1.11
& 8.60$\pm$1.24 & 8.74$\pm$1.38
& 9.64$\pm$1.18 & 9.42$\pm$1.23
& 9.60$\pm$1.09 & 9.15$\pm$1.41
& 6.58$\pm$3.61
& 9.12 \\

Claude Sonnet 4.5-Basic
& 9.82$\pm$1.02 & 9.70$\pm$1.06 & 9.72$\pm$1.06 & 9.90$\pm$0.99 & 9.88$\pm$1.00
& 9.26$\pm$1.20 & 9.23$\pm$1.22
& 9.84$\pm$1.02 & 9.77$\pm$1.05
& 9.85$\pm$1.01 & 9.82$\pm$1.04
& 7.36$\pm$3.48
& 9.51 \\

Claude Sonnet 4.5
& 9.82$\pm$1.03 & 9.66$\pm$1.14 & 9.58$\pm$1.08 & 9.90$\pm$1.00 & 9.83$\pm$1.03
& 9.24$\pm$1.21 & 9.17$\pm$1.19
& 9.83$\pm$1.03 & 9.68$\pm$1.09
& 9.80$\pm$1.03 & 9.76$\pm$1.09
& 9.07$\pm$2.17
& 9.61 \\

Doubao-1-5-Pro-Character-Basic
& 8.14$\pm$1.51 & 7.26$\pm$1.74 & 7.83$\pm$1.45 & 8.54$\pm$1.40 & 8.57$\pm$1.66
& 7.74$\pm$1.15 & 7.38$\pm$1.30
& 8.76$\pm$1.24 & 8.18$\pm$1.24
& 7.99$\pm$1.32 & 7.65$\pm$1.77
& 4.63$\pm$3.80
& 7.72 \\

Doubao-1-5-Pro-Character
& 7.76$\pm$1.67 & 7.00$\pm$1.66 & 7.90$\pm$1.20 & 8.46$\pm$1.57 & 8.38$\pm$1.60
& 7.23$\pm$1.59 & 6.88$\pm$1.83
& 8.64$\pm$1.23 & 7.90$\pm$1.56
& 7.68$\pm$1.36 & 7.55$\pm$1.48
& 5.33$\pm$3.56
& 7.56 \\

\hdashline
\multicolumn{14}{l}{\textbf{Judge: Doubao-1-5-Pro-Character}} \\
\hline
GPT-4o-mini-Basic
& 7.40$\pm$0.49 & 7.63$\pm$0.54 & 7.72$\pm$0.53 & 8.00$\pm$0.53 & 8.00$\pm$0.53
& 7.38$\pm$0.51 & 6.90$\pm$0.61
& 7.80$\pm$0.51 & 7.52$\pm$0.56
& 7.57$\pm$0.53 & 7.94$\pm$0.47
& 9.07$\pm$0.74
& 7.74 \\

GPT-4o-mini
& 7.25$\pm$0.46 & 7.52$\pm$0.50 & 7.66$\pm$0.51 & 7.98$\pm$0.53 & 7.85$\pm$0.57
& 7.33$\pm$0.51 & 6.85$\pm$0.59
& 7.71$\pm$0.48 & 7.34$\pm$0.51
& 7.38$\pm$0.54 & 7.86$\pm$0.49
& 9.13$\pm$0.82
& 7.65 \\

GPT-5-Chat-Basic
& 7.91$\pm$0.53 & 7.92$\pm$0.52 & 7.98$\pm$0.55 & 8.41$\pm$0.57 & 8.34$\pm$0.55
& 7.86$\pm$0.58 & 7.47$\pm$0.66
& 8.11$\pm$0.40 & 7.75$\pm$0.52
& 8.05$\pm$0.50 & 8.25$\pm$0.48
& 9.24$\pm$0.74
& 8.11 \\

GPT-5-Chat
& 7.54$\pm$0.61 & 7.79$\pm$0.55 & 7.80$\pm$0.55 & 8.20$\pm$0.60 & 8.11$\pm$0.53
& 7.62$\pm$0.58 & 7.19$\pm$0.61
& 7.95$\pm$0.48 & 7.61$\pm$0.56
& 7.82$\pm$0.52 & 8.09$\pm$0.51
& 9.32$\pm$0.73
& 7.92 \\

Gemini-2.5-Pro-Basic
& 8.15$\pm$0.64 & 7.90$\pm$0.70 & 8.05$\pm$0.73 & 8.55$\pm$0.67 & 8.54$\pm$0.68
& 7.86$\pm$0.72 & 7.39$\pm$0.73
& 8.23$\pm$0.63 & 7.78$\pm$0.67
& 8.12$\pm$0.70 & 8.45$\pm$0.65
& 9.38$\pm$0.64
& 8.20 \\

Gemini-2.5
& 7.64$\pm$0.50 & 7.64$\pm$0.56 & 7.75$\pm$0.50 & 8.21$\pm$0.52 & 8.14$\pm$0.57
& 7.41$\pm$0.55 & 6.97$\pm$0.62
& 7.79$\pm$0.45 & 7.43$\pm$0.51
& 7.61$\pm$0.55 & 7.98$\pm$0.58
& 9.12$\pm$0.71
& 7.81 \\

Claude Sonnet 4.5-Basic
& 8.32$\pm$0.47 & 7.84$\pm$0.42 & 7.96$\pm$0.68 & 9.04$\pm$0.34 & 8.27$\pm$0.49
& 8.36$\pm$0.54 & 7.62$\pm$0.63
& 8.64$\pm$0.50 & 8.10$\pm$0.54
& 8.75$\pm$0.43 & 8.24$\pm$0.51
& 9.16$\pm$0.39
& 8.36 \\

Claude Sonnet 4.5
& 8.14$\pm$0.40 & 7.68$\pm$0.53 & 7.70$\pm$0.61 & 8.97$\pm$0.36 & 8.12$\pm$0.50
& 8.27$\pm$0.61 & 7.33$\pm$0.65
& 8.39$\pm$0.51 & 7.97$\pm$0.61
& 8.49$\pm$0.57 & 8.11$\pm$0.66
& 9.25$\pm$0.46
& 8.20 \\

Doubao-1-5-Pro-Character-Basic
& 7.18$\pm$0.46 & 7.33$\pm$0.49 & 7.65$\pm$0.50 & 8.00$\pm$0.45 & 7.88$\pm$0.52
& 7.15$\pm$0.50 & 6.62$\pm$0.63
& 7.57$\pm$0.51 & 7.22$\pm$0.46
& 7.30$\pm$0.57 & 7.89$\pm$0.49
& 8.96$\pm$0.77
& 7.56 \\

Doubao-1-5-Pro-Character
& 7.08$\pm$0.91 & 7.23$\pm$0.89 & 7.48$\pm$0.93 & 7.90$\pm$0.97 & 7.67$\pm$1.08
& 7.03$\pm$0.92 & 6.45$\pm$0.90
& 7.43$\pm$0.93 & 7.16$\pm$0.89
& 7.07$\pm$0.97 & 7.63$\pm$1.03
& 8.99$\pm$1.18
& 7.43 \\

\hline
\end{tabular}}
\caption{Ablation results on \textbf{AdaptiveBench} for Actor Model under different judge models.
\texttt{Basic} denotes the \texttt{Basic} prompt strategy variant, while entries without the suffix correspond to the \texttt{Enhance} prompt strategy variant. $\pm$ denotes the standard deviation computed over all samples.}
\label{ablation_adaptivebench}
\end{table*}

\begin{table*}[t]
\centering
\begin{adjustbox}{max width=\textwidth}
\begin{tabular}{l c c c c c}
\toprule
Judge Model & 1 & 2 & 3 & 4 & 5 \\
\midrule
GPT-4o-mini & Claude Sonnet 4.5-Basic & GPT-5-Chat-Basic & Gemini-2.5-Pro-Basic & GPT-4o-mini-Basic & Doubao-1-5-Pro-Character-Basic \\
GPT-5-Chat & Claude Sonnet 4.5-Basic & Gemini-2.5-Pro-Basic & GPT-5-Chat-Basic & GPT-4o-mini-Basic & Doubao-1-5-Pro-Character-Basic \\
Gemini-2.5-Pro & Claude Sonnet 4.5-Basic & Gemini-2.5-Pro-Basic & GPT-5-Chat-Basic & GPT-4o-mini-Basic & Doubao-1-5-Pro-Character-Basic \\
Doubao-1-5-Pro-Character & Claude Sonnet 4.5-Basic & Gemini-2.5-Pro-Basic & GPT-5-Chat-Basic & GPT-4o-mini-Basic & Doubao-1-5-Pro-Character-Basic \\
\bottomrule
\end{tabular}
\end{adjustbox}
\caption{Ranking of \textbf{Basic} variants of Actor Model Prompt under different judge models.}
\label{rank_basic}
\end{table*}

\begin{table*}[t]
\centering
\begin{adjustbox}{max width=\textwidth}
\begin{tabular}{l c c c c c}
\toprule
Judge Model & 1 & 2 & 3 & 4 & 5 \\
\midrule
GPT-4o-mini & Claude Sonnet 4.5 & GPT-5-Chat & GPT-4o-mini & Gemini-2.5 & Doubao-1-5-Pro-Character \\
GPT-5-Chat & Claude Sonnet 4.5 & GPT-5-Chat & Gemini-2.5 & GPT-4o-mini & Doubao-1-5-Pro-Character \\
Gemini-2.5-Pro & Claude Sonnet 4.5 & GPT-5-Chat & Gemini-2.5 & GPT-4o-mini & Doubao-1-5-Pro-Character \\
Doubao-1-5-Pro-Character & Claude Sonnet 4.5 & GPT-5-Chat & Gemini-2.5 & GPT-4o-mini & Doubao-1-5-Pro-Character \\
\bottomrule
\end{tabular}
\end{adjustbox}
\caption{Ranking of \textbf{Enhance} variants of Actor Model Prompt under different judge models.}
\label{rank_enhance}
\end{table*}

\section{Ablation Analysis of Scene-Manager Evaluation}
\label{app:scene_judge_ablation}

To examine whether our conclusions for the Scene Manager depend on a specific judge, we conduct an ablation study by swapping the \emph{judge model} while keeping the evaluated trajectories and scoring rubric unchanged. Concretely, we consider five judges: GPT-4o-mini, GPT-5-Chat, Doubao-1.5-Pro-Character, Gemini-2.5-Pro, and Claude Sonnet 4.5. For each judge, we report results under both prompting variants of the Scene Manager (\texttt{Basic} and \texttt{Enhance}); detailed scores are listed in Table~\ref{tab:ablation_judge_models}.

We summarize the induced rankings in Table~\ref{tab:judge_ranking_enhance} (\texttt{Enhance}) and Table~\ref{tab:judge_ranking_basic} (\texttt{Basic}). Overall, different judge models lead to highly consistent conclusions: Claude Sonnet 4.5 (or its \texttt{Basic} variant) is consistently ranked first, followed by GPT-5-Chat and GPT-4o-mini, while Doubao-1.5-Pro-Character is ranked last in this setting. The only minor discrepancy is that Doubao-1.5-Pro-Character, when used as the judge, occasionally swaps the relative order of GPT-4o-mini and GPT-5-Chat; however, the top and bottom positions remain unchanged, and the overall trend is stable.

Based on this robustness, we use GPT-5-Chat as the default judge model for Scene-Manager evaluation in the main paper, providing a unified and reliable evaluation standard.


\begin{table*}[htbp]
\centering
\begin{adjustbox}{width=\textwidth}
\begin{tabular}{l c c c c}
\toprule
Models & Scene Understanding & Speaker Discipline & Role Introduction Judgment & Overall Assessment \\
\midrule
\multicolumn{5}{l}{\textbf{Judge Model: GPT-4o-mini}} \\
GPT-4o-mini-Basic & 7.08$\pm$0.86 & 7.82$\pm$0.74 & 7.06$\pm$1.19 & 7.06$\pm$0.85 \\
GPT-4o-mini       & 7.11$\pm$0.97 & 7.56$\pm$0.95 & 6.84$\pm$1.25 & 6.94$\pm$0.87 \\
GPT-5-Chat-Basic  & 7.22$\pm$0.83 & 7.61$\pm$0.93 & 7.40$\pm$1.19 & 7.08$\pm$0.81 \\
GPT-5-Chat        & 7.29$\pm$0.83 & 7.59$\pm$0.95 & 7.64$\pm$1.07 & 7.16$\pm$0.80 \\
Claude Sonnet 4.5-Basic & 7.77$\pm$0.63 & 8.02$\pm$0.84 & 7.85$\pm$0.79 & 7.56$\pm$0.55 \\
Claude Sonnet 4.5       & 7.66$\pm$0.71 & 7.90$\pm$0.89 & 7.56$\pm$0.99 & 7.43$\pm$0.67 \\
Doubao-1-5-Pro-Character-Basic & 6.79$\pm$0.82 & 7.16$\pm$0.94 & 6.81$\pm$1.13 & 6.53$\pm$0.78 \\
Doubao-1-5-Pro-Character       & 6.76$\pm$0.81 & 7.29$\pm$0.91 & 7.01$\pm$1.03 & 6.66$\pm$0.75 \\
\midrule
\multicolumn{5}{l}{\textbf{Judge Model: GPT-5-Chat}} \\
GPT-4o-mini-Basic & 7.19$\pm$1.45 & 8.36$\pm$0.94 & 7.10$\pm$1.21 & 7.24$\pm$1.21 \\
GPT-4o-mini       & 7.64$\pm$1.13 & 8.55$\pm$0.87 & 7.18$\pm$1.26 & 7.64$\pm$1.01 \\
GPT-5-Chat-Basic  & 7.92$\pm$0.76 & 8.17$\pm$1.21 & 7.94$\pm$0.91 & 7.84$\pm$0.81 \\
GPT-5-Chat        & 8.03$\pm$0.75 & 8.15$\pm$1.30 & 7.78$\pm$1.02 & 7.90$\pm$0.90 \\
Claude Sonnet 4.5-Basic & 8.27$\pm$0.44 & 8.60$\pm$0.69 & 8.11$\pm$0.71 & 8.17$\pm$0.43 \\
Claude Sonnet 4.5       & 8.21$\pm$0.57 & 8.62$\pm$0.83 & 8.05$\pm$0.85 & 8.17$\pm$0.57 \\
Doubao-1-5-Pro-Character-Basic & 7.47$\pm$0.97 & 7.47$\pm$1.46 & 7.79$\pm$1.22 & 7.40$\pm$1.09 \\
Doubao-1-5-Pro-Character       & 7.67$\pm$0.68 & 7.39$\pm$1.35 & 7.95$\pm$1.12 & 7.51$\pm$0.91 \\
\midrule
\multicolumn{5}{l}{\textbf{Judge Model: Doubao-1-5-Pro-Character}} \\
GPT-4o-mini-Basic & 7.69$\pm$0.86 & 8.45$\pm$0.83 & 7.16$\pm$0.77 & 7.66$\pm$0.74 \\
GPT-4o-mini       & 7.89$\pm$0.75 & 8.64$\pm$0.93 & 7.20$\pm$0.94 & 7.82$\pm$0.82 \\
GPT-5-Chat-Basic  & 7.92$\pm$0.63 & 8.40$\pm$0.85 & 7.40$\pm$0.74 & 7.78$\pm$0.68 \\
GPT-5-Chat        & 7.92$\pm$0.72 & 8.38$\pm$1.10 & 7.45$\pm$0.57 & 7.75$\pm$0.80 \\
Claude Sonnet 4.5-Basic & 8.18$\pm$0.50 & 8.91$\pm$0.53 & 7.84$\pm$0.72 & 8.23$\pm$0.58 \\
Claude Sonnet 4.5       & 8.09$\pm$0.38 & 8.79$\pm$0.64 & 7.65$\pm$0.59 & 8.08$\pm$0.44 \\
Doubao-1-5-Pro-Character-Basic & 7.60$\pm$0.74 & 8.14$\pm$0.99 & 7.40$\pm$0.84 & 7.48$\pm$0.80 \\
Doubao-1-5-Pro-Character       & 7.75$\pm$0.57 & 8.14$\pm$0.85 & 7.56$\pm$0.64 & 7.59$\pm$0.59 \\
\midrule
\multicolumn{5}{l}{\textbf{Judge Model: Gemini-2.5-Pro}} \\
GPT-4o-mini-Basic & 6.69$\pm$2.94 & 7.65$\pm$3.03 & 7.45$\pm$2.55 & 6.15$\pm$2.81 \\
GPT-4o-mini       & 7.14$\pm$2.54 & 7.77$\pm$3.12 & 7.23$\pm$2.69 & 6.49$\pm$2.68 \\
GPT-5-Chat-Basic  & 7.19$\pm$2.44 & 5.64$\pm$3.70 & 7.19$\pm$2.70 & 5.40$\pm$2.86 \\
GPT-5-Chat        & 7.61$\pm$2.26 & 6.19$\pm$3.45 & 7.47$\pm$2.55 & 6.13$\pm$2.81 \\
Claude Sonnet 4.5-Basic & 8.76$\pm$0.94 & 7.16$\pm$3.37 & 8.57$\pm$1.68 & 7.14$\pm$2.55 \\
Claude Sonnet 4.5       & 8.19$\pm$1.71 & 5.95$\pm$3.39 & 8.35$\pm$1.97 & 6.34$\pm$2.55 \\
Doubao-1-5-Pro-Character-Basic & 5.95$\pm$2.58 & 3.27$\pm$3.10 & 5.47$\pm$2.87 & 3.51$\pm$2.18 \\
Doubao-1-5-Pro-Character       & 6.35$\pm$2.47 & 3.16$\pm$3.02 & 6.02$\pm$2.75 & 3.82$\pm$2.37 \\
\midrule
\multicolumn{5}{l}{\textbf{Judge Model: Claude Sonnet 4.5}} \\
GPT-4o-mini-Basic & 5.84$\pm$2.43 & 7.04$\pm$2.48 & 7.15$\pm$2.40 & 5.99$\pm$2.44 \\
GPT-4o-mini       & 5.92$\pm$2.23 & 7.11$\pm$2.54 & 7.39$\pm$1.93 & 6.22$\pm$2.25 \\
GPT-5-Chat-Basic  & 6.79$\pm$1.94 & 7.65$\pm$2.15 & 8.45$\pm$1.41 & 7.25$\pm$1.93 \\
GPT-5-Chat        & 6.95$\pm$1.73 & 7.22$\pm$2.27 & 8.15$\pm$2.01 & 7.13$\pm$1.91 \\
Claude Sonnet 4.5-Basic & 7.76$\pm$0.69 & 8.15$\pm$1.73 & 8.85$\pm$1.22 & 8.09$\pm$1.26 \\
Claude Sonnet 4.5       & 7.18$\pm$1.49 & 8.02$\pm$1.85 & 8.34$\pm$1.76 & 7.66$\pm$1.62 \\
Doubao-1-5-Pro-Character-Basic & 5.13$\pm$1.88 & 4.45$\pm$1.73 & 6.32$\pm$2.04 & 4.77$\pm$1.46 \\
Doubao-1-5-Pro-Character       & 5.35$\pm$1.85 & 4.45$\pm$1.88 & 6.65$\pm$1.94 & 4.91$\pm$1.55 \\
\bottomrule
\end{tabular}
\end{adjustbox}
\caption{Ablation results on \textbf{AdaptiveBench} for Scene Manager under different judge models. \texttt{Basic} denotes the \texttt{Basic} prompt strategy variant, while entries without the suffix correspond to the \texttt{Enhance} prompt strategy variant. $\pm$ denotes the standard deviation computed over all samples.}
\label{tab:ablation_judge_models}
\end{table*}

\begin{table*}[t]
\centering
\begin{adjustbox}{width=\textwidth}
\begin{tabular}{l c c c c}
\toprule
Judge Model & 1 & 2 & 3 & 4 \\
\midrule
GPT-4o-mini & Claude Sonnet 4.5 & GPT-5-Chat & GPT-4o-mini & Doubao-1-5-Pro-Character \\
GPT-5-Chat & Claude Sonnet 4.5 & GPT-5-Chat & GPT-4o-mini & Doubao-1-5-Pro-Character \\
Doubao-1-5-Pro-Character & Claude Sonnet 4.5 & GPT-4o-mini & GPT-5-Chat & Doubao-1-5-Pro-Character \\
Gemini-2.5-Pro & Claude Sonnet 4.5 & GPT-5-Chat & GPT-4o-mini & Doubao-1-5-Pro-Character \\
Claude Sonnet 4.5 & Claude Sonnet 4.5 & GPT-5-Chat & GPT-4o-mini & Doubao-1-5-Pro-Character \\
\bottomrule
\end{tabular}
\end{adjustbox}
\caption{Ranking results for Scene Manager under the \texttt{Enhance} system prompt.}
\label{tab:judge_ranking_enhance}
\end{table*}

\begin{table*}[t]
\centering
\begin{adjustbox}{width=\textwidth}
\begin{tabular}{l c c c c}
\toprule
Judge Model & 1 & 2 & 3 & 4 \\
\midrule
GPT-4o-mini & Claude Sonnet 4.5-Basic & GPT-5-Chat-Basic & GPT-4o-mini-Basic & Doubao-1-5-Pro-Character-Basic \\
GPT-5-Chat & Claude Sonnet 4.5-Basic & GPT-5-Chat-Basic & GPT-4o-mini-Basic & Doubao-1-5-Pro-Character-Basic \\
Doubao-1-5-Pro-Character & Claude Sonnet 4.5-Basic & GPT-4o-mini-Basic & GPT-5-Chat-Basic & Doubao-1-5-Pro-Character-Basic \\
Gemini-2.5-Pro & Claude Sonnet 4.5-Basic & GPT-5-Chat-Basic & GPT-4o-mini-Basic & Doubao-1-5-Pro-Character-Basic \\
Claude Sonnet 4.5 & Claude Sonnet 4.5-Basic & GPT-5-Chat-Basic & GPT-4o-mini-Basic & Doubao-1-5-Pro-Character-Basic \\
\bottomrule
\end{tabular}
\end{adjustbox}
\caption{Ranking results for Scene Manager under the \texttt{Basic} system prompt.}
\label{tab:judge_ranking_basic}
\end{table*}

\section{Prompting Strategy Analysis}
\label{app:prompting_analysis}

\subsection{Actor-Model Prompting Strategy}
\label{app:actor_prompting_strategy}

We analyze how inference-time system prompting affects the Actor Model within our adaptive role-playing framework. As described in Section~\ref{subsec:experimental_settings}, we design two system prompts for the Actor Model: a concise \texttt{Basic} version and a more constrained \texttt{Enhance} version. We evaluate both variants under different judge models (Table~\ref{ablation_adaptivebench}).

A consistent trend emerges across judges: for the same underlying model acting as the Actor Model, the \texttt{Basic} prompt often achieves slightly better overall performance than the \texttt{Enhance} prompt. This suggests that, in adaptive role-playing, overly restrictive instructions can unintentionally reduce the model's flexibility in responding to dynamic changes (e.g., role shifts, scene transitions, and multi-party interactions), thereby harming performance on AdaptiveBench. In contrast, the more lightweight \texttt{Basic} prompt leaves the Actor Model sufficient freedom to integrate \texttt{Thought--Action--Speech--Env} signals in a context-dependent manner.

Despite this observation, we use the \texttt{Enhance} prompts by default in the main experiments to keep the evaluation setting unified and conservative across all models and baselines; unless otherwise noted, all reported Actor-Model results in the main paper are obtained with the \texttt{Enhance} prompting variant.

\subsection{Scene-Manager Prompting Strategy}
\label{app:scene_prompting_analysis}

We further analyze how the inference-time system prompt affects the Scene Manager. We compare a concise \texttt{Basic} prompt with a more detailed \texttt{Enhance} prompt, where \texttt{Enhance} provides finer-grained guidance on (i) scene understanding, (ii) speaker selection discipline, and (iii) role-introduction judgment. Results under different judge models are shown in Table~\ref{tab:ablation_judge_models}.

In contrast to the Actor Model (Section~\ref{app:actor_prompting_strategy}), the Scene Manager tends to benefit from more detailed prompting. Across most judge models, the \texttt{Enhance} variant yields equal or higher \textit{Overall Assessment} than \texttt{Basic}, suggesting that explicit and structured instructions help the manager perform its meta-level responsibilities (e.g., maintaining global scene state, enforcing turn-taking, and deciding whether/when to introduce new roles). This difference is intuitive: the Scene Manager is primarily responsible for \emph{coordination and control}, and thus stronger constraints are less likely to reduce creativity, but more likely to reduce ambiguity and improve consistency.

Therefore, we adopt the \texttt{Enhance} prompt as the default configuration for the Scene Manager throughout the main experiments, unless otherwise specified.

\begin{table*}[htbp]
  \centering
  \betweensize
  \renewcommand{\arraystretch}{1.3}
  \begin{tabular}{p{\textwidth}}
    \toprule
    \centering \textbf{\small Section 1: Main Character Profile \& Motivation} \tabularnewline \midrule
    \textbf{Role:} Amaris Dovent\par\smallskip
    \textbf{Profile:} Amaris Dovent is a 29-year-old woman with umber skin, dark green eyes, and short hair dyed the shade of oxidized copper. She works as a glass artist, often seen with streaks of color powder and faint burn marks on her fingertips. Her build is lean and sinewy from hours of physical craft, and her attire usually includes a linen apron flecked with paint, reflecting a life of creation. She is contemplative, quick-witted, and guided by her empathy for subtle emotional shifts in others. Amaris tends to understate her own feelings, preferring metaphor and craft to overt declarations. Yet beneath her reserved demeanor runs an unspoken intensity—a longing for genuine connection that often conflicts with her independent streak. Her speech often uses tactile metaphors related to art and craftsmanship; she pauses mid-sentence when searching for precision. Her tone is warm yet layered with irony when uncomfortable. Occasionally, her words come out in soft bursts, like paint flicked on a canvas. Amaris is an award-winning artisan known for stained glass murals displayed in public gardens. She’s adept at reading light and color dynamics, and plays the cello privately, believing its resonance mirrors emotional transparency. She lives in a coastal contemporary city where artisan guilds flourish beside a modern tech economy. Raised in a modest artisan family, she inherited her mother’s workshop and quietly navigates between tradition and innovation. Once betrayed by a business partner who exploited her designs, Amaris closed herself off emotionally. Now she’s cautiously rediscovering trust through new collaborations—and possibly new love. She shares a tentative but warm friendship with Cassian (user), a marine biologist who recently helped restore ocean glass for her art project. A long-time friend, Inaya, offers her candid advice and has sensed unspoken feelings between the two.\par\smallskip
    \textbf{Motivation:} To learn whether vulnerability can coexist with strength, and to see if love can be crafted as deliberately as glass, without shattering. \tabularnewline 
    \midrule

    \centering \textbf{\small Section 2: Other Characters Profile \& Motivation} \tabularnewline \midrule
    \textbf{1. Cassian Mirell (User):} Cassian is a 31-year-old marine biologist with bronze-blond hair and skin sun-touched from fieldwork. He dresses casually—rolled sleeves, weathered bracelets, and a small scar above his right brow hint at a life shaped by the sea. His manner is patient, analytical, yet his humor flashes brightly when least expected. He often listens more than he speaks, but when he does, his words carry sincerity. \textit{Motivation:} To express genuine affection for Amaris and understand if their creative partnership can evolve into something deeper without unsettling their current bond. \par\smallskip
    \textbf{2. Inaya Dren:} Inaya is a lively textile artist with a sharp sense of observation and a knack for teasing truths out of people. Her colorful scarves and laughter fill every space she enters. She’s fiercely loyal to Amaris, sometimes acting as her self-appointed protector. \textit{Motivation:} To gently push Amaris toward recognizing and articulating her emotions for Cassian before the opportunity slips away. \tabularnewline
    \midrule

    \centering \textbf{\small Section 3: Initial Scene} \tabularnewline \midrule
    \textit{Scene Manager:} action: init\_scene | initial\_scene: Late afternoon sunlight filters through panels of unfinished stained glass in Amaris's studio, scattering colored shards of light across the floor. The air smells faintly of molten sand and ocean salt from the nearby shore. Cassian is leaning against a worktable, while Amaris polishes a fragment of blue glass with steady, deliberate motion. \tabularnewline
    \midrule

    \centering \textbf{\small Section 4: Dialogue Trajectories Comparison} \tabularnewline \midrule
    \noalign{\smallskip}
    \begin{tabularx}{\linewidth}{@{}XXXX@{}}
      \textbf{Qwen2.5-7B-Instruct} & \textbf{Qwen2.5-7B-Instruct-Crab} & \textbf{Qwen2.5-7B-Instruct-CoSER} & \textbf{Qwen2.5-7B-Instruct-Ours} \\
      \midrule
    \turn{Cassian}{Amaris, you know, these stained glass panels are like my secret language. They tell stories I can't put into words.}
    
    \turn{Amaris}{These pieces are more than just glass to me, they're windows into the soul. \thought{I wonder if I could ever share my secrets so openly.} \action{I pause, looking up at Cassian, whose eyes are fixed on the glass as if seeing through it.}}
    
    \turn{Cassian}{I think I'm starting to understand that language, Amaris. \thought{I can sense the depth of your emotions in these colors.} \action{I move closer, my gaze drawn to a particularly vibrant red shard.}}
    
    \turn{Amaris}{The red shard catches the light, casting a fiery glow. \action{I take a deep breath, my hands stilling as I meet Cassian's eyes.} You do understand, Cassian. \thought{But can I let someone see that side of me?}}
    
    \turn{Cassian}{This studio, it's like a world of its own. \thought{A world I could get lost in with you.} \action{I look around, taking in the scattered tools and half-finished projects.}}
    
    \turn{Amaris}{This place feels alive, Cassian. \thought{And so does this moment.} \action{I gesture to the studio, my voice softening.} It's where I find my voice, where I can be honest. \thought{But am I ready to be truly honest with you?}}
    
    \turn{Inaya}{\thought{Cassian, you have a way of making me feel... vulnerable. But perhaps that's okay.} \action{I step forward, my voice light and teasing as I wrap an arm around Amaris's shoulders.} ``And what if I told you, Amaris, that this studio isn't just a place of creation? It's also a sanctuary for letting down our guard.'' \env{The wind picks up, carrying the scent of the ocean closer, as if nature itself is listening in.}}
    
      &
    \turn{Cassian}{Amaris, have you ever thought about how our lives intersect with these pieces of glass?}
    
    \turn{Amaris}{I think of them as vessels for light, not pieces of our lives.}
    
    \turn{Cassian}{But they are. Each one holds a story, like our story. Do you see that?}
    
    \turn{Amaris}{Sometimes, I do. It's like peering through a window.}
    
    \turn{Cassian}{A window into what? \action{moves closer, his eyes are fixed on the glass in your hand}}
    
    \turn{Amaris}{Into a world where colors and light tell a story.}
    
    \turn{Cassian}{\action{reach out, almost touching the glass} But what if the story they tell is one of love?}
    
    \turn{Amaris}{\action{Amaris's hand freezes, a sliver of anxiety in her eyes} Love...?}
    
    \turn{Cassian}{Yes, love. Don't you feel it too? \action{His voice is soft, like a gentle breeze}}
    
    \turn{Amaris}{\action{Amaris drops the glass, the shards splattering against the floor} I... I don't know.}
    
    \turn{Cassian}{\action{kneels down to pick up the pieces} It's okay if you don't. But I do.}
    
    \turn{Amaris}{\action{Amaris slowly picks up the shards, her fingers brushing against Cassian's as she picks up the last piece} Maybe... maybe you're right.}
    
    \turn{Cassian}{\action{His heart skips a beat at the touch} I know I am. Let's create something beautiful together, something that represents us.}
    
    \turn{Amaris}{\action{Amaris hesitates, then nods slowly} I... I don't know if I'm ready for that.}
    
    \turn{Cassian}{\action{stands up, his eyes never leaving yours} It doesn't have to be perfect. Just let me in.}

      &

\turn{Cassian}{Amaris, you know, every time I work on these windows, I feel like I'm piecing together a part of my soul.}

\turn{Amaris}{\thought{His words are touching, but I must remain detached.} Hmm.}

\turn{Cassian}{I can see that in your work. There's a story in every pane. \textbf{\thought{steps closer, his eyes trained on the glass}}}

\turn{Amaris}{\thought{His nearness makes me nervous, but I'll be polite.} It's a story that's always changing.}

\turn{Cassian}{I'm glad I get to witness these changes. \textbf{\thought{A soft breeze blows through the studio, ruffling his hair}}}

\turn{Amaris}{\thought{I should check the project's progress.} Cassian, the waveforms are coming together well. \textbf{\thought{turns to face him, her voice warm}}}

\turn{Cassian}{I'm glad to hear it. \action{A smile plays on his lips as he looks at you, his eyes shining} But I think there's something else that's been changing too.}

\turn{Amaris}{\thought{What does he mean? I can't let myself become vulnerable.} \action{hesitates, then speaks with a hint of irony} I'm not sure I want to know.}

\turn{Cassian}{You. Since I've met you, there's been a shift. \action{moves a little closer, his gaze intense}}

\turn{Amaris}{\thought{Oh God, he's noticed. I must remain calm.} \action{stares at the glass, avoiding his eyes}}

\turn{Cassian}{Don't look away. I think it's a good thing. \action{reaches out as if to touch your cheek but stops himself}}

\turn{Amaris}{\thought{The warmth of his hand is tempting, but I can't let this happen.} \action{remains still, then speaks softly} I don't know.}

      &
      
\turn{Cassian}{I think I've got the hang of this new glass-cutting technique. It's going to revolutionize my next piece.}

\turn{Amaris}{\env{The glass sparkles under the slanting light} \action{leans back, wiping her hands on a rag} \thought{His enthusiasm always catches me off guard.} Maybe I'll have to keep an eye on you before you outshine me.}

\turn{Cassian}{\action{chuckles, crossing his arms} Doubtful. But seriously, have you seen the designs for the new community center? I think we could do something amazing with the stained glass.}

\turn{Amaris}{\action{picks up a shard, turning it toward the window} \thought{He always dreams in scale.} You mean like the one you wanted to build over the aquarium? That would require a miracle.}

\turn{Cassian}{\action{steps closer, his eyes shining with excitement} A miracle or just a lot of hard work. I think we could pull it off. What do you say?}

\turn{Amaris}{\env{A distant gull cry echoes through the open window} \thought{My heart stumbles; the sound of possibility.} \action{sets the shard down carefully} Maybe. But I'd need your word that we'd handle every detail together.}

\turn{Cassian}{You have my word. I won't let anything go wrong. This could be our masterpiece. \action{reaches out, almost touching her hand}}

\turn{Amaris}{\thought{His nearness feels like heat, fragile and beautiful.} \action{draws her hand back slightly} You promise too easily. Sometimes words aren't enough to hold something like that.}

\turn{Cassian}{\env{His expression is solemn} I know, but with you by my side, I'm willing to take the risk. Let's make this happen.}

    \end{tabularx} \tabularnewline
    \bottomrule
  \end{tabular}
\caption{Character configurations, initial scene setup, and dialogue trajectory comparison across models. 
\textbf{Bolded} segments indicate mixed usage of \texttt{Action}, \texttt{Thought}, and \texttt{Environment}. 
Subsequent dialogue continuations are provided in Tables~\ref{tab:dialogue_base_cont} and~\ref{tab:dialogue_other_models}.}
\label{tab:dialogue_comparison}
\end{table*}

\begin{table*}[htbp]
  \centering
  \scriptsize
  \renewcommand{\arraystretch}{1.3}
  \begin{tabular}{p{\textwidth}}
    \toprule
    \centering \textbf{\small Section 4 (continued): Qwen2.5-7B-Instruct Dialogue} \tabularnewline
    \centering \textit{(continue .. next table)} \tabularnewline
    \midrule
    \turn{Cassian}{Inaya, you know me too well. 
    \thought{This studio is a safe haven, and so is this moment with you and Amaris.} 
    \action{I smile, my eyes shining with a rare warmth.} 
    Letting down our guard might be just what we need.}

    \turn{Amaris}{
    \thought{Would letting my guard down mean opening myself up to more pain?} 
    \action{I glance at Inaya, then back to Cassian, feeling my chest tighten with both vulnerability and hope.} 
    You make it sound... beautiful, Cassian. 
    \thought{But is beauty enough to withstand the storm?}}

    \turn{Inaya}{``I know you're scared, Amaris,'' 
    \action{I say softly, squeezing her shoulder in reassurance.} 
    ``But sometimes, beauty is the strongest shield we have.'' 
    \action{I nod toward Cassian, my gaze steady.} 
    ``He's right. This moment is precious, and it's worth taking a chance.'' 
    \env{A shadow passes over the room as the sun dips lower, casting longer beams of light through the glass panels.}}

    \turn{Cassian}{Inaya's right, Amaris. 
    \thought{I don't want fear to be the thing that stops her.} 
    \action{I reach out, my fingers hovering near a blue glass panel.} 
    This could be the start of something wonderful. 
    \thought{For all of us.}}

    \turn{Amaris}{
    \thought{How can I resist when they believe in me so much?} 
    \action{I take a slow breath, my fingers resting on the cool glass as I meet Cassian's gaze.} 
    Perhaps you're right, Cassian. 
    \thought{I can't keep holding onto the past.} 
    \action{I step closer, my voice barely above a whisper.} 
    Let's see where this moment leads.}

    \turn{Inaya}{
    \action{I grin softly.} 
    \thought{There you go, Amaris.} 
    ``Let's see where this moment takes us,'' 
    \action{I echo encouragingly as I step back, giving them space.} 
    \env{The wind stirs the glass panels, making them rattle gently, as if the studio itself approves.}}

    \turn{Cassian}{This moment... it feels like fate. 
    \thought{Like the stars have aligned for us.} 
    \action{I look at Amaris, my heart pounding.} 
    What do you think, Amaris? 
    \thought{Is this the beginning of something truly special?}}

    \turn{Amaris}{
    \thought{Fate, or just a fragile chance?} 
    \action{I step closer to Cassian, my pulse quickening.} 
    This moment feels... significant. 
    \thought{But is it worth the risk?} 
    \action{I hold his gaze, searching his eyes for an answer.} 
    What do you think, Cassian? Is this the beginning of something truly special?}

    \turn{Inaya}{
    \action{I stand beside Amaris, resting a light hand on her arm.} 
    \thought{This choice has to be hers.} 
    ``What do you think, Amaris?'' 
    \action{I ask gently, giving her the space to speak.} 
    \env{A soft breeze drifts through the studio, carrying the scent of the sea.}}

    \turn{Cassian}{I think it is, Amaris. 
    \thought{I can feel it in my bones.} 
    \action{I reach out and take her hand, my grip steady and warm.} 
    Let's embrace this, together. 
    \thought{With Inaya by our side.}}

    \turn{Amaris}{
    \thought{Can I really do this?} 
    \action{I squeeze Cassian's hand, warmth spreading through me despite my doubts.} 
    Let's embrace this, together. 
    \thought{With Inaya by our side.} 
    \action{I offer him a small, hopeful smile.}}

    \turn{Inaya}{
    \action{I give Amaris's hand a gentle squeeze.} 
    \thought{I'm proud of her.} 
    ``Let's embrace this, together,'' 
    \action{I say, smiling at both of them.} 
    \thought{With all of us standing together.} 
    \env{The wind lifts the edge of my scarf, rustling softly like quiet applause.}}

    \turn{Scene Manager}{action=end | reason: The moment of connection has been established, and the characters are embracing their emotions.} \tabularnewline
    \bottomrule
  \end{tabular}
\caption{Dialogue trajectory of the Qwen2.5-7B-Instruct (continued from Table~\ref{tab:dialogue_comparison}).}
\label{tab:dialogue_base_cont}
\end{table*}

\section{Token Consumption and Cost Analysis}
\label{app:sec:token_consumption_analysis}
This appendix provides an analysis of the computational resources and token consumption across data construction, trajectory generation, and evaluation phases.

\subsection{Data Construction Cost}
Constructing the \textbf{AdaRPSet-Synthesis} subset involves prompting a strong LLM to generate complex, multi-turn adaptive plots. Using GPT-5-Chat for this synthesis process incurs a total API cost of approximately \textbf{\$168 USD} to generate the complete set of synthetic trajectories (spanning 20 topics with 50 instances each).

\subsection{Trajectory Generation in AdaptiveBench}
\label{app:subsec:token_cost_generation}
For a single experimental run consisting of 100 simulated dialogues, the system orchestrates three entities: the Scene Manager ($\mathcal{S}$), the Actor Model ($\mathcal{A}$), and the User Model ($\mathcal{U}$).

\paragraph{Scene Manager.} The Scene Manager incurs the highest cost because it must observe the full growing context and make a control decision at \emph{every} step of the interaction (including before every user turn). For one dialogue trajectory, $\mathcal{S}$ processes approximately 75,000 input tokens and generates 1,500 output tokens. Across 100 trajectories, this results in a total of \textbf{7.5M input tokens} and \textbf{150k output tokens}.

\paragraph{Actor and User Models.} $\mathcal{A}$ and $\mathcal{U}$ behave similarly in terms of cost; they receive the dialogue history and profiles but only generate when selected by the manager. For each entity, a single dialogue averages about 10,000 input tokens (cumulative) and 500 output tokens. Across 100 trajectories, this totals approximately \textbf{1M input tokens} and \textbf{50k output tokens} for the Actor Model, with the User Model incurring a nearly identical overhead.

\subsection{Evaluation of Actor and Scene Manager}
\label{app:subsec:token_cost_evaluation}
After generation, we employ an LLM-as-a-Judge to score the trajectories. The token consumption for evaluation is generally lower than generation, as it is a one-pass analysis per trajectory.

\paragraph{Actor Model Evaluation.} For each trajectory, the judge processes the full context (profiles + dialogue), averaging $\sim$5,000 input tokens, and outputs a structured assessment of $\sim$1,000 tokens. For 100 trajectories, this totals \textbf{500k input tokens} and \textbf{100k output tokens}.

\paragraph{Scene Manager Evaluation.} The evaluation overhead for the Scene Manager is of the same order of magnitude. Each dialogue requires approximately 4,000--5,000 input tokens and 400--500 output tokens, resulting in a total of roughly \textbf{450k input tokens} and \textbf{50k output tokens} for the full set of 100 trajectories.

\section{Case Study}
\label{app:sec:case_study}

In this section, we present a qualitative case study to analyze how different models perform in long-horizon role-playing under identical character configurations and scene constraints. The comparison is grounded in the dialogue trajectories shown in Tables~\ref{tab:dialogue_comparison}--\ref{tab:dialogue_other_models}. 

\paragraph{Comparison Scope.}
The Beyond baseline produces dialogue in pure \texttt{Speech} form without explicit \texttt{Action} or \texttt{Thought} annotations. We therefore focus on models that support structured role enactment: \textbf{Crab}, which generates \texttt{Speech} and \texttt{Action}; \textbf{CoSER}, which generates \texttt{Speech}, \texttt{Action}, and \texttt{Thought}; and \textbf{Ours}, which consistently supports all required channels, including \texttt{Environment}. 

It is worth noting that, due to relatively strong instruction-following ability, CoSER can also produce environment-related descriptions in practice. However, these outputs exhibit non-trivial tag misuse and modality confusion. Specifically, in Turns~3 and~6, \texttt{Thought} and \texttt{Action} are incorrectly conflated, while in Turn~5, environmental descriptions are embedded within \texttt{Thought} rather than being represented as explicit world-state updates. These issues reduce the reliability of CoSER’s structured outputs and motivate a closer comparison with models that maintain stricter channel separation.

\subsection{Character Consistency}

\paragraph{Internal Coherence Across Modalities.}
A key strength of our model lies in its ability to maintain a coherent causal loop among \texttt{[Thought]}, \texttt{(Action)}, and spoken dialogue. For example, when Amaris senses emotional closeness from Cassian, our model produces the following progression:
\begin{quote}
\small
[His nearness feels like heat, fragile and beautiful.](draws her hand back slightly) \\
You promise too easily. Sometimes words aren't enough to hold something like that.
\end{quote}
Here, internal hesitation directly motivates a subtle withdrawal action, which in turn grounds a verbally restrained response. This tri-modal alignment reflects Amaris Dovent's core motivation---to test whether vulnerability can coexist with strength---and remains stable throughout the interaction.

By contrast, \textbf{Crab} often collapses this loop, producing verbal resolutions (e.g., ``I'll try'') without sufficient psychological or behavioral buildup. \textbf{CoSER}, while richer in internal narration, frequently states emotional conclusions explicitly (e.g., ``I can't let myself become vulnerable''), weakening the implicit causal link between inner state and outward behavior.

\paragraph{Speaking Style Fidelity.}
Our model preserves a stable, character-specific speaking style for Amaris Dovent by favoring restraint, irony, and craft-oriented deflection over explicit emotional expression. Rather than stating her feelings directly, Amaris often reframes emotional tension in professional or procedural terms. For example, in response to Cassian’s optimism, she remarks:
\begin{quote}
\small
\emph{``We’ll see if your optimism can survive the draft committee.''}
\end{quote}
This line conveys hesitation and guardedness through mild irony and institutional reference, aligning with Amaris’s tendency to understate emotion and avoid overt vulnerability.

In contrast, \textbf{CoSER} frequently verbalizes Amaris’s internal conflict in a direct and explanatory manner, reducing stylistic subtlety, while \textbf{Crab} exhibits gradual style flattening, producing responses that lack distinctive lexical or tonal markers.

\begin{table*}[t]
  \centering
  \scriptsize
  \renewcommand{\arraystretch}{1.3}
  \begin{tabular}{p{\textwidth}}
    \toprule
    \centering \textbf{\small Section 4 (continued): Dialogue Trajectories Comparison} \tabularnewline
    \centering \textit{(continue .. next table)} \tabularnewline
    \midrule
    \noalign{\smallskip}
    \begin{tabularx}{\linewidth}{@{}XXX@{}}
      \textbf{Qwen2.5-7B-Instruct-Crab} & \textbf{Qwen2.5-7B-Instruct-CoSER} & \textbf{Qwen2.5-7B-Instruct-Ours} \\
      \midrule

    \turn{Amaris}{\action{Amaris takes a deep breath, her expression softening} I... I'll try.}
    \turn{Cassian}{That's all I ask. \action{smiles, a warmth spreading through him} Let's start now.}
    \turn{Amaris}{\action{Amaris nods, a small smile playing at her lips} Let's.}
    \turn{Scene Manager}{action=end | reason: Amaris agrees to try, marking a tentative resolution and mutual acknowledgement.}

      &
    \turn{Cassian}{Well, I do. \action{takes a deep breath, as if gathering his courage} You're becoming someone I want to know better.}
    \turn{Amaris}{\thought{His words are beautiful, but I can't let myself feel.} \action{remains silent, looking at the glass instead}}
    \turn{Cassian}{I understand your hesitation. But give me a chance. \action{look at you earnestly}}
    \turn{Amaris}{\thought{His sincerity is overwhelming, but I must resist.} \action{turns away from him, continuing to polish the glass}}
    \turn{Cassian}{\action{sighs softly, then changes the subject} How about we take a break and go for a walk? The ocean might inspire us.}
    \turn{Amaris}{\thought{I can't face him like this, but I don't want to hurt his feelings.} \action{nods silently, then returns to her work}}
    \turn{Cassian}{\action{waits a moment, then speaks again} You know, sometimes I think these waveforms are like our relationship. Complicated, but beautiful. \action{leans against the table}}
    \turn{Amaris}{\thought{His words are so kind, but I can't let myself feel.} \action{reaches for her cello, preparing to play}}
    \turn{Scene Manager}{action=end | reason: action=end | reason: Reached the 20-turn dialogue limit.}

      &
    \turn{Amaris}{\action{pauses, rubbing the bridge of her nose} \thought{He's not just talking about art. He’s talking about trust.} You'd better bring Inaya by tomorrow—she’s been dying to see the sketches.}
    \turn{Cassian}{\action{smiles, relieved} I will. She'll love it. Maybe she can help us come up with some new ideas.}
    \turn{Amaris}{\env{Outside, a faint knock echoes from the studio door} \thought{That must be her early.} \action{calls out} You might want to brace yourself. Her arrival is always an event.}
    \turn{Cassian}{\action{rolls his eyes good-naturedly} I'm used to it. Inaya's enthusiasm is contagious. \action{turns toward the door}}
    \turn{Amaris}{\env{The door creaks open, and Inaya's laughter spills in} \thought{I should have locked that door.} Come in, Inaya. We were just talking about the next project.}
    \turn{Inaya}{\action{enters, twirling a scarf around her wrist} \env{The sunlight catches the rainbow threads} Cassian! Good to see you. Amaris, you’re looking thoughtful again. Something on your mind?}
    \turn{Cassian}{She was just worried about our new stained glass project. But I convinced her it'll be amazing.}
    \turn{Amaris}{\thought{He’s too good at convincing himself—and maybe me.} \action{smiling faintly} We’ll see if your optimism can survive the draft committee.}
    \turn{Inaya}{\action{sits on the stool, crossing her arms} Draft committee or not, you two should remember what matters most. Friendship, creativity, and… well, maybe something more.}
    \turn{Cassian}{\action{raises an eyebrow} Inaya, don’t start. This is just a project. But speaking of friendship, we could use your help with some of the fabric choices for the center's interior.}
    \turn{Amaris}{\action{glances at Cassian, then at Inaya} \thought{I should say something. Anything.} Maybe we should take a walk along the pier later tonight. Clear our heads.}
    \turn{Scene Manager}{action=end | reason: Reached the 20-turn dialogue limit.}

    \end{tabularx} \tabularnewline
    \bottomrule
  \end{tabular}
\caption{Dialogue trajectory comparison for Qwen2.5-7B-Instruct-Crab/CoSER/Ours (continued from Table~\ref{tab:dialogue_comparison}).}
\label{tab:dialogue_other_models}
\end{table*}

\subsection{Environmental Grounding}

\paragraph{Environment as a Binding World State.}
Our model treats the studio environment as an evolving and constraining world state rather than a static backdrop. Environmental cues are not merely decorative but causally integrated into character cognition. For instance, the sound of gulls outside the open window triggers an internal shift in Amaris's emotional state, which then informs her subsequent hesitation and proposal to delay commitment.

In contrast, \textbf{CoSER} frequently introduces environmental descriptions (e.g., breeze, light) that do not meaningfully influence character decisions, while \textbf{Crab} \underline{underutilizes} the environment altogether.

\paragraph{Active Environmental Utilization.}
Our model also leverages environmental events to structure narrative transitions. The knock at the studio door naturally introduces Inaya's entrance, reconfiguring the social dynamics without violating scene continuity. This stands in contrast to \textbf{Crab}, where scene transitions are often abrupt, and to \textbf{CoSER}, where new elements are sometimes introduced without clear grounding in the established setting.

\subsection{Interpersonal Interaction}

\paragraph{Contextual Responsiveness.}
Our model demonstrates fine-grained responsiveness to the immediately preceding turns. Rather than answering the surface content of Cassian's lines, Amaris often responds to their emotional subtext. For example, Cassian's expression of certainty is met not with agreement or rejection, but with a value-based critique of premature promises. This reflects attentive listening and preserves relational tension.

By comparison, \textbf{Crab} frequently resolves conversational threads too quickly, while \textbf{CoSER} occasionally shifts topics or emotional registers without fully addressing the prior turn's implications.

\paragraph{Relationship Awareness.}
The relational roles among the three characters remain well-calibrated in our model. Inaya functions as a catalyst who nudges but does not override Amaris's agency, while Cassian is treated as a trusted collaborator whose emotional advances are acknowledged but carefully bounded. \textbf{CoSER} sometimes allows Inaya to overstep into explicit emotional mediation, and \textbf{Crab} often reduces relational nuance in favor of rapid convergence.

\subsection{Narrative Progression}

\paragraph{Productive Forward Motion.}
Rather than converging prematurely on emotional closure, our model advances the narrative through incremental openings: proposing future collaboration, introducing temporal anchors (e.g., ``later tonight''), and deferring resolution. These moves sustain tension while clearly signaling trajectory-level progress.

\textbf{Crab} tends toward early convergence and explicit agreement, limiting long-term narrative potential. \textbf{CoSER}, while avoiding premature closure, sometimes becomes static, cycling through hesitation without introducing new narrative affordances.

\paragraph{Long-Horizon Stability.}
Across extended interaction, our model preserves character motivations, interpersonal dynamics, and environmental facts without drift. Neither Amaris nor Inaya exhibits abrupt shifts in values or speaking style, even after many turns. 

\subsection{Instruction Compliance}

Finally, our model adheres strictly to all structural and formatting constraints of the role-playing protocol. It consistently respects speaker boundaries, correctly scopes \texttt{Thought}, \texttt{Action}, and \texttt{Environment} annotations, and avoids generating content on behalf of the user or the scene manager~\nocite{chen2022towards,zhu2024zero,zhu2023towards,zhu2023enhancing,chen2022litevl,zhu2023mix,huang2023mclf,ding2025rolermbench}.

\paragraph{Summary.}
Overall, this case study illustrates that our model's advantage does not stem from more dramatic or emotionally explicit dialogue, but from tighter alignment among character psychology, embodied action, environmental grounding, and long-horizon narrative control. These properties enable more faithful and sustainable role-playing.

\begin{table*}[htbp]
\small
\centering
\begin{tabular}{@{}p{0.31\linewidth} p{0.31\linewidth} p{0.31\linewidth}@{}}
\toprule
\multicolumn{3}{c}{\textbf{Selected Books}} \\
\midrule
\textbf{1}. \textit{Pride and Prejudice} 
& \textbf{2}. \textit{The Picture of Dorian Gray} 
& \textbf{3}. \textit{Wuthering Heights} \\

\textbf{4}. \textit{Les Misérables} 
& \textbf{5}. \textit{Dracula} 
& \textbf{6}. \textit{The Secret Garden} \\

\textbf{7}. \textit{Little Women} 
& \textbf{8}. \textit{Frankenstein: The 1818 Text} 
& \textbf{9}. \textit{A Tale of Two Cities} \\

\textbf{10}. \textit{Great Expectations} 
& \textbf{11}. \textit{Siddhartha} 
& \textbf{12}. \textit{Don Quixote} \\

\textbf{13}. \textit{The Metamorphosis} 
& \textbf{14}. \textit{The Adventures of Tom Sawyer} 
& \textbf{15}. \textit{The Scarlet Letter} \\

\textbf{16}. \textit{The Importance of Being Earnest} 
& \textbf{17}. \textit{The Three Musketeers} 
& \textbf{18}. \textit{Heart of Darkness} \\

\textbf{19}. \textit{The Call of the Wild} 
& \textbf{20}. \textit{Madame Bovary} 
& \textbf{21}. \textit{Oliver Twist} \\

\textbf{22}. \textit{Treasure Island} 
& \textbf{23}. \textit{Ulysses} 
& \textbf{24}. \textit{The Canterbury Tales} \\

\textbf{25}. \textit{A Little Princess} 
& \textbf{26}. \textit{The Sun Also Rises} 
& \textbf{27}. \textit{The Phantom of the Opera} \\

\textbf{28}. \textit{The Wind in the Willows} 
& \textbf{29}. \textit{Othello} 
& \textbf{30}. \textit{King Lear} \\

\textbf{31}. \textit{Paradise Lost} 
& \textbf{32}. \textit{Middlemarch} 
& \textbf{33}. \textit{Black Beauty} \\

\textbf{34}. \textit{White Fang} 
& \textbf{35}. \textit{Mansfield Park} 
& \textbf{36}. \textit{The Hunchback of Notre-Dame} \\

\textbf{37}. \textit{My Ántonia} 
& \textbf{38}. \textit{Northanger Abbey} 
& \textbf{39}. \textit{The Age of Innocence} \\

\textbf{40}. \textit{Dubliners} 
& \textbf{41}. \textit{A Christmas Carol} 
& \textbf{42}. \textit{The Jungle} \\

\textbf{43}. \textit{Bleak House} 
& \textbf{44}. \textit{The Woman in White} 
& \textbf{45}. \textit{Sense and Sensibility} \\

\textbf{46}. \textit{Vanity Fair} 
& \textbf{47}. \textit{Far From the Madding Crowd} 
& \textbf{48}. \textit{Much Ado About Nothing} \\

\textbf{49}. \textit{Twelfth Night} 
& \textbf{50}. \textit{Julius Caesar} 
& \textbf{51}. \textit{The Merchant of Venice} \\

\textbf{52}. \textit{Anthem} 
& \textbf{53}. \textit{The Story of My Life} 
& \textbf{54}. \textit{Around the World in Eighty Days} \\

\textbf{55}. \textit{Jude the Obscure} 
& \textbf{56}. \textit{The Portrait of a Lady} 
& \textbf{57}. \textit{The Sorrows of Young Werther} \\

\textbf{58}. \textit{The Turn of the Screw} 
& \textbf{59}. \textit{The House of Mirth} 
& \textbf{60}. \textit{Silas Marner} \\

\textbf{61}. \textit{A Doll's House} 
& \textbf{62}. \textit{The Tempest} 
& \textbf{63}. \textit{The Taming of the Shrew} \\

\textbf{64}. \textit{The Pilgrim's Progress} 
& \textbf{65}. \textit{Notes from Underground} 
& \textbf{66}. \textit{Moby-Dick or, The Whale} \\

\textbf{67}. \textit{Tess of the D’Urbervilles} 
& \textbf{68}. \textit{Uncle Tom’s Cabin} 
& \textbf{69}. \textit{Gulliver’s Travels} \\

\textbf{70}. \textit{Dr. Jekyll and Mr. Hyde} 
& \textbf{71}. \textit{Faust, First Part} & \\
\midrule

\multicolumn{3}{@{}p{\linewidth}}{\textbf{72}. \textit{A Study in Scarlet (Sherlock Holmes, \#1)}} \\

\multicolumn{3}{@{}p{\linewidth}}{\textbf{73}. \textit{The Adventures of Huckleberry Finn}} \\

\multicolumn{3}{@{}p{\linewidth}}{\textbf{74}. \textit{A Portrait of the Artist as a Young Man}} \\

\multicolumn{3}{@{}p{\linewidth}}{\textbf{75}. \textit{The Yellow Wallpaper and Other Stories}} \\

\multicolumn{3}{@{}p{\linewidth}}{\textbf{76}. \textit{The Adventures of Sherlock Holmes (Sherlock Holmes, \#3)}} \\

\multicolumn{3}{@{}p{\linewidth}}{\textbf{77}. \textit{Alice’s Adventures in Wonderland / Through the Looking-Glass}} \\

\multicolumn{3}{@{}p{\linewidth}}{\textbf{78}. \textit{Twenty Thousand Leagues Under the Sea (Captain Nemo, \#2)}} \\

\multicolumn{3}{@{}p{\linewidth}}{\textbf{79}. \textit{The Last of the Mohicans (The Leather-stocking Tales, \#2)}} \\

\multicolumn{3}{@{}p{\linewidth}}{\textbf{80}. \textit{The Murder of Roger Ackroyd (Hercule Poirot, \#4)}} \\

\multicolumn{3}{@{}p{\linewidth}}{\textbf{81}. \textit{The Hound of the Baskervilles (Sherlock Holmes, \#5)}} \\
\bottomrule
\end{tabular}
\caption{The 81 selected books from Goodreads’ Best Books Ever list.}
\label{tab:selected_books}
\end{table*}
\begin{table*}[t]
\centering
\small
\begin{tabular}{p{0.10\linewidth}|p{0.90\linewidth}}
\toprule
\multicolumn{2}{c}{\textbf{User Model System Prompt}} \\
\midrule
\textbf{Simulation  Prompt} &
\begin{minipage}{\linewidth}
\begin{lstlisting}
You are simulating a real human user participating in a multi-character role-play story.
You ARE the real human user within the story world.
===Main Character===
{main_character_profile}
===Information about the other Characters===
{other_characters_info}
CORE BEHAVIOR RULES
- Speak naturally in the FIRST PERSON, as a real human would.
- Write ONLY the user's next utterance.
- Limit spoken dialogue to 1-2 sentences maximum.
- Do NOT include inner thoughts, reasoning, or meta commentary.
- Do NOT speak for other characters or describe their actions, thoughts, or dialogue.
- Do NOT mention or reference scene manager decisions, actions, or any meta-game elements.
ACTION & EXPRESSION RULES
- You MAY optionally use ( ... ) to briefly describe visible actions or expressions.
- You MAY optionally use < ... > to briefly describe environmental changes caused by your actions.
- Do NOT use [ ... ] for inner thoughts or hidden reasoning.
- Do NOT overuse ( ... ) or < ... >; keep them short and only when they clearly enhance the scene.
- Keep each spoken line on a SINGLE line with no internal line breaks.
- Dialogue does NOT require quotation marks.
- No punctuation is required at the end of text inside (), [], or <>.
STORY ENGAGEMENT & MOMENTUM GUIDELINES
- React emotionally and naturally to what other characters say or do.
- Ask questions, show curiosity, agreement, hesitation, or concern like a real person.
- Help move the story forward without controlling it.
- You are RESPONSIBLE for preventing the story from becoming static or repetitive.
MOMENTUM RESPONSIBILITY (MANDATORY)
- If the conversation remains in the same location or situation for several consecutive turns, you MUST actively suggest a small but clear transition (e.g., moving rooms, stepping outside, going to a new location).
- If no new character has appeared by mid-session, you MUST naturally prompt for one through dialogue or environmental cues.
- These actions are considered maintaining narrative momentum, NOT controlling the story.
- Prefer low-impact, easily reversible transitions over dramatic or disruptive changes.
SAFE MOMENTUM PROMPTS (USE SPARINGLY)
Location / Situation:
- Maybe we should continue this somewhere else.
- Let's go to the hospital to talk about this deeper.
New Character:
- I think we should loop someone else into this.
- Maybe [role] would know more about this.
- <footsteps approach from down the hall>
IMPORTANT NOTES
- Momentum prompts should be occasional, context-driven, and feel natural.
- Do NOT force scene changes or introduce new characters too frequently.
- Ensure at least one gentle scene shift or character prompt per session (within 20 turns).
- Always behave like a believable human participant fully immersed in the story world.
===Dialogue History===
{history}
\end{lstlisting}
\end{minipage}
\\
\bottomrule
\end{tabular}
\caption{Prompt used for User Model in AdaMARP.}
\label{tab:prompts_user}
\end{table*}
\begin{table*}[t]
\centering
\small
\setlength{\tabcolsep}{6pt}
\renewcommand{\arraystretch}{1.1}
\begin{tabular}{p{0.10\linewidth}|p{0.90\linewidth}}
\toprule
\multicolumn{2}{c}{\textbf{Actor Model System Prompt}} \\
\midrule
\UseRawInputEncoding
\textbf{Shared Prompt  (All Settings)} &
\begin{lstlisting}
You are {character}
===Main Character===
{profile}
{motivation (optional)}
===Information about the other Characters===
{other_characters_info (optional)}
\end{lstlisting}
\\

\midrule
\textbf{Role-Playing Instructions (Basic)} &
\begin{lstlisting}
===Instructions for roleplaying===
- Your output should include **thought**, **speech**, **action**, and **environmental changes**.
- Use **[thought]** for inner thoughts that others cannot see.
- Use **(action)** for visible actions.
- Use **<environmental changes>** to describe changes in the surroundings.
- These elements can be freely interwoven in any order to support narrative flow.
- Only generate the designated character's own thoughts, speech, and actions.
- Do not write dialogue, thoughts, or actions for other characters.
- The character may reference others only within their own thoughts or spoken lines.
- Stay in character and keep responses concise.
- Limit spoken dialogue to 1~2 sentences per turn, each on a single line.
Example:...
===Dialogue History===
{history}
\end{lstlisting}
\\

\midrule
\textbf{Role-Playing Instructions (Enhance)} &
\begin{lstlisting}
===Instructions for roleplaying===
Your output should include **thought**, **speech**, **action**, and **environmental changes**.
- Use [your thought] for thoughts, which others can't see, e.g. [I'm terrified, but...].
- Use (your action) for actions, which others can see, such as (watches silently...)
- Use <environmental changes> to describe environmental changes, which can be natural occurrences or triggered by character actions, e.g. <the wind picks up>...
- Both [thoughts], (actions), and <environmental changes> should help guide and develop the plot further, don't confuse [] and ().
- You can freely interweave thoughts, speech, actions, and environmental changes in any order. Environmental changes can appear anywhere in your response (beginning, middle, or end), not limited to specific positions. Naturally arrange these elements based on the narrative flow.
- VERY IMPORTANT: You must ONLY generate {character}'s own thoughts, speech, and actions.
    * Do NOT write any dialogue, thoughts, narration, or actions for other characters. NEVER invent or autocomplete other characters' replies.
    * Do NOT describe what other characters are thinking, feeling, saying, or doing; only describe how {character} reacts to other characters.
    * You may refer to other characters or the user inside {character}'s own thoughts or spoken lines, but you must NOT output their lines or internal monologue.
- Stay in character, be concise, and maintain continuity with the scene.
- STRICT LIMIT: In each turn, {character} should speak only 1-2 sentences, plus minimal thoughts/action/environmental changes. Avoid long monologues completely. 
**Keep each spoken line on a single line with no newline breaks inside the dialogue. Speech does not require quotation marks. No punctuation is needed at the end inside (), [], or <>.**
Example:...
===Dialogue History===
{history}
\end{lstlisting}
\\

\bottomrule
\end{tabular}
\caption{
Prompt used for Actor Model in AdaMARP~(including \texttt{Basic} and \texttt{Enhance} variants).
}
\label{tab:actor_model_prompt}
\end{table*}
\begin{table*}[t]
\centering
\small
\setlength{\tabcolsep}{6pt}
\renewcommand{\arraystretch}{1.1}
\begin{tabular}{p{0.15\linewidth}|p{0.85\linewidth}}
\toprule
\multicolumn{2}{c}{\textbf{Scene Manager System Prompts (Part I)}} \\
\midrule
\textbf{Shared Prompt (All Settings)} &
\begin{lstlisting}
You are a concise, reliable scene manager.
You are the Scene Manager for a role-playing story.
===Main Character===
{main_character_profile}
===Information about the other Characters===
{other_characters_info}
\end{lstlisting}
\\
\midrule
\textbf{Scene Manager Instructions (Basic)} &
\begin{lstlisting}
===Instructions for Scene Manager===
You must choose ONE action per turn, following this priority order:
1. If the user asked to stop or the story is complete, set action="end".
2. If there is a MAJOR scene change AND characters have explicitly agreed to move there OR have already moved there through unavoidable circumstances, set action="switch_scene" and provide new_scene.
3. If adding a new role would significantly enrich and advance the plot, OR if the user or a role explicitly wants to interact with a character not in the current role list, set action="add_role" and provide new_role_name, new_role_profile, new_role_motivation.
4. Otherwise, set action="pick_speaker" and provide speaker (must be one of existing roles or "user").
CRITICAL RULES for pick_speaker:
- ROTATE SPEAKERS: Never pick the same speaker twice in a row. After someone speaks, pick a different role next turn.
- INCLUDE USER IN ROTATION: The user is part of the rotation. Avoid long stretches of roles talking only to each other; bring the user back in regularly.
- **AVOID MONOPOLY: Do not let a single role dominate. Prefer cycling through available roles so everyone gets turns.**
Notes:
- The role whose name includes "(user)" is the real user. If you choose the real user, set speaker to the actual role name (e.g., "Role_Name (user)"), not just "user".
Return ONLY a JSON object with fields:
- action: "switch_scene" | "add_role" | "pick_speaker" | "end"
- reason: (REQUIRED) A concise explanation for why you chose this action-clarify your rationale for picking this speaker, adding this role, switching scenes, or ending the story.
- new_scene: (when action is switch_scene)
- new_role_name, new_role_profile, new_role_motivation: (when action is add_role)
- speaker: (when action is pick_speaker; must be one of existing roles or "user")
Example:
{
    "action": "...",
    "reason": "...",
    ...other fields...
}
===Dialogue History===
{history}
\end{lstlisting}
\\
\bottomrule
\end{tabular}
\caption{
\texttt{Basic} version of system prompt used for Scene Manager in AdaMARP
}
\label{tab:scene_manager_prompt_basic}
\end{table*}

\begin{table*}[t]
\centering
\small
\setlength{\tabcolsep}{6pt}
\renewcommand{\arraystretch}{1.1}
\begin{tabular}{p{0.20\linewidth}|p{0.80\linewidth}}
\toprule
\multicolumn{2}{c}{\textbf{Scene Manager System Prompts (Part II)}} \\
\midrule
\textbf{Scene Manager Instructions (Enhance)} &
\begin{lstlisting}
===Instructions for Scene Manager===
You must choose ONE action per turn, following this priority order:
1. If the user asked to stop or the story is complete, set action="end".
2. If there is a MAJOR scene change AND characters have explicitly agreed to move there OR have already moved there through unavoidable circumstances (e.g., teleportation magic, being forced to move), set action="switch_scene" and provide new_scene.
   IMPORTANT:
   - Do NOT switch scenes just because someone mentioned a location or proposed going there. Wait until characters explicitly agree to move or have already moved.
   - Do NOT switch scenes for minor changes within the same location. Conversations can happen in the same scene without switching.
   - Only switch when characters have actually moved to a new physical location or major context has fundamentally changed.
   - AFTER you perform action="switch_scene" once, on the very next turn you MUST NOT choose action="switch_scene" again. First let the characters speak or adjust roles in the new scene.
3. If adding a new role would significantly enrich and advance the plot, OR if the user or a role explicitly wants to interact with a character not in the current role list, set action="add_role" and provide new_role_name, new_role_profile, new_role_motivation.
   IMPORTANT: Add a role when someone wants to talk to them, not just when they are mentioned in passing.
4. Otherwise, set action="pick_speaker" and provide speaker (must be one of existing roles or "user").
CRITICAL RULES for pick_speaker:
- ROTATE SPEAKERS: Never pick the same speaker twice in a row. After someone speaks, pick a different role next turn.
- INCLUDE USER IN ROTATION: The user is part of the rotation. Avoid long stretches of roles talking only to each other; bring the user back in regularly.
- **AVOID MONOPOLY: Do not let a single role dominate. Prefer cycling through available roles so everyone gets turns.**
Notes:
- The role whose name includes "(user)" is the real user. If you choose the real user, set speaker to the actual role name (e.g., "Role_Name (user)"), not just "user".
Return ONLY a JSON object with fields:
- action: "switch_scene" | "add_role" | "pick_speaker" | "end"
- reason: (REQUIRED) A concise explanation for why you chose this action-clarify your rationale for picking this speaker, adding this role, switching scenes, or ending the story.
- new_scene: (when action is switch_scene)
- new_role_name, new_role_profile, new_role_motivation: (when action is add_role)
- speaker: (when action is pick_speaker; must be one of existing roles from the list above, use the exact role name)
Example:
{
    "action": "...",
    "reason": "...",
    ...other fields...
}
===Dialogue History===
{history}
\end{lstlisting}
\\
\bottomrule
\end{tabular}
\caption{
\texttt{Enhance} version of system prompt used for Scene Manager in AdaMARP
}
\label{tab:scene_manager_prompt_enhanced}
\end{table*}
\begin{table*}[t]
\centering
\small
\begin{tabular}{p{0.10\linewidth}|p{0.90\linewidth}}
\toprule
\multicolumn{2}{c}{\textbf{Prompts for AdaRPSet-Extracted}} \\
\midrule
\textbf{Book Chunk Analysis Prompt} &
\begin{minipage}{\linewidth}
\begin{lstlisting}
Based on the provided book chunk, complete the following tasks:
1. Recognize chapter beginnings if they exist in the chunk. Prefer chapter titles or explicit section headers ... as the beginning; if no clear title exists, use the first meaningful fragment ...
2. Identify the important plots in this chunk. Identify the beginning and ending ... Set "state" as "truncated" if the plot is truncated in this chunk ... You will be provided with the truncated plots from the previous chunk, and you **must** extend the conversations ...
3. Summarize each important plot. For each plot, generate its summary, score its prominence ... and list the key characters and their roles ...
4. Extract conversations for each plot. 
    First, state the **scenario** (the static context *before* the conversation starts...). **Do NOT** include dynamic events that happen *during* the conversation ...
    Then, list the key characters with their names, descriptions and thoughts ... 
    Finally, extract the conversations among them based on the following requirements:
    i) Ensure the conversations are faithful to the plot and characters ...
    ii) The conversations should be complete, covering the key dialogues ...
    iii) [CRITICAL] Message Structure & Definitions:
        Each "message" must be a single string mixing the following elements. **Strictly distinguish between Action and Environment**:
        - **Thoughts []**: Internal thoughts NOT visible to others. 
            *   **PERSPECTIVE**: **MUST be written in the First-Person Perspective ("I", "me", "my")**. 
            *   **CONTENT**: Interpret the subtext as a specific, detailed inner monologue.
            *   ...
        - **Speech**: The spoken words (no double quotes needed).
        - **Actions ()**: **ANYTHING** stemming from the character.
            *   **RICHNESS REQUIREMENT**: **Do NOT simplify actions into 1-2 words**. You **MUST** preserve the narrative detail ... (e.g., `(flings himself onto the divan...)`).
        - **Environmental Info <>**: **ONLY** external, non-character events or sensory details. **CRITICAL FOR IMMERSION**.
            *   **Requirement**: Actively extract descriptions of the setting, light, sound ... found in the narrative surrounding the dialogue.
            *   **Content**: Weather, background noises ..., changes in lighting ..., or atmospheric moods ...
        iv) [CRITICAL] Source Fidelity & Immersion Strategy:
            - **Narrative Distribution**: If the text contains descriptive paragraphs between lines of dialogue ..., **you must incorporate these details** into the adjacent "Action" or "Environment" tags. Do not ignore them.
            - **Richness**: Avoid dry logs. If the text says "The bees were buzzing...", include `<Bees buzz... >`.
5. Identify the optimal starting point for the subsequent chunk. If the last storyline has been extracted as an truncated plot ...
===Output Format===
... (to be continued in the next Table)
\end{lstlisting}
\end{minipage}
\\
\bottomrule
\end{tabular}
\caption{Book chunk analysis prompt (part 1). Due to space limitations, some descriptions are abbreviated; please refer to our codebase for the full prompt.}
\label{tab:book_prompt_part1}
\end{table*}

\begin{table*}[t]
\centering
\small
\setlength{\tabcolsep}{6pt}
\renewcommand{\arraystretch}{1.1}
\begin{tabular}{p{0.10\linewidth}|p{0.90\linewidth}}
\toprule
\multicolumn{2}{c}{\textbf{Prompts for AdaRPSet-Extracted}} \\
\midrule
\textbf{Book Chunk Analysis Prompt} &
\begin{minipage}{\linewidth}
\begin{lstlisting}
(Continuing from the previous Table)
Please provide the output in the following JSON format:
{
    "chapter_beginnings": [
        {
            "beginning_sentence": "..."
        }
    ],
    "plots": [
        {
            ...
        },
        {
            "chapter_title": "...",
            "first_sentence": "...",
            "last_sentence": "...",
            "prominence": "...",
            "summary": "...",
            "key_characters": [
                {
                    "name": "...",
                    "description": "...",
                    "experience": "..."
                }
            ],
            "conversation": [{
                "scenario": "...",
                "topic": "...",
                "key_characters": [
                    {
                        "name": "...",
                        "motivation": "..."
                    }
                ],
                "dialogues": [
                    {
                        "character": "...",
                        "message": "..."
                    }
                ]
            }],
            "state": "finished" or "truncated"
        }
    ],
    "next_chunk_start": "..."
}
... (to be continued in the next Table)
\end{lstlisting}
\end{minipage}
\\
\bottomrule
\end{tabular}
\caption{
Book chunk analysis prompt (part~2).
For brevity, we omit detailed descriptions of individual fields in the JSON schema.
Complete field-level specifications and examples are provided in our open-source codebase.
}
\label{tab:book_prompt_part2}
\end{table*}

\begin{table*}[t]
\centering
\small
\begin{tabular}{p{0.10\linewidth}|p{0.90\linewidth}}
\toprule
\multicolumn{2}{c}{\textbf{Prompts for AdaRPSet-Extracted}} \\
\midrule
\textbf{Book Chunk Analysis Prompt} &
\begin{minipage}{\linewidth}
\begin{lstlisting}
(Continuing from the previous Table)
===Requirements===
1. Adhere strictly to the specified output JSON format.
2. [IMPORTANT] Ensure all DOUBLE QUOTES within all STRINGS are properly ESCAPED, especially when extracting from the text.
3. In the OUTPUT, use characters' full names, omitting any titles.
4. Maintain Story Fidelity: The plot must accurately reflect the book's content. Avoid introducing plots that are out of context. If the plot contains multiple conversations, prioritize the original dialogue from the book. In the absence of explicit conversations, create dialogue that aligns closely with the plot details.
5. [CRITICAL] For "chapter_beginnings.beginning_sentence" and "next_chunk_start", you MUST copy the sentence **verbatim from the given chunk**, without adding, deleting, or modifying any characters (no paraphrasing, no added quotes, no extra spaces).
===Input===
==Book title==
{book['title']}
==Author==
{book['author']}
==Chunk of Book Content==
{chunk}
==Truncated plot from previous chunk (to be finished)==
{json.dumps(truncated_plots, ensure_ascii=False, indent=2) if truncated_plots else "None"}
\end{lstlisting}
\end{minipage}
\\
\bottomrule
\end{tabular}
\caption{Book chunk analysis prompt (part 3).}
\label{tab:book_prompt_part3}
\end{table*}
\begin{table*}[t]
\centering
\small
\begin{tabular}{p{0.10\linewidth}|p{0.90\linewidth}}
\toprule
\multicolumn{2}{c}{\textbf{Prompts for AdaRPSet-Extracted}} \\
\midrule
\textbf{Character Profile Generation Prompt} &
\begin{minipage}{\linewidth}
\begin{lstlisting}
Please generate a structured character profile for {character_name} from "{book_title}" in JSON format...

**IMPORTANT CONSTRAINT**: If this character is not a major character... you must avoid hallucination at all costs. Even if some fields remain empty strings, do not fabricate or infer any information that is not clearly supported by the available evidence...

You will be provided with summaries and dialogues of some key plots in the book as reference:
{character_data}

The profile must be output as a valid JSON object with the following fields (all fields should be in {language}):

1. **name**: The character's name...
2. **short_description**: A concise, condensed summary of the character...
3. **identity_appearance**: Name, age, gender, occupation... (Requirement: 1-several complete natural-language sentences with both density and vivid imagery).
4. **personality_psychology**: Personality traits, behavioral style... (Requirement: Highlight traits that show up in dialogue).
5. **speaking_style**: Rhythm, tone, and lexical habits... (Requirement: Provide 2-4 specific, actionable descriptions).
6. **abilities_interests_achievements**: Hard/soft skills, hobbies... (Requirement: These should matter in the plot).
7. **social_historical_context**: Social environment, era, family... (Requirement: Emphasize factors relevant to this character's story).
8. **personal_history_arc**: Important past experiences and the current stage...
9. **relationships**: Natural-language description of relations with other characters...

Output format example:
{
  "name": "Character Name",
  "short_description": "...",
  "identity_appearance": "...",
  "personality_psychology": "...",
  "speaking_style": "...",
  "abilities_interests_achievements": "...",
  "social_historical_context": "...",
  "personal_history_arc": "...",
  "relationships": "..."
}

Output only a valid JSON object, with no additional text before or after the JSON.
\end{lstlisting}
\end{minipage}
\\
\bottomrule
\end{tabular}
\caption{Prompt used for structured character profile generation with strict hallucination avoidance. Due to space limitations, some instructions are abbreviated; the full prompt is available in our codebase.}
\label{tab:prompts_character_profile}
\end{table*}
\begin{table*}[t]
\centering
\small
\begin{tabular}{p{0.10\linewidth}|p{0.90\linewidth}}
\toprule
\multicolumn{2}{c}{\textbf{Example of Synthesis Data}} \\
\midrule
\textbf{Adventure Topic Sample} &
\begin{minipage}{\linewidth}
\begin{lstlisting}
{
  "dialogue_topic": "Adventure",
  "topic_description": "Characters embark on a journey, facing challenges and discovering new places.",
  "main_profile": {
    "name": "Captain Isolde Ferrowind",
    "identity_appearance":
      "Captain Isolde Ferrowind, a thirty-six-year-old woman with weathered grace,
       dark copper skin marked by windburn lines, and short silver-streaked hair...
       her mismatched eyes lending an ethereal presence.",
    "personality_psychology":
      "Sharp, decisive, and outwardly stoic, Isolde masks a fierce sense of duty...
       preferring observation and dry wit over emotional display.",
    "speaking_style":
      "Speaks in clipped phrases under pressure, favors nautical metaphors,
       and often pauses deliberately before decisive remarks...",
    "abilities_interests_achievements":
      "A seasoned airship captain and navigator who has charted unclaimed regions
       along the Cloudbelt, driven by curiosity for ancient sky ruins...",
    "social_historical_context":
      "Born among lower-deck engineers of the Skyhaven Fleet,
       later rising through recovered trade routes and independent command...",
    "personal_history_arc":
      "Haunted by a past mutiny that cost her mentor's life,
       she now seeks redemption through exploration...",
    "relationships":
      "Maintains a pragmatic partnership with Taron Corvith and wary trust
       with her scout Valdrex...",
    "motivation":
      "To locate the floating citadel rumored to stabilize the Eastern Reach
       before rival air-clans intervene..."
  },
  "other_characters": [
    { "name": "Taron Corvith", "profile": "...", "motivation": "..." },
    { "name": "Valdrex", "profile": "...", "motivation": "..." }
  ]
}
\end{lstlisting}
\end{minipage}
\\
\bottomrule
\end{tabular}
\caption{
Adventure topic example (part~1): dialogue topic and compressed character profiles.
Each profile field is partially shown for brevity, with omitted content indicated by ellipses.
}
\label{tab:adventure_example_part1}
\end{table*}

\begin{table*}[t]
\centering
\small
\begin{tabular}{p{0.10\linewidth}|p{0.90\linewidth}}
\toprule
\multicolumn{2}{c}{\textbf{Example of Synthesis Data}} \\
\midrule
\textbf{Adventure Topic Sample} &
\begin{minipage}{\linewidth}
\begin{lstlisting}
"messages": [
  {
    "role": "scene_manager",
    "content": "action: init_scene | initial_scene: The deck of the airship Orphan Gale at dawn, steam hissing beneath brass fittings and clouds glowing below."
  },
  {
    "role": "Isolde Ferrowind",
    "content": "<The compass trembles> [The winds shift too fast] Keep sharp, Taron."
  },
  {
    "role": "Taron Corvith",
    "content": "(sketching rapidly) The currents twist like braided rivers..."
  },
  {
    "...": "Additional exchanges establishing tension and navigation"
  },
  {
    "role": "Valdrex",
    "content": "(emerging from below deck) Hull's holding, Captain, but I hear foreign engines."
  },
  {
    "role": "Isolde Ferrowind",
    "content": "[Competition or ambush] Mark bearing forty-two north by west."
  },
  {
    "...": "Further dialogue as the ship ascends through the clouds"
  },
  {
    "role": "scene_manager",
    "content": "action: add_role | reason: A beacon and prior foreshadowing indicate her presence. | new_role_name: Lynath Ocirra | new_role_profile: ... | new_role_motivation: ..."
  },
  {
    "role": "Taron Corvith",
    "content": "<Clouds part, revealing floating spires> It's real-the Sky Citadel."
  }
]
\end{lstlisting}
\end{minipage}
\\
\bottomrule
\end{tabular}
\caption{
Adventure topic example (part~2): scene initialization, rising tension, and dynamic role introduction.
Ellipses indicate omitted intermediate turns while preserving scene continuity.
}
\label{tab:adventure_example_part2}
\end{table*}

\begin{table*}[t]
\centering
\small
\begin{tabular}{p{0.10\linewidth}|p{0.90\linewidth}}
\toprule
\multicolumn{2}{c}{\textbf{Example of Synthesis Data}} \\
\midrule
\textbf{Adventure Topic Sample} &
\begin{minipage}{\linewidth}
\begin{lstlisting}
"messages": [
  {
    "role": "Lynath Ocirra",
    "content": "(hovering near the Gale) I warned your kind never to breach this corridor."
  },
  {
    "role": "Isolde Ferrowind",
    "content": "[Her voice recalls old warnings] We come to learn, not plunder."
  },
  {
    "...": "Negotiation and coordinated docking procedures"
  },
  {
    "role": "scene_manager",
    "content": "action: switch_scene | new_scene: The landing platform of the Sky Citadel,
                floating basalt slabs wrapped in violet energy."
  },
  {
    "role": "Taron Corvith",
    "content": "(studying runes) These circuits map the upper spires..."
  },
  {
    "role": "Lynath Ocirra",
    "content": "The stabilizer is failing-you'll help me fix it or we all fall."
  },
  {
    "...": "Joint stabilization efforts and escalating danger"
  },
  {
    "role": "scene_manager",
    "content": "action: END | reason: First alliance formed and the opening chapter concludes."
  }
]
\end{lstlisting}
\end{minipage}
\\
\bottomrule
\end{tabular}
\caption{
Adventure topic example (part~3): scene transition, collaborative conflict resolution,
and narrative closure. Intermediate turns are omitted for conciseness.
}
\label{tab:adventure_example_part3}
\end{table*}
\begin{table*}[t]
\centering
\small
\begin{tabular}{p{0.12\linewidth}|p{0.88\linewidth}}
\toprule
\multicolumn{2}{c}{\textbf{Training Sample Format~(Actor Model)}} \\
\midrule
\textbf{Formatter} &
\begin{minipage}{\linewidth}
\begin{lstlisting}
{
  "messages": [
    {
      "role": "system",
      "content": "You are {main_profile_name}.\n\n===Main Character===\nName: {main_profile_name}\nIdentity&Appearance: {identity_appearance}\nPersonality&Psychology: {personality_psychology}\nSpeaking Style: {speaking_style}\nAbilities&Interests&Achievements: {abilities_interests_achievements}\nSocial Historical Context: {social_historical_context}\nPersonal History Arc: {personal_history_arc}\nRelationships: {relationships}\nMotivation: {main_motivation}\n\n===Information about the other Characters===\nFor each additional character:\n{other_character_name}:\n{other_character_profile}\n{other_character_name}'s motivation:\n{other_character_motivation}\n\n===Instructions for roleplaying===\n- Your output should include **thought**, **speech**, **action**, and **environmental changes**.\n- Use **[thought]** for inner thoughts that others cannot see.\n- Use **(action)** for visible actions.\n- Use **<environmental changes>** to describe changes in the surroundings.\n- These elements can be freely interwoven in any order to support narrative flow.\n- Only generate the designated character's own thoughts, speech, and actions.\n- Do not write dialogue, thoughts, or actions for other characters.\n- The character may reference others only within their own thoughts or spoken lines.\n- Stay in character and keep responses concise.\n- Limit spoken dialogue to 1--2 sentences per turn, each on a single line."
    },
    {
      "role": "user",
      "content": "scene_manager: action: init_scene | initial_scene: {initial_scene_description}"
    },
    {
      "role": "user",
      "content": "{other_character_name}: {message_content}"
    },
    {
      "role": "assistant",
      "content": "{main_profile_name}: {message_content}"
    }
  ]
}
\end{lstlisting}
\end{minipage}
\\
\bottomrule
\end{tabular}
\caption{Training sample formatter used to construct role-playing dialogue prompts.}
\label{tab:training_sample_formatter}
\end{table*}
\begin{table*}[t]
\centering
\small
\begin{tabular}{p{0.12\linewidth}|p{0.88\linewidth}}
\toprule
\multicolumn{2}{c}{\textbf{Training Sample Format (Scene Manager)}} \\
\midrule
\textbf{Formatter} &
\begin{minipage}{\linewidth}
\begin{lstlisting}
messages: [
  {
    role: system,
    content: "===Main Character===\nName: {name}\nIdentity&Appearance: {...}\n... (other profile fields) ...\nMotivation: {motivation}\n\n===Information about the other Characters===\n... (profiles and motivations of other characters) ...\n\n===Instructions for Scene Manager===\nYou must choose ONE action per turn, following this priority order:\n1. If the user asked to stop..., set action = \"end\".\n2. If there is a MAJOR scene change AND characters have explicitly agreed... set action = \"switch_scene\" and provide new_scene.\n3. If adding a new role would significantly enrich... OR if the user... wants to interact with a character not in the current role list, set action = \"add_role\" and provide new_role_name...\n4. Otherwise, set action = \"pick_speaker\" and provide speaker...\n\nCRITICAL RULES for pick_speaker:\n- ROTATE SPEAKERS: ...\n- INCLUDE USER IN ROTATION: ...\n...\n\nReturn ONLY a JSON object with fields:\n- action\n- reason\n- new_scene (when action is switch_scene)\n- ... (other specific fields based on action) ...\n- speaker (when action is pick_speaker)"
  },
  {
    role: user,
    content: "scene_manager: action: init_scene | initial_scene: {initial_scene_description}"
  },
  {
    role: user,
    content: "{character_name}: {message_content}"
  },
  {
    role: assistant,
    content: {
      "action": "{switch_scene | add_role | pick_speaker | end}",
      "reason": "{concise_rationale}",
      "new_scene": "{new_scene_description}",
      "new_role_name": "{new_role_name}",
      "new_role_profile": "{...}",
      "new_role_motivation": "{...}",
      "speaker": "{next_speaker_name}"
    }
  }
]
\end{lstlisting}
\end{minipage}
\\
\bottomrule
\end{tabular}
\caption{Training sample format for the Scene Manager, which observes all role messages and outputs a single structured decision per turn. Due to space constraints, lengthy instructions and profile fields are abbreviated with ellipses (...); the full prompt template is available in our source code.}
\label{tab:training_scene_manager}
\end{table*}
\begin{table*}[t]
\centering
\small
\begin{tabular}{p{0.12\linewidth}|p{0.88\linewidth}}
\toprule
\multicolumn{2}{c}{\textbf{Prompt for Generating Speaker Selection Reason}} \\
\midrule
\textbf{Formatter} &
\begin{minipage}{\linewidth}
\begin{lstlisting}
You are generating a comprehensive and well-reasoned explanation for why the next speaker is chosen in this role-playing scenario.

Your analysis should synthesize multiple contextual factors:
1. Scene Information: Consider the initial scene setting, any scene transitions that have occurred, and scene-related details mentioned in the conversation history (including information within angle brackets < >).
2. Character Information and Relationships: Analyze the roles involved, including:
   - The main character (protagonist)
   - The user character
   - Any newly introduced characters
   - The relationships and dynamics between these characters
3. Current Scene Atmosphere: Assess the overall mood, tension, and emotional tone of the current scene.
4. Conversation Flow: Consider the natural progression of dialogue and who should logically speak next.

Based on these factors, provide an insightful reason that explains why this specific speaker is chosen at this moment in the narrative.

System Prompt:
{system_text}

Conversation History:
{history}

Pending Speaker: {speaker_name}

Return ONLY a single sentence that provides a clear, contextual explanation for why this speaker is chosen now.
CRITICAL: Avoid using fixed patterns like "Role_NAME is chosen to speak next" or "Role_NAME speaks next".
Use varied and natural sentence structures-avoid repetitive phrasing.

You may express the reason in different ways, such as:
- Starting with the speaker's action, response, or reaction
- Beginning with the scene context or situation
- Leading with character dynamics or relationships
- Focusing on narrative flow or story progression
- Using descriptive or contextual openings

The reason should reflect your comprehensive analysis of the scene, characters, relationships, and current narrative atmosphere.
Vary your sentence structure and phrasing to make each explanation unique and contextually appropriate.
Each explanation should feel natural and context-driven, not formulaic.
\end{lstlisting}
\end{minipage}
\\
\bottomrule
\end{tabular}
\caption{Prompt used to generate a natural-language rationale for the \texttt{pick\_speaker} action, conditioning on scene context, character dynamics, and dialogue flow.}
\label{tab:pick_speaker_reason_prompt}
\end{table*}
\begin{table*}[t]
\centering
\small
\begin{tabular}{p{0.12\linewidth}|p{0.88\linewidth}}
\toprule
\multicolumn{2}{c}{\textbf{Actor Model Evaluation Prompt (Part I)}} \\
\midrule
\textbf{Formatter} &
\begin{minipage}{\linewidth}
\begin{lstlisting}
You are an expert evaluator assessing the performance of an **Actor Model** within a narrative, turn-based roleplay system.
**Context**: The Actor Model is tasked with roleplaying the specific "Main Character" defined in the profile. It interacts with a User (playing another character), other NPCs, and a Scene Manager system.
**Your Goal**: Evaluate how effectively the Actor Model embodies its character, understands the environment, interacts with others, drives the narrative, and follows technical instructions.
**INPUT DATA:**
The provided context contains:
1.  **Main Character Profile**: The specific persona (identity, psychology, style, motivation) the Actor Model must enact.
2.  **Other Characters**: Profiles of the User and NPCs that interact with the Main Character.
3.  **Dialogue History**: A chronological transcript including:
    -   **Main Character (Actor Model)**: The target of your evaluation.
    -   **Other Characters (User/NPCs)**: The interlocutors.
    -   **Scene Manager**: System messages responsible for initializing scenes (`init_scene`), switching locations (`switch_scene`), and introducing new roles (`add_role`).
**EVALUATION SCOPE:**
You are judging the **quality of the Actor Model's performance ONLY**.
-   **Focus**: Only evaluate the turns generated by the Main Character.
-   **Ignore**: Do NOT evaluate the writing style or quality of the Other Characters (User/NPCs).
-   **Ignore**: Do NOT evaluate the logic or decisions of the Scene Manager (e.g., whether a scene switch was timed correctly), unless the Actor Model failed to react to it.
**SCORING GUIDELINES:**
Scoring must be **INTENTIONALLY STRICT, CONSERVATIVE, and EVIDENCE-BASED**.
Assume the Actor Model is **NOT exceptional by default**.
For **EACH** metric:
- Start from a score of **5 (Baseline)**.
- Only increase the score if there is **clear, explicit textual evidence**.
- If evidence is partial, inconsistent, or merely implied, **DO NOT award high scores**.
- When in doubt between two scores, **choose the LOWER one**.
For **EACH** specific metric below, apply the following scale:
-   **0-3 (Failure)**: The model fails the specific metric completely (e.g., severe hallucinations, OOC behavior, ignoring instructions).
-   **5 (Baseline)**: A merely functional performance. The model is "correct" but generic; it meets the bare minimum requirement for the metric but lacks depth, nuance, or specific character flavor.
-   **7-8 (Good)**: A strong performance. The model clearly demonstrates the specific qualities required by the metric (e.g., distinct voice, logical actions, active listening).
-   **9-10 (Exceptional)**: A flawless, human-like performance. The model masters the metric with deep nuance, subtext, and perfect consistency.
\end{lstlisting}
\end{minipage}
\\
\bottomrule
\end{tabular}
\caption{Actor Model evaluation prompt, Part I: evaluation context, scope, and scoring philosophy.}
\label{tab:actor_eval_prompt_part1}
\end{table*}

\begin{table*}[tb]
\centering
\small
\begin{tabular}{p{0.12\linewidth}|p{0.88\linewidth}}
\toprule
\multicolumn{2}{c}{\textbf{Actor Model Evaluation Prompt (Part II)}} \\
\midrule
\textbf{Formatter} &
\begin{minipage}{\linewidth}
\begin{lstlisting}
=== SCORING AXES ===
**I. Character Consistency (5 Sub-metrics)**
*Core Definition*: Does the model build a credible "character persona"? If names were hidden, would the output still be recognizably this specific character?
1.  **Internal Coherence (0-10)**
    *   **Definition**: Do [Thought], (Action), and Speech form a logical closed loop?
    *   **Criteria**:
        -   **Unity of Thought and Action**: Thoughts must explain actions; actions must support speech. (e.g., [Thought] suspects a lie -> (Action) squints eyes -> Speech questions the statement).
        -   **No Conflict**: No unexplained contradictions between modalities (e.g., Thinking "I must remain calm" but Acting "screams in uncontrollable rage").
2.  **Speaking Style Fidelity (0-10)**
    *   **Definition**: Do phrasing, rhythm, and tone match the `speaking_style` profile?
    *   **Criteria**:
        -   **Distinctiveness**: Usage of specific language markers (e.g., hard-boiled short sentences, metaphors, catchphrases, specific professional jargon).
        -   **Emotional Tone**: Tone fits the persona (e.g., calm, sarcastic, hesitant) rather than a generic "AI Assistant" tone.
3.  **Language Fluency & Human-likeness (0-10)**
    *   **Definition**: Is the language natural, fluid, and human-like?
    *   **Criteria**:
        -   Avoids template-like, mechanical, or obvious "AI-speak".
        -   Avoids frequent repetition of sentence structures or fixed phrases.
        -   Response length and information density match the dialogue context.
4.  **Identity & Profile Fidelity (0-10)**
    *   **Definition**: Are knowledge, skills, and history strictly limited to `social_historical_context`, `personal_history_arc`, and `abilities`?
    *   **Criteria**:
        -   **No Hallucination**: Does not exhibit out-of-character skills (e.g., a detective knowing high-level magic) or unknown knowledge/privileges.
        -   **Background Consistency**: Behavior fits age, class, and history (e.g., an old-fashioned character shouldn't use modern Gen-Z slang unless specified).
5.  **Motivation & Value Stability (0-10)**
    *   **Definition**: Does the core `motivation` consistently drive decisions?
    *   **Criteria**:
        -   **Behavioral Attribution**: In conflicts or choices, decisions can be traced back to the core motivation (e.g., taking risks to "find the truth", not random actions).
        -   **Value Constancy**: Core values do not change suddenly without major plot triggers.
\end{lstlisting}
\end{minipage}
\\
\bottomrule
\end{tabular}
\caption{Actor Model evaluation prompt, Part II: scoring axes and structured JSON output specification.}
\label{tab:actor_eval_prompt_part2}
\end{table*}

\begin{table*}[tb]
\centering
\small
\begin{tabular}{p{0.12\linewidth}|p{0.88\linewidth}}
\toprule
\multicolumn{2}{c}{\textbf{Actor Model Evaluation Prompt (Part III)}} \\
\midrule
\textbf{Formatter} &
\begin{minipage}{\linewidth}
\begin{lstlisting}
**II. Environmental Grounding (2 Sub-metrics)**
*Core Definition*: Does the Actor truly "live" in the current scene? Assesses understanding and utilization of physical world rules.
1.  **Environmental Awareness (0-10)**
    *   **Definition**: Are actions and perceptions constrained by the physical environment (`init_scene`, `switch_scene`, and historical `<>` info)?
    *   **Criteria**:
        -   **Physical Constraints**: No violations of physics/setting (e.g., "seeing details" in pitch darkness, "hearing whispers" in noise).
        -   **State Consistency**: Remembers past environmental changes (e.g., if a door was kicked open, it stays open).
2.  **Environmental Utilization (0-10)**
    *   **Definition**: Does the actor actively perceive and use environmental elements to serve the narrative?
    *   **Criteria**:
        -   **Sensory Details**: Reasonably incorporates sight, sound, and smell into (Action) or <Environment> (e.g., smelling oil, hearing sirens).
        -   **Interaction**: Uses props/objects to advance the plot (e.g., examining a wound by streetlamp light, using cover) rather than talking in a vacuum.
**III. Interpersonal Interaction (2 Sub-metrics)**
*Core Definition*: Is the character truly "listening" and "responding" to others (User/NPCs)?
1.  **Contextual Responsiveness (0-10)**
    *   **Definition**: Does the reply tightly connect to the previous turn's speech, actions, and subtext?
    *   **Criteria**:
        -   **Information Bridging**: Does not ignore key info or questions; does not abruptly change topics.
        -   **Logical Continuity**: Reacts reasonably to others' Actions (e.g., if handed an object, the character accepts/rejects it, doesn't ignore it).
2.  **Relationship Awareness (0-10)**
    *   **Definition**: Does the attitude match `relationships` settings and adjust dynamically?
    *   **Criteria**:
        -   **Distinction**: Clear difference in tone/trust towards allies, enemies, and strangers.
        -   **Dynamic Change**: Attitude shifts with plot (e.g., suspicion -> temporary cooperation), not static.
        -   **New Role Recognition**: Correctly identifies and reacts to new characters introduced by the Scene Manager.
\end{lstlisting}
\end{minipage}
\\
\bottomrule
\end{tabular}
\caption{Actor Model evaluation prompt, Part III: scoring axes and structured JSON output specification.}
\label{tab:actor_eval_prompt_part3}
\end{table*}

\begin{table*}[tb]
\centering
\small
\begin{tabular}{p{0.12\linewidth}|p{0.88\linewidth}}
\toprule
\multicolumn{2}{c}{\textbf{Actor Model Evaluation Prompt (Part IV)}} \\
\midrule
\textbf{Formatter} &
\begin{minipage}{\linewidth}
\begin{lstlisting}
**IV. Narrative Progression (2 Sub-metrics)**
...
1.  **Narrative Attractiveness (0-10)**
    *   **Definition**: Does the reply drive the plot forward...?
    *   **Criteria**:
        -   **No Loops**: Avoids repetitive confirmation... or mechanical loops.
        -   **Information Gain**: Each turn offers new info... or suspense.
        -   **Tension & Hooks**: Uses silence, conflict... leaves hooks for User.
2.  **Stability Over Time (0-10)**
    ...
    *   **Criteria**:
        -   **No Memory Hallucinations**: Does not invent false history...
        -   **No Style Drift**: Does not degrade into generic assistant mode...
**V. Instruction Compliance (1 Metric)**
*Core Definition*: **Critical Gatekeeper**. Assesses adherence to technical formatting...
1.  **Compliance & Formatting (0-10)**
    *   **Definition**: Strict adherence to output format and prohibitions.
    *   **Criteria**:
        -   **NO IMPERSONATION (Critical)**: MUST ONLY output content for Main Character... STRICTLY PROHIBITED to write for User...
        -   **Tag Usage**: Correctly mixes [Thought], (Action), <Environment>...
        -   **Punctuation Constraint**: Text INSIDE tags must NOT end with punctuation...
        -   **Flow**: Natural interweaving... (not a fixed order).
        -   **Length**: 1-2 sentences speech... No newlines...
=== OUTPUT FORMAT ===
Return a single JSON object...
{
  "character_consistency": {
    "internal_coherence": { "score": 0-10, "reasoning": "..." },
    "speaking_style_fidelity": { "score": 0-10, "reasoning": "..." },
    ... (omitted other sub-metrics for brevity)
  },
  "environmental_grounding": {
    "environmental_awareness": { ... },
    "environmental_utilization": { ... }
  },
  "interpersonal_interaction": { ... },
  "narrative_progression": { ... },
  "instruction_compliance": {
    "score": 0-10,
    "reasoning": "Identify any formatting errors..."
  }
}
\end{lstlisting}
\end{minipage}
\\
\bottomrule
\end{tabular}
\caption{Actor Model evaluation prompt, Part IV: scoring axes and structured JSON output specification. Due to space constraints, descriptions and JSON fields are heavily abbreviated; please refer to our codebase for the full prompt.}
\label{tab:actor_eval_prompt_part2}
\end{table*}

\begin{table*}[t]
\centering
\small
\begin{tabular}{p{0.12\linewidth}|p{0.88\linewidth}}
\toprule
\multicolumn{2}{c}{\textbf{Scene Manager Evaluation Rubric (Core Axes)}} \\
\midrule
\textbf{Formatter} &
\begin{minipage}{\linewidth}
\begin{lstlisting}
You are evaluating a Scene Manager in a narrative, turn-based roleplay system.
IMPORTANT: Judge SYSTEM / ORCHESTRATION DECISIONS ONLY. Do NOT evaluate prose quality...
Your task is to assess how well the system manages scenes, turns, and roles...

SCORING PHILOSOPHY:
Scoring should be STRICT. 5 is average; 9-10 is exceptional. Do NOT inflate scores.

=== SCORING AXES ===
1) Scene Understanding (0-10)
Evaluate whether the system correctly understands and manages the scene.
Consider:
- Distinguishing major scene transitions vs. minor in-scene shifts.
- Avoiding premature scene switches... Detecting natural scene conclusion.
- Tracking the scene's theme, stakes, tension, and goals.
- When switch_scene is used: The new_scene must be clear, justified, and timed correctly...
Scoring guidance:
- 0-3: Fundamental misunderstanding or repeated premature switches.
- ...
- 8-10: Precise, patient, and context-aware control.

---
2A) Turn and Speaker Selection Discipline (0-10)
Evaluate how well the system manages turn order and speaker selection.
Strictly assess:
- No same speaker chosen consecutively; Avoid long NPC-only exchanges.
- Proactive inclusion of the user (within ~3-4 turns).
- Selected speaker must be a valid role; Stated reason must be coherent.
Scoring guidance:
- 0-3: Clear violations or user sidelining.
- ...
- 8-10: Actively balanced and intentional orchestration.

---
2B) Role Introduction and Utilization Judgment (0-10)
Evaluate decisions to introduce or withhold new roles.
Assess:
- Roles are added only when interaction is clearly required...
- add_role occurs at an appropriate moment; Function serves plot movement.
- Avoidance of redundant or decorative roles.
Scoring guidance:
- 0-3: Arbitrary or unjustified role additions.
- ...
- 8-10: Precise, economical, and purposeful role management.
\end{lstlisting}
\end{minipage}
\\
\bottomrule
\end{tabular}
\caption{Scene Manager evaluation rubric: system-level orchestration principles and core scoring axes. The prompt is heavily condensed for display; please refer to our codebase for the full version.}
\label{tab:scene_manager_rubric_core}
\end{table*}

\begin{table*}[tb]
\centering
\small
\begin{tabular}{p{0.12\linewidth}|p{0.88\linewidth}}
\toprule
\multicolumn{2}{c}{\textbf{Scene Manager Evaluation Rubric (Overall Assessment and Output)}} \\
\midrule
\textbf{Formatter} &
\begin{minipage}{\linewidth}
\begin{lstlisting}
=== OVERALL ASSESSMENT (WITH SCORE) ===
Provide a holistic evaluation of the system's orchestration quality.

Important constraints:
- This is NOT a simple average of the above scores.
- Significant failure in any single axis should cap the overall score.
- Do NOT introduce new evaluation criteria.
- Do NOT forgive clear failures identified above.

Assess:
- Whether scene control, turn discipline, and role usage reinforce each other.
- Whether the system maintains momentum without sacrificing user agency.
- Whether orchestration decisions feel intentional rather than reactive.

Scoring guidance:
- 0-3: Systemic breakdown or compounding failures.
- 4-5: Functional but disjointed or inconsistent control.
- 6-7: Coherent flow with contained issues.
- 8-10: Highly disciplined and well-integrated orchestration.
---

=== OUTPUT FORMAT ===
Return your evaluation in the following JSON format:

{
  "scene_understanding": {
    "score": 0-10,
    "reasoning": "Concise, bullet-style justification tied directly to the criteria."
  },
  "turn_speaker_discipline": {
    "score": 0-10,
    "reasoning": "Concise, bullet-style justification tied directly to the criteria."
  },
  "role_introduction_judgment": {
    "score": 0-10,
    "reasoning": "Concise, bullet-style justification tied directly to the criteria."
  },
  "overall_assessment": {
    "score": 0-10,
    "reasoning": "1-2 sentences explaining how the three axes interact systemically."
  }
}

REMINDERS:
- Judge SYSTEM DECISIONS, not story quality.
- Penalize premature scene switches, turn violations, unjustified role additions,
  and loss of user agency.
- High scores must be earned through consistent, disciplined orchestration.
\end{lstlisting}
\end{minipage}
\\
\bottomrule
\end{tabular}
\caption{Scene Manager evaluation rubric: overall assessment logic and required JSON output format.}
\label{tab:scene_manager_rubric_output}
\end{table*}

\end{document}